# Abstractive Text Summarization for Contemporary Sanskrit Prose: Issues and Challenges

*Thesis submitted to Jawaharlal Nehru University in partial fulfillment of the requirements for the award of the degree of*

**DOCTOR OF PHILOSOPHY**

by

Shagun Sinha

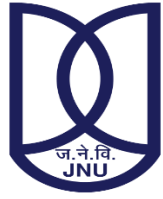

School of Sanskrit and Indic Studies

Jawaharlal Nehru University

New Delhi – 110067

India

**December 2022**

# Acknowledgments

*om gaṇapataye namaḥ, om sarasvatyai namaḥ, om lakṣmyai namaḥ, om durgāyai namaḥ, om sai nath!*

I cannot believe I am finally writing my acknowledgments page. I had so deeply hoped for this day to come and it is finally here. Over the past few years, this thesis has stayed with me. This thesis – my Pikachu – is the first thing I will ever acknowledge. I have seen people treat their thesis as their baby that gets to see the world after years of love and hard work poured in by its academic parent. But for me, this thesis is not my baby – it is my academic life partner! We started as strangers and years later, here we are – ready to be partners forever. We have seen so many ups and downs in this journey, yet here we are stronger than ever. Cheers to us, Pikachu, we did it!

This research work deserves attention for reasons more than one. It might be the first work of its kind in Sanskrit. But that is not its only unique aspect. It carries with itself years of hard work and love – not of me, but of the people in my life who made it possible for me to even think of doing something like this. I, for one, am an average learner and I give up way too easily. Yet, I persisted for the last few years working on this thesis and more importantly on myself, honing my research skills all because people around me made me realize how important Pikachu is to me.

I commence by thanking my supervisor, Prof. Girish Nath Jha. He has been such an amazing supervisor and an indispensable source of support for me. His courses, his guidance, and his support have been so crucial in getting this thesis done. I have always admired him for his skills and knowledge and as I complete this work, I must say, my admiration of his knowledge has only grown stronger. Thank you for being such an amazing supervisor and an equally amazing human being, Prof. Jha.

The one professor I want to dedicate this thesis to, the best professor in the world, is the late Prof. Satyamurti. It shatters my heart to use 'late' with his name. Prof. Satyamurti, wherever you are, I am sure that place is far better than this. You transformed my life and learning. I will always miss you. I am here because of you. I wish you could read my thesis.

Special thanks and immense gratitude to Dr. Santosh K. Shukla immensely for his indispensable guidance in the initial years of my research journey. I also thank all my professors from the Sanskrit school for their constant support and guidance. Gratitude to Prof. Sudhir Kumar, the current Dean of the Sanskrit school for his unmatched support during the submission and advisory process. I have learnt greatly from my teachers here, Prof. Shashi Prabha Kumar, Prof. Ram Nath Jha, Prof. C. U. Rao, Prof. Rajnish Kumar Mishra, Prof. Hari Ram Mishra, Prof. Brajesh Pandey, Prof. Gopal Meena, and Dr. Jyoti. Million thanks to them.

While my teachers have been amazing, my first set of Gurus to come into my life truly were my Parents and my sister. I want to absolutely thank Papa first. I wonder if a thank you is ever going to suffice for the indispensable support he has given me. He inspired me to think of NLP early on in my life and even though I was a complete novice to the field back then, it was because of his guidance – and solely because of him – that Pikachu is here today. Thank you, Papa! Thank YOU. YOU are the reason this thesis is complete today.

Pikachu came into existence because of Papa, but *my own* existence was sustained because of two other people - Maa and Swadha. Cannot thank you both for *constantly* listening to my breakdown rants and speeches of emotional outbursts. For the hundreds of chai round, millions of chats, billions of hugs, and uncountable love and support – I love you both! Maa's unconditioned love is something I will never stop cherishing. It is a tough job to be a mother, even tougher to be *my* mother. But she aced the job like a queen! Rare are parents who become best friends to their children. My parents *are* my bestest friends ever. Talking of best friendship, my sister, Swadha, is the bestestest friend (even more than my parents, yes :D). She is t.h.e. person I need in times of distress to pour me with love (and Maggie! :D). As I often joke (but mean in all totality) – what would I do without the love of you all!

Talking of love, I have to deeply thank the crowning member of our family, Dadi – a big thank you for nurturing us all. The four of us are nurtured by your affection and kindness. Not to miss, Dadu, Nana, and Nani – wherever they are, they are nurturing us remotely! Deep love to all elders of the family, my cousins, and my extended family members for their brilliant company and encouragement always. Special mention to Satyendra Mausa whose wise words from his PhD days were the most needed during moments of distress.

I have always maintained that my love affair with Sanskrit started in my adolescent years. I fell in love with the beauty of Panini's grammar and continue to be in awe of it. I would have never even thought of pursuing Sanskrit had it not been for my teacher Mrs. Madhu Prakash who introduced me to Sanskrit and Panini. I am forever indebted to her. As they rightly say, a student is only as good as the teaching of her teachers. So, here I am, learning with fun all due to the teaching of my teachers!

Talking of good teaching, I also wish to express my deepest gratitude to my best public speaking mentor, Mrs. Neeta Pant, who made me fall in love with the stage! And and and, Special thanks to Mrs. Madhubala, Mrs. Sarla Seth, Mrs. Nirmal Saxena, Mrs. Asha Srivastava, Mrs. G. R. Khare, Mrs. Veena Michael, Mrs. Hema Tiwari, Mrs. Sophia Elizabeth, Mrs. Alka Verma, Dr. Pankaja Ghai, Dr. Shivani Dubey, and all other teachers from school and college who totally made me the confident woman I am today!

The trail of good teaching and supervision has been present even during my research journey. Special gratitude to Dr. Garima Dalal from the LEC who introduced me to the idea of research terminologies and their usage.

I learned a great deal of NLP from my friend, Ankur, whose guidance was the initial impetus required for me to move forward in the field. I remain indebted to him for his professional guidance. Speaking of professional guidance, I am lucky to have had the opportunity to teach throughout the duration of my research and be in touch with some of the best students in the country, especially the COP batch of 2019-20, and the elementary Sanskrit batches of 2020 to 2022. Their questions and doubts have enhanced my thinking in numerous ways. I am deeply grateful to them.

I am ever grateful to the human evaluators and assessors who graciously agreed to undertake the tasks and give invaluable inputs. Thank you very much for your timely help. The researcher in me also thanks her seniors, students, and friends – Shiladitya Bhattacharya, Kabi bhaiya, Srishti, and friends from college who provided the needed inputs from time to time. The list of supportive peers is endless. I am indebted to all the scholars from computer science and linguistics who helped me in clarifying the myriad questions I inundated them with almost every day during the final 6 months of this work. They have all been indispensable to this work's progress and I am deeply grateful to each of them.

My very special friend, Dr. Shipra Verma, has been instrumental in getting things in place during my illness. Thank you, Shipra, for everything you did for me and my family!

This acknowledgment can never be complete without thanking Shabnam Mam, Manju Mam, and other members of the department who have now become family. I thank them for being such amazing people to be around!

And finally, the great loves of my life – Sanskrit, public speaking, teaching, along with the institutions I have been associated with, namely, KV, LSR, JNU, UofZurich, and Ashoka – thank you for being with me always. The most special love of my life, Sanskrit, has been the gift of some supreme power to me. I always thank God for that one day in 2006 when I randomly decided to pick Sanskrit as an optional in 9th grade, not just because it was a high-scoring subject but also because I wanted to 'try it out and see how I do in it'. Little did I know that that one random decision was going to give me a lifetime of joyful learning. Thank God Sanskrit exists (I might have been a mediocre civil engineer somewhere, otherwise :P). I cherish the Sanskrit journey that I have completed so far as well as the journey I am yet to mark with it. Come what may, Sanskrit always has my back! All my institutions and my dearest Sanskrit, I carry you in my heart. I always will!

अत्यन्तमेव प्रसन्नचित्ताऽस्म्यहम्!

Shagun

# List of Abbreviations

| ATS | Abstractive Text Summarization |
|---|---|
| AMR | Abstract Meaning Representation |
| BERT | Bidirectional Encoder Representations from Transformers |
| CNN | Convolutional Neural Network |
| Colab | Google Colaboratory |
| CV | Computer Vision |
| DistilBERT | Distilled BERT |
| DL | Deep Learning |
| EL | Evaluation Loss |
| ELMo | Embeddings from Language Model |
| ETS | Extractive Text Summarization |
| FFNN | Feed Forward Neural Network |
| GAN | Generative Adversarial Network |
| GCP | Google Cloud Platform |
| GPT | Generative Pre-Training |
| GRU | Gated Recurrent Unit |
| HRL | High Resource Language |
| IL | Indian Language |
| ILTS | Indian Language Text Summarization |
| iNLTK | Indic NLTK |
| LM | Language Model |
| LLM | Large Language Model |
| LR | Low-Resource |
| LRL | Low-Resource Language |
| LSA | Latent Semantic Analysis |
| LSTM | Long Short-Term Memory |
| MDS | Multi-Document Summarization |
| MKB | Mann Ki Baat |
| ML | Machine Learning |
| MLLM | Multilingual Language Model |
| Mono-LM | Monolingual Language Model |
| MTL | Multi-Task Learning |
| MURIL | Multilingual Representations for Indian Languages |
| NLG | Natural Language Generation |
| NLP | Natural Language Processing |
| NLU | Natural Language Understanding |
| NN | Neural Network |
| OOV | Out of Vocabulary |
| PTLM | Pre-trained Language Model |
| RND | Random |
| RNN | Recurrent Neural Network |
| RoBERTa | Robustly Optimized BERT Pretraining Approach |
| ROUGE | Recall-Oriented Understudy of Gisting Evaluation |
| RQ | Research Question |
| RST | Rhetorical Structure Theory |
| SATS | Sanskrit Abstractive Text Summarization |
| SC | Sentence Compression |
| SDS | Single Document Summarization |
| SM | Summarization Model |
| SOTA | State of the Art |
| SR | Semantic Representation |
| TRC | TPU Research Cloud |
| TRL | Transfer Learning |
| TL | Training Loss |
| TS | Text Summarization |
| ULMFiT | Universal Language Modeling and Fine Tuning |
| VAE | Variational Auto-Encoder |
| WSD | Word Sense Disambiguation |

# List of Tables

# List of Figures

# Table of Contents

# Chapter 1: Introduction

This thesis presents abstractive text summarization models for contemporary Sanskrit prose and the challenges faced in developing the models. Different methods for text summarization (TS) training have long existed. However, some of the prevalent deep learning (DL) methods that have impressively succeeded in English and other high-resource languages (HRLs) serve as the motivation to apply the models to Sanskrit[1] TS.

Natural language processing (NLP) is a field of covering computer science and linguistics which develops computational models for processing, generating and analyzing natural language inputs (Lusetti, 2018). TS is a field under NLP in which models are trained to summarize important information from source texts. Two approaches to TS are popular in literature, extractive TS (ETS) and abstractive TS (ATS).

ETS summarizes the key information by picking sentences which contain the important information of the source text (Barve et al., 2016). ATS, on the other hand, summarizes the key information using new words. As a result, ATS involves more linguistic analysis for better processing and hence, it is more human-like in its summarization process (Chen & Bansal, 2018; Mishra & Gayen, 2018).

The approach with which an NLP task is performed has evolved over time. In TS, the earlier works were based on surface and linguistic features like word frequency, cue phrases etc.

---

[1] Also written as *saṃskṛta* although 'Sanskrit' has been used throughout this work.

(Luhn, 1958). However, with the advent of machine learning (ML) and deep learning (DL), more semantic-based methods have emerged as possible methods which exploit the underlying structure of a text to process it (Chowdhury et al., 2021; Dohare et al., 2018). Since ATS requires more linguistic processing, DL methods have greatly advanced ATS.

In ATS, systems produce summaries of long documents using new words. Different approaches and methods have been deployed in different languages to train various algorithms since the late 1950s. However, with neural methods, the training of TS algorithms has gained greater heights and has been applied on more languages than ever before – such as Slovenic (Žagar & Robnik-Šikonja, 2021), Bengali (Sultana et al., 2022), etc. The wide-range utility of DL models in TS is further testified by the fact that despite the typical large training data requirement of neural (DL) models, such models have been successfully trained and tested in low-resource settings as well (Bajaj et al., 2021; Maimaiti et al., 2019). As a result, it is now possible for low-resource settings and DL models of NLP to meet at crossroads.

This thesis is set in the backdrop of this crossroad. It queries the challenges in the implementation of the existing neural methods of ATS in Sanskrit. By reporting and analyzing the challenges, this work aims to add to the current understanding of the interaction between machines and languages for the sequence generation task of ATS. This is the first attempt at Sanskrit ATS (SATS) so far.

## 1.1 Motivation – Why This Research?

I envisage two probable objections to this research.

The first objection may concern the utility of this project. "*Why should a summarizer for Sanskrit be built at all?"* some may seek to know. Three key facts answer that question and hence, serve as the motivation for this work.

- First, an ATS may automatically summarize the vast literature of Sanskrit and provide to the research community for better access. The abundant literature of Sanskrit, including a large body of manuscripts, *may* be automatically summarized if a researcher aims to get an overview of manuscripts before studying them. As Reddy et al. (2018) observe, with large-scale digitization efforts in Sanskrit like the Göttingen Register of Electronic Texts in Indian Languages (GRETIL)[2], the Digital Corpus of Sanskrit (DCS) (Krishna et al., 2017), Digital Corpus of Sanskrit[3] and Sanskrit Library of India[4], many manuscripts and texts have reached the public domain. However, due to technological and processing challenges, they are mostly inaccessible (Reddy et al., 2018). A summarization system would enhance accessibility by summarizing the contents of a text thereby enabling a prospective reader to decide whether or not to read it.
- Second, Sanskrit offers an interesting field of research in NLP in general. The inflectional property of Sanskrit is one characteristic to be studied for summarization challenges just like in NLP which is increasingly seeking to adapt to inflectional languages (Hellwig, 2015a).
- The third and simplest reason is that no abstractive summarization system has yet been built in a particular setting of Sanskrit which is a low-resource language (LRL). Literature offers many examples of ATS built in low-resource settings like low-resource data (Khemchandani et al., 2021) and low-resource domains (Bajaj et al., 2021). However, Sanskrit has not yet been tested for ATS.

---

[2] gretil http://gretil.sub.uni-goettingen.de/gretil.html
[3] Digital Corpus of Sanskrit, http://www.sanskrit-linguistics.org/dcs/
[4] Sanskrit Library https://sanskritlibrary.org/

In other words, the first motivation for this work is the literary abundance of Sanskrit that can benefit from TS – that is, the large body of Sanskrit manuscripts could be summarized using SATS making those manuscripts more accessible to researchers. The second point of motivation is to broaden the scope of NLP for inflectional and morphological languages in general and Sanskrit, in particular. The final reason is the fact that Sanskrit is a language with low resources and a research of this kind may improve our understanding of ATS for LRLs. The last reason forms the core of this thesis.

The other probable objection to the motivation behind this work could be why does this thesis focus on an ATS and not ETS? Two facts answer that question: First, ATS has been regarded as a better form of TS than ETS, owing to its coherent summaries and human-like language (Kouris et al., 2021). Hence, building an ATS for Sanskrit would enhance the current Sanskrit processing abilities of machines just as it would for any other language. Second, an ETS for Sanskrit has already been attempted with promising results (Barve et al., 2016).

Additionally, some preprocessing requirements specific to Sanskrit, including for tasks that are not conventionally applicable in other languages, have emerged as promising avenues for applying current neural methods. For example, Sanskrit texts usually contain *sandhi* and *samasa* the automatic segmentation of which has been approached as a sequence-to-sequence processing task by many scholars (Hellwig, 2015b; Reddy et al., 2018). A double-decoder sequence has also been used to segment *Sandhi* (Aralikatte et al., 2018). Thus, present-day neural sequence-to-sequence methods have been applied to Sanskrit-specific tasks. This thesis is an extension in that direction, attempting the sequence-to-sequence task of ATS in Sanskrit.

## 1.2 Research Question

Given the *why* of this research work above, the *what* of this thesis is focused on training neural models for SATS and the key research question I answer is – what are the key challenges faced when ATS is developed for Sanskrit?

To answer the primary research question given above, the following are the sub-research questions based on four different themes that I have posed throughout this work:

1. **About training method**
    a. What are the ways of training an ATS model in the literature?
    b. Which methods can be used in this research work?
    c. Why did this research work choose a given method?
2. **About data domain**
    a. What is the desired dataset format for training?
    b. What are the challenges of building such a dataset?
    c. Why contemporary Sanskrit prose?
    d. What issues does data preprocessing pose?
3. **About the training process**
    a. How was a specific training method used in this research?

b. What are the challenges of deploying the chosen training method?

4. **About generated summaries**

   a. What are the properties of the system-generated summaries?

   b. How should the summaries be evaluated?

   c. Why did some systems perform better?

   d. How can the summary quality be improved in the future?

The four themes above form the basis for the four core chapters of this work.

- Questions based on the first theme have been answered in the chapter, literature review, which surveys the key works in TS and also establishes the methodology used in this research. The literature review also answers one question from the second theme, question 2(c).
- Questions other than 2(c) from the second theme have been answered in the chapter, data preparation, which describes the preparation and preprocessing challenges.
- Likewise, questions from the third theme are presented in the chapter on experiments and results which presents the steps for implementing the training and the resources use for it.
- Finally, questions from the fourth theme have been covered in the chapter on results evaluation and discussion which analyzes the system-generated summaries, presents the evaluation scheme, and suggests path for future work.

The research questions will be answered within the limits of the following:

- First, in developing the dataset, I assume that texts from Sanskrit literature, especially poetry and related literary domain have complexities as a result of which literature as a whole is an unsuitable domain for the first attempt at ATS.
- Second, I do not develop any new approach but I only train and evaluate the existing models on the dataset developed in this work.

This research is based on a set of concepts which will be used to describe the existing works. The next section describes the framework of the key concepts and related terminologies used throughout this work.

## 1.3  Conceptual Framework – The How of this Work

Text summaries are mainly of two types – extractive (ETS) and abstractive (ATS). The first attempt at TS was by Luhn (1958) in which summaries of scientific articles were generated using surface-level features like word frequency, ranking sentences based on significance, and selecting the top-n best ranked sentences as the summary. His methods were extractive. The first attempt at ATS, however, was made very late after Luhn (Jones, 1999; Torres-Moreno, 2014). As noted by van Yperen et al. (2021), the headline-generation approach to TS by (Banko et al., 2000) was the first ATS-like system. FRUMP system developed at Yale is also one of the earliest ATS systems developed although very few attempts were made initially (Torres-Moreno, 2014). ATS is considered a preferred way of summarization because of its coherent summaries and consequently better quality. However, the metrics for evaluating the quality of TS systems varied for a long time before a metric called Recall-Oriented Understudy for Gisting Evaluation (ROUGE) was first proposed in (Lin, 2004). ROUGE has become a standard metric for most TS systems evaluation. Chapter 5 will explore evaluation in detail.

### 1.3.1 Shallow Features vs Semantic Representation

TS in the initial years was more extractive owing to the lack of resources needed for deeper analysis required in abstraction. Statistical NLP dominated TS research for a long time with most efforts being extractive in nature as well as using surface-level features to gauge importance of a sentence before extracting it (Ferreira et al., 2013; Luhn, 1958). As a result, initial TS works analyzed a source text for its important sentences and ranked them according to certain threshold. If a sentence crossed that threshold, it was considered in the summary (Luhn, 1958). The sentence ranking process was based on shallow surface features used to represent the text such as word frequency, presence of certain cue-phrases, keyword presence, etc. (Edmundson, 1969). Such features are surface-level linguistic features which rely on the presence of tokens or specific phrases in a sentence to assess its importance.

On the other hand, representing the meaning of source text through graphs like abstract meaning representation (AMR) graphs and other techniques is called semantic representation (SR) which helps in improving semantic understanding (Mishra & Gayen, 2018). It may at times include syntactic representation methods such as part of speech (POS)-tagging, graph representation, named entity recognition (NER)-based methods, etc. (Embar et al., 2013; Liu et al., 2018).

**1.3.2   Machine Learning**

The growth of ML and DL paved way for networks that could generate deeper representations of the source text enabling better understanding of the source text by the machine. With rise in efficient computational resources, abstractive summarization also became more popular and implementable. Starting with vanilla feedforward neural network (FFNN), NLP gradually moved to recurrent neural network (RNN) and long-short term memory (LSTM), gated recurrent unit (GRU) networks to handle sequential data such as natural language text (Goldberg, 2017). Sequential data like natural language sentences have words with long-range dependencies, i.e., a word at the starting of a sequence may be important to the last or later word in a sentence. As Section 1.3.4. explains, mapping such related words is an important challenge in NLP which the DL models continue to solve. Thus, neural networks and different learning methods could dig deeper than linguistic features to study the semantic features of a given input in natural language.

Both ML and DL need a lot of data to train on (Han et al., 2021). They primarily offer three methods of learning patterns from data – supervised learning (SL), unsupervised learning (USL), and reinforcement learning (RL). SL methods involve large-scale parallel dataset to learn input-output labeling. SL methods are widely used in classification tasks. USL refers to learning on large-scale unlabeled data and is mostly used in clustering, grouping, etc. RL trains a model while constantly giving feedback to the system in terms of rewards for improving its performance (Narayan et al., 2018b).

### 1.3.3 Representations of Text

In moving from the word-level and surface-level to deeper semantic representations, one important development came with vector-based representation. The vector-based representations of text made textual analyses more meaningful, especially after the introduction of distributional hypothesis leading to distributed representations. Distributional hypothesis is the idea that words that occur in related contexts tend to have the same meaning (Firth, 1957; Harris, 1954). This hypothesis paved way for distributed representation of words with the aim to make machines better understand the context of a word, understanding similar words based on their meaning, and differentiating between them based on their contexts and semantics (Goldberg, 2017). Algorithms for distributed representations learn word representations from large scale monolingual corpora through unsupervised learning. Word2Vec algorithm for word embeddings is an algorithm for distributed representations introduced by (Mikolov, Sutskever, et al., 2013; Mikolov, Yih, et al., 2013). These vector models made reasoning possible with words given their ability to capture underlying semantics (Hobson et al., 2017).

### 1.3.4 Encoder-Decoder Structures

Despite enhancements in semantic representations and advances in neural network architectures which could process sequential inputs, certain sequence generation tasks such as machine translation and text summarization posed immense challenges (Sutskever et al., 2014). Sequence-to-Sequence (Seq2Seq) learning problems required mapping an input sequence to an output sequence. Issues in these were solved by encoder-decoder architectures made their entry proving to be extremely beneficial for the implementation of sequence-to-sequence tasks (Sutskever et al., 2014). Encoder portion converts the input sequence into a hidden vector which is used by decoder to sequentially decode and convert into a desirable output sequence.

The authors note that deep neural networks (DNN) are useful for classification tasks where input data dimension is not significant. However, for sequence generation tasks, input dimensions must be considered. The method used by Sutskever et al. (2014) used LSTMs for encoding as well as decoding inputs sequences thereby permitting the use of sequences.

However, LSTM- and RNN-based encoder decoder setups could not handle long-term dependencies. A solution to this problem was introduced in 2014 called attention mechanism. Attention was introduced by (Bahdanau et al., 2014) for better alignment in MT tasks. Attention is a context vector that is conveys which portions of the input are important (Bahdanau et al., 2014). However, attention also has some disadvantages for which coverage or distraction mechanism was introduced (Lin & Ng, 2019, p. 9819).

While sequence-to-sequence models by Sutskever et al. (2014) and attention mechanism by Bahdanau et al. (2014) had led to great improvements in DL-based NLP, it was the work by Rush et al. (2015) that first used neural methods for a paraphrase generation task which is close to summarization. The authors used the encoder-decoder architectures have been a great boon in resolving many sequential language processing tasks. With neural methods advancing in TS, many other solutions have also developed mostly in the form of modifications to the encoder-decoder models. One such advancement has been the advent of Transformers which led to a great development in NLP and TS.

### 1.3.5 Transformer Architecture

Vaswani et al. (2017) introduced Transformer architecture which is a special type of neural encoder-decoder model that uses self-attention and is aimed at improving the ability of the model to capture important portions of the source text. Self-attention or intra-attention is based on Key (K), Query (Q) and Value (V) vectors which are added as per the formula in (Vaswani et al., 2017):

$$Attention\ (Q, K, V) = softmax\left(\frac{QK^T}{\sqrt{d_k}}\right)V \quad \ldots (1)$$

Transformers enhanced sequence processing through self-attention. The next section elaborates on their use in the direction of enhanced language understanding.

### 1.3.6 Distributed Representation and Transformers

Word embeddings such as Word2Vec (Mikolov, Yih, et al., 2013) outperformed the earlier language models in capturing relevant semantic content. However, those embeddings were not contextualized – that is, for a given word, the embeddings remained the same across different contexts. With ELMo (Peters et al., 2018), contextualized embeddings became possible. Models in NLP and TS have moved from non-contextualized word embeddings like word2vec, to contextualized embeddings like ELMo (Peters et al., 2018), and then to large Transformer-based language models (LM) which are the current state-of-the-art (SOTA) (Devlin et al., 2019; Rothe et al., 2020). With BERT and Transformer-based models (Devlin et al., 2019), the contextualized LMs became huge enough to capture language generalities and began being fine-tuned on downstream tasks while achieving results better than ELMo (Martin et al., 2020; Nemeskey, 2020). Section 2.5 discusses Transformers at length.

The key terms in current NLP space thus include – DL, ML, Transformers, LM. The next chapter will explore these in detail.

## 1.4 Contribution of this Thesis

In addition to initiating the first work in SATS and improving our understanding of neural methods for Sanskrit processing, this thesis contributes the following resources:

1. A cleaned dataset for Sanskrit language model.
2. Transformer-based checkpoints for SATS.
3. A pipeline for building future works in the direction.

## 1.5 Resources Acknowledgment

This work uses the following resources:

1. Google Colab Pro+ services, TPU Research Cloud (TRC), and Google Cloud Platform (GCP) support systems for training environments.
2. HuggingFace Platform that made some of the most popular research works and models accessible free of cost (Wolf et al., 2019).
3. Methodology from Rothe et al. (2020) which form the crux of this work.
4. Publicly available Sanskrit resources including open access journal Anantaa, translated version of the speeches of the Prime Minister of India, collectively titled, *Mann ki Baat*, shared by Dr. Baldevanand Sagar, and OSCAR data shared by the developers (Suárez et al., 2019).

# Chapter 2: Literature Survey

This chapter surveys TS systems and approaches developed in English and Indian Languages (ILs). The larger aim of this chapter is to establish the what of this research – that is, the survey begins with the traditional methods of approaching TS and gradually moves to establish the core methodology used in this work – the use of pre-trained language models (PTLMs) for the downstream task of TS. Through this survey, I argue that TS can be approached in Sanskrit via PTLMs and fine-tuning for a summarization task. Therefore, I deal with research questions related to the first theme – the training method, particularly, RQs marked [1(a), 1(b), 1(c)](), and [question 2(c)]() from the second theme.

PTLMs with fine-tuning on downstream tasks have been accepted as current SOTA methods for many sequence generation and inference tasks (DeLucia et al., 2021; Nadeem et al., 2020). This chapter summarizes the history of the TS and how LM-based methods came to be used for sequence-to-sequence tasks like TS. Since this work is focused on an inflectional language, this chapter briefly looks at how the case of inflectional languages is handled through tokenization enhancement in the literature of PTLMs. The key research questions of this thesis – can ATS in Sanskrit be initiated and if yes, what are the issues and challenges in building a Sanskrit ATS – will be answered through the lens of PTLMs and fine-tuning.

In any case, the development of technology has largely impacted the TS pipeline. The technology used for summarization – is the basis for division into classical and neural methods. In both eras all other factors remained constant to the goal – summarize a certain number of document(s) of a given language and domain through some degree of source text analysis. However, classical methods were non-neural and thus, they approached source text analysis differently than how neural methods did (Lin & Ng, 2019). Classical methods stayed on the surface and involved many subtasks for saliency capturing in summarization while neural methods could refer to a complete *end-to-end* system that performs summarization (Lin & Ng, 2019, p. 9818, [emphasis original]).

Torres-Moreno (2014) observes different types of summaries based on their purpose including - indicative, informative. Additionally, he classifies TS systems on the following criteria (pp. 11-12):

**Type:** extractive if the summaries are collected from the source text and abstractive if the summaries are produced using new words. A third type of summary categorized by him under this division is sentence compression (SC) which I discuss in a later section.

**Purpose:** Indicative – Summaries that reflect the main ideas or topics of a text, may be used to gain an overview of what the source text is about. Indicative summarization is aimed at summarizing the key idea of the source text. Informative – Summaries that may be used as a replacement of the document and whose purpose is to contain all important information in the text

**Genre of Source Text:** Summaries pertaining to a specific domain of the source text (Specialized summaries), or literary genre, or news genre etc.

**Based on Context:** Query-guided – Summaries that are targeted at a specific word or concept, usually given by a user. Update – When a user who knows all background information on a topic only needs the new information to get updated. Generic – is a general summary type with no focus on any specific type of information.

**Number of Source Text Documents:** Single Document or Multi-document TS systems exist. When only one source text/document is summarized, the TS system is called single document summarization (SDS) (Nagwani & Verma, 2011). On the contrary, when more than one documents are summarized, the system is called multi-document summarization (MDS) (Li, 2015).

Thus, TS classification has many layers. Both ETS and ATS are impacted by different factors. They include:

- purpose of summarization,
- number of source texts,
- type of source text analysis,
- language of the source text (as well as summary)
- domain of the source text

In addition to the above, the following two factors also impact TS pipeline:

- desirable characteristics[5] of the generated summaries, and
- the technology used for summarization.

This thesis aims at generic single document abstractive summarization system. The upcoming sections survey the TS systems based on one or more of the remaining factors above. The next section surveys the current text analysis methods for TS which is one of the core factors impacting the work presented in this thesis.

---

[5] compression, language quality, etc.

## 2.1 Methods of Source Text Analysis

Different processing methods are used in developing a TS system or machine to make it understand the source text before summarizing it. Given that, the first criterion of TS classification is based on how the machine processes the text for 'understanding' it better. Mani and Maybury (1999) divide TS approaches into three groups based on the level of text processing – surface-level, entity-level, and discourse-level. Similar to them, Radev et al. (2002) discuss the idea of three types of analysis done by an algorithm for ETS: surface-level, intermediate level, and deep parsing. The earlier surface-level methods dominated the classical approaches while neural methods used the deeper analysis methods.

### 2.1.1 Surface Level Analysis

Surface-level algorithms operate on the word-level of a text without considering any deeper linguistic meanings (Torres-Moreno, 2014, p. 31). Surface-level algorithms use methods of analysis that are based on cue-phrases, word location, term frequency etc. The first work in TS (Luhn, 1958) was based on surface-level features where the presence of significant words at certain positions in a sentence was used to calculate its significance factor and in turn, its eligibility to be included in the summary. Edmundson (1969); McCargar (2004); Myaeng and Jang (1999) used statistical methods exploiting surface-level information to summarize.

Entity-level methods described by Mani and Maybury (1999) exploit the relatedness among words and sentences that helps gauge the importance of sentences to be included in the summary. The other similar domain of intermediate methods involved extracting strong lexical chain relationships between sentences (Torres-Moreno, 2014, p. 33).

## 2.1.2 Semantic Representation

Surface-level methods are shallow in their analysis and they barely consider the deeper semantic representation of content. Such methods use shallow linguistic hints in the text like cue phrases or locating text positions (Barzilay & Elhadad, 1999). As a result, most of these methods are extractive in nature.

For ATS, improved SR methods are sought. SR in classical methods was not entirely a based on deep parsing but was able to serve as intermediate representation which was enough to process meaning to an extent (Torres-Moreno, 2014). Such methods of analysis include lexical chains (Barzilay & Elhadad, 1999; Brunn et al., 2001). Abstraction may be achieved by analyzing the coherence and cohesion of text units and sentences (Barzilay & Elhadad, 1999). Ontology, used mostly in domain-specific summarizations, is a way to represent domain-specific terminologies and their relationship (Sotudeh et al., 2020). As a result, TS systems incorporate domain ontologies to enhance text representations and generate more domain-specific summaries.

Another SR technique is discourse-representation. Discourse-level analysis is a similar approach of extracting text information from discourse units in which the global structure of the document is processed to find the summary-worthy content. Discourse-level processing paved way for deeper understanding of the text. Myaeng and Jang (1999) indicate that more semantic understanding of text leads to summaries that are closer to human-generated summaries. It is from such thoughts that abstractive methods gain traction.

Discourse Analysis (Disc-Ana) aims to exploit the discourse structure of a given text to analyze a given text. Such analysis has been used for empirical evaluation of human summaries (Seidlhofer, 1991). In the NLP sphere, Disc-Ana methods find the underlying relationships between different units of a text (Smith, 2011). Disc-Ana methods create graphs of discourse units to better understand the relationship between different discourse units for better analysis. Such methods greatly enhance the meaning extraction part of NLP. A Disc-Ana theory, called, Rhetorical Structure Theory (RST) (Mann & Thompson, 1987), is popularly used to annotate rhetorical units in a text and to exploit the argumentative structure of a text (Chengcheng, 2010). That rhetorical analysis is then used for finding important portions in the text and summarizing them. Similar to RST, cohesion theory has been used to study the relationship between sentences and then use those relationships to better summarize the documents (Lebanoff et al., 2020). Discourse-based parsing delves on a deeper structural level for identifying source information (Torres-Moreno, 2014, p. 34).

Among the other approaches to understanding the semantics of a document, latent semantic analysis (LSA) is one which is a method to find semantically similar words and phrases and to also calculating the vector scores for similarity (Hobson et al., 2017). LSA-based TS is developed in Gong and Liu (2001); Ozsoy et al. (2011) but those are more extractive in nature. A related approach to understanding the role of individual words in a document with respect to a given task is a statistical technique called Term Frequency-Inverse Document Frequency (TF-IDF) where the frequency of a given word in a document is multiplied with the inverse of its frequency across documents. The inverse frequency serves as the basis for ranking a given document or sentence and this method is popular in ETS along with other information retrieval (IR) methods (Barve et al., 2016).

SR for improving textual understanding is achieved through graph-representations like abstract meaning representation (AMR) graphs (Liu et al., 2018) and tree-based representations. Graphs mostly help in analyzing the relatedness among different units like words, sentences, concepts or discourse-units. Graph-based methods represent the entities of a text (words, sentences, etc.) into nodes and the relation between them as edges (Balaji et al., 2016; Moawad & Aref, 2012; Tan et al., 2017; Veena et al., 2016). Graphs are then used for either extracting important sentence or generating abstracts.

Tree-based methods represent the text as a tree and important concepts/sentences are accommodated accordingly (Marcu, 1999; Sunitha et al., 2016). This process of capturing saliency requires that the algorithm should first understand the underlying meaning of the source text through some methods. All the methods of analysis described above lead to either a structure- or semantic-based TS. The former is based on collecting salient sentences in a structure or template whereas semantic-based approaches use SR or some mechanism for analyzing the meaning of the source text before summarizing the text (Sunitha et al., 2016). Source text analysis, therefore, is a lengthy task which requires many tasks to better understand complete source text.

## 2.2 Language of the Source Text

The next important factor in developing a TS is the language of the source text. As noted earlier, low-resource domains are challenging areas for TS. Similar to the domain of the source text, the language of source text can also be low-resource. Languages that lack resources for tool building are called low-resource languages (LRLs). Indian Languages (ILs) – languages which are spoken in the Indian subcontinent - are also LRLs and often lack digital resources due to which developing NLP tools becomes tough. Since Sanskrit is also an LRL, this low-resource problem is the second challenge that this thesis tackles. A survey of ATS and TS in ILs indicates that both traditional and DL methods have been used for developing Indian Language Text Summarization (ILTS) systems.

### 2.2.1 Indian Language Text Summarization

ILTS have also made use of different approaches for summarization. Like English, ILTS has also witnessed both extractive and abstractive summarization. The initial methods were focused on surface-level and linguistics features including sentence-ranking methods thereby leading to extractive systems. Through Rich Semantic Graph (RSG), Hindi ETS has been developed in (Dalal & Malik, 2013; Thaokar & Malik, 2013). Graph methods have also been used for Marathi ETS (Sarwadnya & Sonawane, 2018). Other extractive works in ILs include sentence-framing technique in Malayalam (Kishore et al., 2016), keyword extraction- and similarity-based ETS for Telugu (Naidu et al., 2018), Urdu ETS based on sentence weighting and other extraction techniques (Nawaz et al., 2020), key phrases-extraction based ETS for Punjabi (Gupta & Lehal, 2012, 2013), Gujarati (Ranchhodbhai, 2016), Information Retrieval (IR)-based extractor for Bengali (Islam & Al Masum, 2004). In addition to all these methods, ILTS has used linguistic resources like WordNet since WordNet helps in tracing important concepts (Baruah et al., 2019; Kalita et al., 2012).

ATS efforts through conventional techniques have seen a rise. Subramaniam and Dalal (2015) use RSG for Hindi ATS. Bengali ATS have used different methods to abstract important information (Chowdhury et al., 2021; Masum et al., 2019; Sarker, 2021). Similar efforts to generate abstractive summaries through conventional techniques in Tamil (Priyadharshan & Sumathipala, 2018), Konkani (D'Silva & Sharma, 2021). ATS efforts in Kannada through POS-tagging and information extraction methods (Embar et al., 2013; Kallimani & Srinivasa, 2011; Shilpa & DR, 2019) are commendable. Neural methods in ILTS are discussed in section 2.5.1.

## 2.3 Classification of TS based on source text domain

Domain refers to the field or topic on which the source text is based. Capturing information or jargon specific to the field is, therefore, crucial. In classical TS, ontology serves as a knowledge base for a given domain on which knowledge is derived for the summarization process. Ontology is another way of accessing information of the text thereby improving its understanding of the text and capturing the information contained therein (Mohan et al., 2016; Verma et al., 2007). Ontology is also important in accessing domain-specific terminologies which is used for ATS in (MacAvaney et al., 2019). Ontology-based analysis thus provides the TS module with ample information for summarization like clinical domain (MacAvaney et al., 2019). The neural methods, on the other hand, have been used for domain-specific summarization without ontologies (Ma et al., 2022).

The first effort at building a TS was made in Luhn (1958) in which sentences were evaluated on the presence of significant words and their relative positions to decide the significance. The significant sentence so selected form the summary. Luhn's focused on the domain of scientific articles. Similar works in the initial phases of TS focused on scientific articles (Pollock &

Zamora, 1999), technical documents (Edmundson, 1969). Similarly, other domain could be news articles (Liu et al., 2022), geosciences (Ma et al., 2022), biomedical (Mishra et al., 2014; Plaza & Carrillo-de-Albornoz, 2013; Reeve et al., 2006). A slightly similar aspect of source document is the format of source document. The source to be summarized could be books, news articles, or medical documents, etc. The format of a given data often influences the organization of information in the document – for example, information in news articles is often said to be organized under "layout bias" with important information located in the beginning (Kryściński, Keskar, et al., 2019, p. 544). Since source documents are often analyzed by an algorithm before being processed for summarization, the format of the source document plays a crucial role in deciding the information extraction method. Besides news articles, some other source text types have been used in the literature as follows:

**Book and Blog Summarization:** Some have also explored summarizing books (Ceylan & Mihalcea, 2009). Blogs are another domain that have seen summarization given the last amount of text and information content contained in there. Avinesh et al. (2021) develop a live blog summarization method**.**

**Chat, Meeting, and Discussion Summarization**: Online Chats have been used at the source text in (Forsythand & Martell, 2007) and the summarization method is based on discourse analysis for modeling the chats. Although the domain of chats is not relevant to this research work, the domain of chat modeling through discourse analysis is interesting. A related domain is of meeting summarization is meeting summarization (Feng et al., 2020). Online news discussions have been treated as a multi-document text which can be analyzed for various common links between different posts (Tampe et al., 2021).

**Story and Narrative Summarization**: Processing narrative threads in a story is also a subject of summarization and narrative texts has been used as the source text even in the pre-neural network-era. For example, stories have been summarized through combination of plot units in (Lehnert, 1999). Related to this, DeLucia et al. (2021) study the different NLG methods in generating a narrative summary.

A detailed discussion on any of the aforementioned domains of summarization is out of the scope of this thesis.

Most ATS systems are built using news data which is often criticized for layout bias (Narayan et al., 2018a), (Sharma et al., 2019). That is, important portions may appear on some selected positions and training models on such data may train it to forever focus on those specific text positions only. A prospective solution to summarizing news articles without any positional biases is segmentation-based TS. Segmentation-based TS summarizes every segment or paragraph and then present the set of sectional summaries as the final summary (Liu et al., 2022). Therefore, domain impacts how information may be presented in a given text and helps in deciding the right form of preprocessing. Thus, once a domain-specific data has been finalized, the next expected step is to process the document so as to sift important information. Selection of the important information, given a domain is the key challenge in developing any TS (Radev et al., 2002, p. 399).

## 2.4 Synthesis

The previous sections discussed three major aspects influencing the pre-development stage of a TS tool development – method of analyzing source text for tracing important information, language of source text, and domain and format of source text. The classical methods were largely surface-level based and failed to capture the underlying structure and meaning. As a result, linguistic resources like WordNet and deep text analysis methods like discourse analysis, knowledge-based methods like ontology-based summarization were used (Kogilavani & Balasubramanie, 2009).

In the next section, this chapter surveys TS based on neural methods which have greatly advanced NLP and TS (Devlin et al., 2019; Rush et al., 2015). When using neural networks - which have recently been proven to be excellent sequence processors too - for summarization, the three aforementioned factors – domain, language, document analysis methods continue to remain important factors. However, these neural methods differ from classical methods in that they require huge data for training (Adadi, 2021). As a result, to make the most of the efficiency of neural networks, large training data must be built which might make one think if such methods can be used for low-resource languages like Sanskrit which is the focus of this work. This research work deploys a specific neural approach to Sanskrit summarization – fine-tuning monolingual pre-trained Transformer-based language models. The next section surveys research in neural TS and later argues why the method of monolingual pre-trained language models (PTLMs) and finetuning is suitable for this research.

## 2.5 Neural Methods

Neural methods are the currency in NLP. SL methods have enhanced the outcome of many NLP tasks like question-answering tasks (Van et al., 2021, p. 2116) and other DL-based tasks in general (Shorten et al., 2021, p. 1). SL also has a variant in weak or distant supervised learning which involves a semi-automatic process of labeling unlabeled data through additional sources like databases and dictionaries (Hedderich et al., 2020, p. 2548). Most of these SL and USL methods are used widely in natural language classification tasks (Goldberg, 2017). However, they are definitely not limited to only classification tasks anymore (Chowdhury et al., 2021; Clark & Lappin, 2010; Tampe et al., 2021). A complete neural TS pipeline includes various techniques of finding saliency, techniques for generating relevant words such as copy mechanism, coverage and attention methods (Hou et al., 2021).

A form of training used for learning different parameters of a TS system is multi-task learning (MTL) which helps the model learn different tasks and related modules simultaneously. A key application of the technique has been in learning nuances of different domains as well as syntax of a languages simultaneously (Lu et al., 2019). Related to that, MTL has also been used in simultaneous learning[6] of ETS and ATS (Deng et al., 2021). On a similar line of work, Chen et al. (2019) train ETS and ATS to improve long-term dependency capturing and saliency capturing through MTL.

RL has been used in Chen and Bansal (2018) to combine extractive and abstractive summarization. Kryściński et al. (2018) used RL to improve abstraction in summaries in an encoder-decoder model where the decoder was trained on an LM of summarization data. Their model compete with copy mechanism and PG network of See et al. (2017). A neural attentive encoder-decoder model has been found useful for English ATS (Krantz & Kalita, 2018).

---

[6] See Section 2.5.4 for survey of hybrid approaches combining ATS and ETS.
.

Some other prominent methods have also enhanced solutions to summarization problems through other forms of ML. For example, USL and semi-supervised learning methods have led to word embeddings. With Word2Vec, ELMo, and then FastText (character-based), contextualized word embeddings greatly improved machine's understanding of language. Contextualized word embeddings, large language models (LLMs), etc. are being developed for many ILs. Relatedly, the pre-training of Transformer-based LMs and their subsequent use in a downstream task have become SOTA for current NLP space (Devlin et al., 2019).

Language generation is a crucial part of TS and some variants of on the seq2seq models have become popular. For example, deep generative models like the variational autoencoders (VAE) and generative adversarial network (GAN) have been used in text generation for TS and otherwise in NLP. A graph-to-sequence model (graph2seq) is an interesting variant although it may not yet be a good performer (Jin et al., 2020). Such methods may help enhance generation outcomes in TS in the future. A complete evaluation of other generation methods is out of the scope of this survey (Jin et al., 2020; Mishra & Gayen, 2018).

This thesis is based on pretraining (a language model) and finetuning it on a downstream task (through Transformer-based LMs). Given this context, two questions need to be answered. First, can one use vanilla Transformers or does one need to modify the architecture in summarization tasks? Second, can Transformers be used for low-resource TS?

The next section ponders over these questions by tracing the development of Transformers and their use for summarization from the literature and then establishes that use Transformers for summarization through language model pre-training has been a successful method.

### 2.5.1 Pre-Transformer Era

Before the advent of Transformer models, RNN- and LSTM-based seq2seq models were used in sequential processing tasks like translation (Sutskever et al., 2014). Some popular and successful implementations include the sentence paraphrase task using attention mechanism which was carried out on the Gigaword dataset (Rush et al., 2015). SR was enabled through added resources like the WordNet (Kouris et al., 2021). With Nallapati et al. (2016), the attention-based sequence model was further enhanced for ATS.

Like English and other high resource languages (HRLs), ILTS has also witnessed an impetus since the advent of neural networks. Punjabi neural TS by Jain et al. (2021) achieved 90% recall with neural methods. Nambiar et al. (2021) developed an ATS for Malayalam using seq2seq models. Rani et al. (2021) used LSTM-based model for summarizing Telugu news articles. Similarly, (Mohan Bharath et al., 2022) use neural models for Telugu summarization. Nepali has witnessed TS through LSTM in ETS (Khanal et al., 2021). Kannada is one of the few ILs in which both traditional and neural methods that can potentially be used for ATS have been actively developed (Ebadulla et al., 2021; Embar et al., 2013).

Bengali, Malayalam, and Telugu have used neural methods for TS relatively more intensively than other ILs. Sultana et al. (2022) deployed attention model for Bengali ATS while an RNN-based encoder-decoder for Bengali ATS has been used in M. Talukder et al. (2019); M. A. I. Talukder et al. (2019). None of these works report ROUGE scores. However, Malayalam seq2seq models for sentence summarization task attain ROUGE-1, ROUGE-2 and ROUGE-L scores of 24.33, 23.11, and 24.02 (in percentage), respectively (Nambiar et al., 2021).

### 2.5.2 Transformer-based LM and TS

The advent of Transformers led to large pre-trained language models (PTLMs) like BERT (Devlin et al., 2019) which could be adapted to downstream sequence processing tasks which is a form of Transfer Learning (TRL) (Tunstall et al., 2022). Transformers reused attention and encoder-decoder architectures and became the SOTA architectures for sequence processing. They can be called the culmination point of many existing methods and techniques. LMs help algorithms in understanding word distributions. Transformer-based LMs have been useful in acquiring LM for language understanding tasks like classification - thus, a key area of implementation of these architectures is LM training. Transformers are trained on very large corpora to learn the word distribution thereby acquiring the language model. BERT, RoBERTa are some names of some popularly used Transformer-based LMs.

PTLMs enabled efficient NLP because a lot of time could be saved in developing an application since no time was required to train LMs from scratch (Rothe et al., 2020). Adapting their methodology, different combinations of Transformer-based LMs in the encoder-decoder are the second part of the work that this thesis does.

**2.5.3 Different LM training methods**

Transformers have been used in all use-cases where RNNs or LSTMs were previously used including in TRL-based approached. As a result of transfer learning with Transformers, we see Transformer-based LMs and their subsequent implementation on downstream tasks. The key LMs used in this work are described as follows:

*2.5.3.1 Bidirectional Encoder Representations from Transformers (BERT)*

BERT refers to Bidirectional Encoder Representations from Transformers introduced by (Devlin et al., 2019). Earlier models like the RNNs and LSTMs did not suffice for handling long texts. Transformers were expected to manage those long-term dependencies and the derivatives of the Transformers, the LMs like BERT, began begin used for managing those dependencies. BERT revolutionized the field. However, as literature demonstrates, BERT has not been absolutely successful in managing such long-range dependencies (J. Xu et al., 2020). While it may have advanced way more than its predecessor architectures, BERT has been inadequate in handling instances of very long texts and in tasks involving the connections of such texts, say in discourse-extraction (J. Xu et al., 2020).

BERT as a language model is a way to study the relationship and probabilities among different units/words of a text. As a result, two key training objectives are a part of BERT training – Masked Language Modeling (MLM) and Next Sentence Prediction (NSP).

*2.5.3.2 Generative Pre-Training (GPT)*

Generative Pretraining (GPT) introduced by a group of researchers at OpenAI, in Radford et al. (2018), is an autoregressive Transformer-based LM which is used for text generation on a given prompt. GPT uses only the decoder portion of the Transformer architecture and has been trained on large-scale corpus of nearly 70,000 books to learn world knowledge resulting in good performance on downstream discriminative tasks like question answering. The authors suggest that unsupervised method enables the model to learn world knowledge well. GPT-2 is a larger version of with 1.5-billion parameters (Radford et al., 2019). GPT-3 has 175 billion parameters. Texts generated by GPT models are found to be very human-like suggesting high quality text generation capability.[7,8]

### 2.5.3.3 *Robustly Optimized BERT Pretraining Approach - RoBERTa*

Introduced by Y. Liu et al. (2019), RoBERTa was meant to improve the performance of BERT on a smaller data size and fewer parameters. The authors had also removed NSP as a pre-training objective. As per the developers, Roberta achieves SOTA performance on SQuAD benchmark (Section 7). The key argument of the authors is that BERT system can be improved by increasing data size, batch size, duration of training, sequence length; modifying the pattern of masking, and leaving out next sentence prediction task (Section 7). The authors of RoBERTa argue that change in hyperparameters impacts the quality of training. RoBERTa showed improved the downstream task performance in language understanding as compared with BERT (Y. Liu et al., 2019). Use of RoBERTa for sequence generation task has resulted in mixed results and RoBERTaSHARE worked good on sentence summarization task while BERT setups succeeded in summarizing documents (Rothe et al., 2020). However, those research works focus on the HRL English data. In the LRL sphere, what must be paid attention to is the poor performance of RoBERTa for morphologically rich ILs like Hindi and Telugu which has resulted in RoBERTa being rejected for these languages (Jain et al., 2020).

---

[7] https://www.scientificamerican.com/article/we-asked-gpt-3-to-write-an-academic-paper-about-itself-mdash-then-we-tried-to-get-it-published/ accessed December 26, 2022 at 16.41 hours.
[8] https://openai.com/blog/gpt-3-apps/ accessed December 26, 2022 at 17.17 hours.

In this research work, the pre-training objectives of both BERT and RoBERTa have been limited to masked language modeling (MLM) which is the objective of predicting a masked token in a sentence.

*2.5.3.4 Other Models*

Multilingual and smaller-size variants of the aforementioned models have also been developed. BERT has mBERT and many other LMs have multilingual versions. DistilBERT is an additional model which is trained on fewer parameters than RoBERTa thereby reducing the memory and resource constraints on training (Jain et al., 2020). ALBERT reduces the BERT parameter size so that the memory requirements of BERT may be reduced for improved training (Lan et al., 2019). Similar models include a cross-lingual XLM-RoBERTa (Conneau et al., 2020), model that has been trained only on English without any multilingual versions (Jain et al., 2020), autoencoder-based BART (Lewis et al., 2019), and T5 (Raffel et al., 2020). Likewise, other model variants exist a complete review of which is out of the scope of the present work.

The literature in Transformer-based LM for TS has advanced significantly in the last 5 years. The problems which persisted in the pre-transformer era, were solved through Transformers using the same techniques. The difference, however, was in using Transformer architecture in place of RNN or the LSTM architectures. For example, just as MTL has been used to learn many tasks through LSTMs, it has been used for learning "summary-worthy named entity" through a BART model head for classification (Nan et al., 2021, p. 2729), and saliency capturing (W. Xu et al., 2020) among other tasks. MTL's power of simultaneous learning has therefore enabled diverse training with Transformers as well.

LMs are not limited to tasks which require deep linguistic knowledge and language generation, such as ATS where machine must understand the source text before summarizing, pre-trained LMs have been fine-tuned on a specific abstraction task (Liu & Lapata, 2019). Contrary to expectations, LMs have also been used in ETS which has relatively lesser need of understanding a text. J. Xu et al. (2020) enable learning of language nuances through DISCOBERT so that discourses may be extracted from the source. They argue that DISCOBERT can handle long-range dependencies that are important in ETS for sentence extraction and also prevent redundant information from flowing into the summary (p. 5021).

Models like the DISCOBERT are apt use-cases of Transformer-based LMs being used for traditional information extraction methods like discourse unit-extraction for summarization. Use of these LMs with Disc-Ana methods have been used to enhance the source text analysis by the model and then produce summaries. DISCOBERRT extracts discourse units from news text and then produces the summaries. Similarly, Cohan et al. (2018) proposed a Transformer-model based ATS that considered discourse units in summarization but it only uses the decoder portion of the Transformer architecture to analyze discourse units.

Maimaiti et al. (2019) suggest Multi Round Transfer Learning for LRs using High Resource Languages as the parent model for the LRL child model to incorporate through multi-round transfer. By using the baseline of Vaswani et al (2017) they report that with Multi Round Transfer, BLEU shoots to +5 points. They suggest improving the processing of LRLs by utilizing HRLs through transfer learning. An idea closely related to TRL is Distillation Learning (Dis-L) which deals with a basic question - How to distill information contained in a larger model and use it on a smaller model? Dis-L has been used by (Turc et al., 2019) to train teacher models and train student models on them.

LM-pretraining and fine-tuning, though, has been a popular and successful method for TS. Fine-tuning of a pretrained BERT-based model for fine-tuning on a TS task was initially devised in two works. First, pretraining of BERT-based encoders for use in downstream task of summarization was implemented by Liu and Lapata (2019). Their work indicated that ATS can be approached through encoding sentences. While the encoder is pretrained, the decoder is trained from scratch and both are trained with different optimizers and learning rates. Liu and Lapata (2019) indicate that pre-trained encoders may not succeed in generating many new n-grams in the abstractive summarization task – that is, the model is a poor generator on the CNN/DM dataset, but not for XSum dataset. In the combined abstractive-extractive summarization setup, extraction is higher (that is it tends to extract sentences more than it abstracts them). The pretraining approach of Liu and Lapata (2019) was devised with the goal of keeping a simple model which despite not requiring the use of copy mechanism or reinforcement learning to add to the model performance, achieves good results (p. 3731). Second, Zhang et al. (2019) used the same technique on CNN/DM and NewYork datasets to with promising results.

Such approaches were then adapted to different domains later. Ma et al. (2021) fine-tuned BERT for producing topic-aware TS. Similarly, medical articles summarization through BERT and GPT-2 has been approached in (Kieuvongngam et al., 2020). BERT for geosciences domain has been approached in (Ma et al., 2022). Notably, BERT fine-tuning for ETS has challenges (Miller, 2019). Therefore, Transformer-based architecture/LM and its fine-tuning has served as a promising method for ATS.

### 2.5.4 Hybrid Approaches

ETS and ATS have, at times, been treated not as separate but complementary tasks. That is, for a good TS system, important sentence(s) from the source text are extracted through ETS and are then summarized in an abstractive manner (Chen & Bansal, 2018). A line of research in suggests that many primitive TS systems claiming to be abstractive are actually extractive only which indicates the fact that researchers have long been wanting to generate abstracts, also called condensed texts. However, they have used methods that are extractive at their core (Kasper, 1999). This pattern seems to have been followed later as well but not because resources were limited but because despite limited resources, the best method seems to have been achieved through a combination of both.

#### *2.5.4.1 Sentence Compression as an ATS-ETS link*

Sentence compression (SC) is related to ATS. The goal in this task is to reduce a sentence to its shorter form without losing any significant information or order of subject matter (Cohn & Lapata, 2013, p. 3). ATS has often been used in combination with SC to improve abstraction. Liu and Liu (2009) suggest using SC as a method for improving extractive summaries and bring them closer to abstractive summaries (p. 261). Thus, SC has been used as a way to improve ATS. van Yperen et al. (2021) rightly notes that the earlier works on using SC for TS like (Cohn & Lapata, 2008) were more extractive than abstractive. However, as Liu and Liu (2009) observe, SC does not work well without a well performing generation module (p. 264).

On a similar note, Kryściński et al. (2018) emphatically note that abstractive summarization pipeline involves extraction of summary portions followed by paraphrasing the extracted portion (p. 1808). Their observation indicates that extraction may never completely discarded from the ATS pipeline. Additionally, some also see the combination of ETS with ATS as the dependence of ATS on ETS to resolve factual errors and to also use SC for the resolution (Xu & Durrett, 2019, p. 3292).

Similarly, J. Xu et al. (2020) take cue from Liu and Lapata (2019) to treat ETS of discourse units as a sequence labeling task where discourse units (the sequences) of a text are ranked or labeled by a neural network and only some units are extracted at the end (p. 5024). Similarly, Chen and Bansal (2018)'s method first extracts salient sentences and then rewrites them to produce abstractive summaries, whereas Droog-Hayes (2019) uses Disc-Ana to first extract the important sentences in the text and then summarize them in an abstractive manner. A detailed survey of methods which combine extractive and abstractive summarization techniques can be found in Hou et al. (2021, p. 641).

The key point emerging out of all the above works is that ATS systems may not be able to capture saliency on their own. Especially in the case of long document ATS, summarization even through Transformer-based LMs is ineffective at capturing saliency as well (Pilault et al., 2020). Most approaches, then, first use ETS to find important parts before summarizing them resulting in hybrid approaches (Chen & Bansal, 2018; Kryściński, Keskar, et al., 2019). This forms a part of some open research problems in the field and may be cited as drawbacks of the data-driven models. A probable solution for resolving the shortcomings of the neural or data-driven models has been proposed in the literature, as the next section explains.

### *2.5.4.2 Clubbing Deep Learning and Traditional Methods*

Data-driven DL methods train on large-scale datasets to learn linguistic and world knowledge (Han et al., 2021). Yet, as noted above, the present-day results indicate that it may not be totally true. DL seq2seq models are found to be severely limited in many aspects like capturing saliency or generating keywords (Li et al., 2018; W. Xu et al., 2020). As a result, new methods are deployed to incorporate the traditional methods of using linguistic resources in the DL pipeline to enhance the output such as combining WordNet (Fellbaum, 1998), Word Sense Disambiguation (WSD) and generalization in the summarization process (Kouris et al., 2019, 2021). Hence, while using DL methods are expected to learn deep representations of text to capture all linguistic and world knowledge, linguistic resources may still be required in NLP as much as in TS.

PTLMs, however, seem to capture linguistic knowledge better than any other methods (Han et al., 2021; N. F. Liu et al., 2019). Their power is attested to be more than the earlier DL methods thereby making them current SOTA techniques.

ATS through the aforementioned methods has been successful in language like English which have a strict word order. The morphologically rich languages like Sanskrit, however, pose different challenges. The next section explains the case of such languages.

## 2.6  The Case of Morphologically Rich Languages

While training LMs for NLP tasks has been a popular idea, literature has cited concerns about training LMs for ILs, among other similar languages, given their morphologically rich nature. First, the concerns were cited earlier for LSTM-based LMs for inflectional languages such as LSTM-based LMs in Kannada (Ebadulla et al., 2021). (Cotterell et al., 2018) note that LSTM-based LMs do not suit inflectional languages. As a result, a better way for modeling inflectional languages was needed. Additionally, tokenization in morphologically rich languages may lead to large vocabulary sizes (Baykara & Güngör, 2022). Tokenization, which is an important step in any NLP preprocessing pipeline, has traditionally been on the word-level.

Handling rare or out-of-vocabulary (OOV) words is an open research question in TS as well as NLP with some very promising solutions having been devised so far. As noted by Wu et al. (2016) - the developers of WordPiece tokenization for Google's neural machine translation (NMT) system - rare words are handled by two popular mechanisms, among others – 1) copy mechanism, and 2) sub-word tokenization (p. 7).

The first method was introduced through pointer networks (Vinyals et al., 2015). When a rare or OOV word is seen in the input, the pointer network points to the word to use it as it is (Lin & Ng, 2019). The network is said to copy the text when the network captures an entire unit of input instead of a word (p. 9819). Relatedly, pointer-generator with coverage mechanism was adapted to ATS task in (See et al., 2017). The coverage mechanism was used to monitor novelty by ensuring that a previously summarized portion was not summarized again. Although copying enhanced the capability of networks to manage OOV words, it was critiqued to be defeating the very purpose of abstraction and novelty in ATS making them closer to ETS (Lin & Ng, 2019). Such methods have been widely adapted in bottom up ATS (Gehrmann et al., 2018) and in handling OOV words as well as concepts for novelty of concepts in summaries (Wang et al., 2019).

The second method is of sub-word tokenization which is at the core of Transformer-based LMs and those subword tokenization methods do away with the need for copy mechanisms (Wu et al., 2016). Subword tokenization considers parts of a word as the tokenizing unit making the vocabulary size manageable. Given a vocabulary of n words during training, if a new word is encountered in the test or prediction corpus, the model will call it an OOV word failing to handle it (since it was not a part of the training). The probability of finding an OOV word becomes even higher in inflectional languages where inflected forms of the same root may be treated as two words instead of being treated as semantically related to the root. Thus, rich morphology of a language may give rise to new OOV words thereby posing a problem at the text generation stage (Baykara & Güngör, 2022). Therefore, morphologically rich languages seek tokenization at levels different from that of English.

### 2.6.1  Tokenization on different levels

Models like BERT enhance vocabulary learning of models by incorporating different tokenization schemes. The WordPiece (Wu et al., 2016) tokenization scheme used in BERT, DistilBERT models is a point of balance between character- and word-level tokenization since it tokenizes on the subword level to manage OOV words or rare words. It is modeled on the subword tokenization scheme which has proven an effective tokenization scheme for speech recognition tasks in morphologically rich and agglutinative languages like Korean and Japanese (Schuster & Nakajima, 2012). Relatedly, Byte-Pair Encoding (BPE) tokenization, introduced by Sennrich et al. (2016), helps manage large vocabulary in tokenization and is very similar to WordPiece.[9] BERT uses WordPiece Tokenization while GPT-2 and RoBERTa use

---

[9] https://huggingface.co/docs/transformers/tokenizer_summary#wordpiece

BPE. A detailed analysis of tokenization suitability is out of the scope of this thesis. An excellent source for comparative study on tokenization schemes of BERT and its impact on vocabulary size can be found in Nemeskey (2020).

Research has also indicated that during the LM pretraining phase, the subword or character-level tokenization may be very useful for morphologically rich languages and also that techniques like WordPiece are better than BPE (Baykara & Güngör, 2022; Bostrom & Durrett, 2020) Thus, WordPiece tokenization seems more promising for Sanskrit. An example of subword tokenization is presented below: During the LM training conducted for this work, BERT split words into subword units:

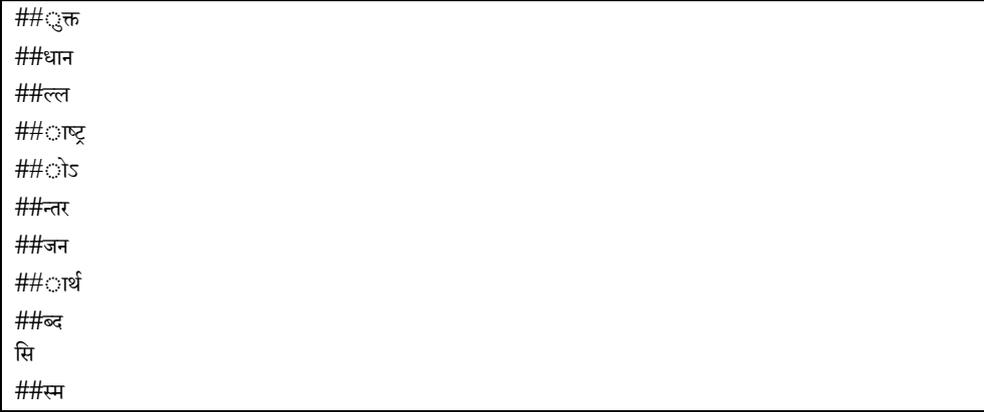

*Figure 1: Tokenization of Sanskrit data in LM pretraining*

These subword units are then used to identify words making the vocabulary much easier to handle. It then naturally follows that when Liu and Lapata (2019) suggest that PTLMs for summarization is a "minimum requirements" model which does not require copy mechanisms for handling these issues, they depend on the power of these tokenization systems (p. 3735). Thus, PTLMs for ATS are effective even without any added methods or resources.

The impact of BERT-based LMs on languages other than English is an open research question. Martin et al. (2020) observe that large scale LMs like BERT and RoBERTa have not been trained on any other language in the same quantity as English and consequently the impact of BERT-like LMs for morphologically rich languages has not been not adequately evaluated owing to the large training and resource requirements of training such models. They also use OSCAR corpus for training CamemBERT, an LM for French.[10] Baykara and Güngör (2022) make similar argument in the light of Turkish and Hungarian.

Thus, research on the impact and usability of Transformer-based models for ATS especially for morphologically rich languages is highly needed. Some related efforts in certain morphologically rich languages have indicated promising results. For Turkish, MLLM BERT was fine-tuned to give promising ROUGE scores indicating that MLLM might also have a promising role to play with good-sized datasets (Baykara & Güngör, 2022). Alternately, Japanese ATS has also been approached through BERT-based encoder although the summaries suffer from redundancy (Iwasaki et al., 2019).

In this research work, the use of PTLMs is aimed at summarizing texts because of the aforementioned issues which are also relevant to Sanskrit. However, this work presents the challenges of deploying PTLMs for ATS in a specific domain of contemporary Sanskrit prose because of the different characteristics of this domain as the next section explains.

---

[10] This research has used the Sanskrit portion of the OSCAR corpus for training the models. Chapter 3 elaborates more on the data collection methods.

**2.6.2 The Case of Contemporary Sanskrit Prose**

SATS is an unexplored field. With or without PTLMs, SATS has not invited the interest of many. So far, the only attempt at Sanskrit TS has been by Barve et al. (2016) which was a query-based ETS developed through TF-IDF, and vector-space model. That is the first work in Sanskrit TS and the only work so far. The authors collected Wikipedia data and used user-given query as the cue to selecting important sentences. One can observe that the choice for Wikipedia, an easily available data, is very natural to a language that has not witnessed any research in the direction. Wikipedia data does not contain long literary constructions making it closer to the HRL datasets currently available. Additionally, as Hellwig (2015a) notes, processing classical Sanskrit literature has challenges of orthography, metrics handling, prosody, lexicon, etc. in addition to the issues of resolving *Sandhi*s. Thus, despite its vast literature, Sanskrit resources that can be effectively used for NLP model training are low.

Taking cue from the earlier Sanskrit works in NLP, this thesis also focuses on Sanskrit data that does not contain long complex sentences and would not pose many challenges in terms of orthography and complex word order – the contemporary Sanskrit prose comprising of web crawled data, Sanskrit speeches, and journal articles. This thesis uses contemporary prose to initiate Sanskrit ATS hoping that ATS will be implemented for different genre of Sanskrit literature in the future.

The non-literary contemporary prose is also a feasible choice for summarization because summarizing plots of literature books, or literary texts in general, is more difficult than summarizing short documents although objective summaries can be generated through extraction (Ceylan & Mihalcea, 2009, p. 592). Kakwani et al. (2020), who developed large-scale monolingual corpora for different ILs, have also focused on corpora containing instances of contemporary language possibly due to the short sentences and ease of availability of such data.

LMs for specific NLP tasks seeks availability of large language resources or datasets. As a result, such methods have often been performed on high resource languages (HRLs). The success that deep learning models have exhibited on HRLs languages makes them a lucrative approach to NLP tasks most of which are sequence-to-sequence processing tasks. This thesis looks at the challenges of applying neural methods to an LRL, Sanskrit. For a language like Sanskrit, the size of training data is often distinctively lower than that in HRLs (Krishna et al., 2021). The crux of this thesis, then, is - given the LRL status of Sanskrit, would it be possible to still initiate an SATS system?

Having surveyed the PTLMs approaches in HRLs along with the advantages of using Transformer-based LMs for morphologically rich language like Sanskrit, in the next section, I enquire whether despite the high advantages of Transformer-based PTLM for downstream task of ATS they can be used for Sanskrit which is an LRL. In other words, the next section seeks to know if literature indicates that PTLMs can be used for the downstream task of ATS for Sanskrit (or similar languages) even if Sanskrit is resource-scarce.[11]

---

[11] I have used 'resource scarce' synonymously with 'low resource.' Some scholars prefer to separate the two.

### 2.6.3 Low-Resource Challenge

In the case of ILs, the low-resource challenge is a major roadblock to using Transformer-based language models which seek large amounts of data for training and are immensely useful in downstream tasks (Kakwani et al., 2020). Despite the LR nature of these languages, efforts are on a rise to develop more monolingual resources and evaluation benchmarks (Kakwani et al., 2020; Kunchukuttan et al., 2020; Kunchukuttan et al., 2017).

Scholars have noted the difficulties in working in LR TS. As noted in Section 2.3, either the domain of the source text is LR (Bajaj et al., 2021; D'Silva & Sharma, 2021) or the language of the source text is LR (Sinha & Jha, 2022). The former is domain-based data scarcity whereas the latter is language-based data scarcity. One way of mitigating low-resource challenges is data augmentation. Literature survey reveals that data augmentation has been one popular way of creating more data to train neural models (Parida & Motlicek, 2019).

Second approach is called Transfer Learning (TRL). The Transformer-based PTLMs are used in downstream tasks through TRL (Martin et al., 2020) Originally used in Computer Vision (CV), TRL is the process of training a model on a large dataset and transferring the learned parameters to a language or domain with low data (Tunstall et al., 2022). The specific use case of TRL in language model (LM) pre-training and fine-tuning has proven to be a great advancement in NLP including TS. An example of TRL is, zero-shot cross-lingual summarization helps in using models trained in high resource settings to be adapted to LR settings without the algorithm seeing even a single example of the LRL (Khemchandani et al., 2021), (P. J. Liu et al., 2019). Cross-lingual summarization - where summarization model trained in a high-resource language is transferred to a low-resource language – has also been attempted through unified decoders (Bai et al., 2021). Earlier, these LMs were limited to NLU but they have now been used to generate language units too. Hence, LMs can assist in not only understanding the language distribution but also generating units from it (Rothe et al., 2020). Transformers offer effective sequential processing capabilities to neural networks thereby becoming a top choice for most sequential processing tasks.

In this thesis, large-scale PTLMs have been used for fine-tuning on downstream tasks of TS and they have been used to develop SM for Sanskrit given that they have proved to be effective in LR scenarios.

The requirement for Transformer-based PTLM training also is to have large labeled datasets. The most pressing problem in LM training is lack of data which also one of the primary concerns of this work. ILs, which are LRLs, lack large-scale corpora typically needed for development of these LMs (Kakwani et al., 2020). Thus, scarcity of resources may exist for either the monolingual corpus for LM training or the downstream task (say, lack of parallel translation data), or both. This research work faces but also resolves resource crunch on both fronts.

Some other limitations of PTLMs have been regularly pointed in literature. J. Xu et al. (2020) argue that pre-trained LMs can handle only a limited size of text and are not suited to tasks that require understanding of long documents (p. 5022). Similarly, as noted earlier, in Section 2.5.4, incorporating knowledge resources to enhance model learning has been a popular topic in DL (Goldberg, 2017) and that PTLMs are commonly considered good acquirers of knowledge. However, a different line of debate in PTLMs suggests that incorporating knowledge-based elements in these PTLMs may enhance their performance on downstream tasks (Wang et al., 2022). OOV words is also the problem that has led to DL models being combined with linguistic resources to learn the best way to link an OOV word with the best possible general term through knowledge based like WordNet (Kouris et al., 2021). As a result, Knowledge-enabled BERT (Liu et al., 2020) and similar PTLMs for downstream tasks have also been proposed. Evaluating them is out of the scope of this work.

While I was finalizing this research work, a new ATS model enhanced specifically for low-resource summarization, Z-Code++, was released which claims to improve TS (He et al., 2022). I do not test that model in this work. Additionally, a decoder-only model treating summarization as an LM task is proposed by (Khandelwal et al., 2019). However, due to technical issues, I am unable to attempt a decoder-only model for ATS either.

Since Sanskrit is also an IL, the best possible sources were used to develop data. Like Kakwani et al. (2020), this thesis also crawls Wikipedia data and augments it through OSCAR to prepare data, as the next chapter explains (Suárez et al., 2019).

### 2.6.3.1  *How low is low-resource?*

An important question to ask in dataset preparation method is – what number of data points constitute low-resource exactly? There is no consensus on the lower limit to the number that can be called low-resource. Some works like Bajaj et al. (2021) use 120 long document-summary pairs thereby becoming one of the lowest sized datasets in literature. Yet another line of literature indicates that even a few million data points would be low-resource - LR settings, especially cases with a "few hundred labeled instances" have been considered inadequate for neural methods. (Hedderich et al., 2020, p. 2547). As noted earlier, Sanskrit datasets are not able to match the size of regularly used datasets (Krishna et al., 2021). This thesis makes the best use of available Sanskrit resources to develop the best possible number of sentences (for LM) and document-summary pairs (for SM). While use of LRLs in neural methods is indeed tough, that is where the crux of the challenge lies. In other words, while LR is a challenge in general for NLP, one may still want to develop a system for LRLs using neural methods since neural methods have greatly improved the NLP performance (Devlin et al., 2019; Rush et al., 2015). Therefore, despite the LRL nature of Sanskrit, summarization is being attempted through neural methods here.

The low-resource challenge posed by Sanskrit or ILs is shared by many other languages of the world. While this thesis developed datasets with the best possible sources, the data numbers may not match the current SOTA methods.

Thus, the core question that this thesis deals with is how can one use neural methods in the LRL setting of Sanskrit? Literature has dealt with this question for LRLs in general suggests that TRL especially usable in downstream tasks with low-resource settings (Tunstall et al., 2022). A famous use-case of TRL is in cross-lingual summarization (Bai et al., 2021; Žagar & Robnik-Šikonja, 2021). Transformers further enhanced the TRL mechanism leading to easier processing of languages with little data – that is, Pre-training network on a task and then fine-tuning it on another task improves the task performance greatly especially for tasks for which supervised labeled data is negligible (Tunstall et al., 2022). TRL is a potentially good method for NLP downstream tasks when data is scarce (Nikolov, 2020). In other words, literature recommends a simple solution to the problem of using NNs for LRLs – pre-training LM and fine-tuning on a target task. This thesis uses this solution of pre-training and fine-tuning - in the particular LR setting of Sanskrit. That is where the core contribution of this work lies.

BERT-like large LMs for ILs have been trained such as MURIL (Khanuja et al., 2021), ALBERT-based MLLM for major ILs (Kakwani et al., 2020), TransformerXL in iNLTK (Arora, 2020), BERT (Kumar et al., 2020), IndicTransformers (Jain et al., 2020), a "ROBERTA-like model" (ROB-like, hence) for Hindi, Gujarati, and Sanskrit hosted on huggingface[12], GPT-2 for Sanskrit literary documents.[13]

The models above have four characteristics which deserve attention:

- First, except GPT-2, most of the trained LMs are multilingual. Multilingual LMs have been found to be inferior to monolingual models (Rust et al., 2021).
- Second, none of them has been evaluated on a downstream task of TS.
- Third, barring iNLTK, ROB-like, GPT-2 for Sanskrit and MURIL, models rarely are developed for Sanskrit.[14]
- Fourth, models are highly varied making it tough to finalize which models are suitable to the morphologically rich nature of ILs. Jain et al. (2020) observe that mBERT does not obtain sufficient performance for Hindi, Bangla, and Telugu for it is trained on very low data. They argue that models should be selected based on the task given at hand. On the other hand, Kakwani et al. (2020) use ALBERT for training an MLLM in 12 ILs, called IndicBERT, citing its ease of use and small size and thus, model selection parameters are largely different. Thus, Transformer-based IL models seek more research especially for sequence generation tasks. However, present research suggests that fine-tuning from pre-trained Transformer-based LMs – monolingual or multilingual LMs - is universally recognized as a popular method to mitigate the low-resource problem (Ogueji et al., 2021; Tunstall et al., 2022). In other words, PTLMs have proven to be extremely effective in performing downstream tasks even in low-resource settings.

---

[12] https://huggingface.co/surajp/RoBERTa-hindi-guj-san
[13] https://github.com/Vaibhav2001/VedaLearn/blob/main/README.md
[14] See the next Section for why MURIL and other available models were not used in this work.

Therefore, I have trained three Transformer-based LMs in Sanskrit before fine-tuning them on a summarization dataset.

### 2.6.4 LMs in Sanskrit

One possible objection to training LMs could be this - In the presence of a multilingual pre-trained LM, MURIL which has been trained on Sanskrit too, why does this thesis train monolingual LM from scratch?

As noted earlier, a simple answer to that objection is that the TS literature indicates that there are two major disadvantages of using MLLM for downstream tasks like TS – they fail to perform as good as monolingual LMs (mono-LMs) (Martin et al., 2020). The difference in their tokenization abilities of mono- and MLLM may be the reason for such difference in performance (Rust et al., 2021). Mono-PTLMs and their use on downstream task of ATS have been found to be exceptionally good for LRLs (Martin et al., 2020). Second, training MLLMs with low quantity data for a language never leads to good performance. Sanskrit is not a part of mBERT. Hence, I train mono-LMs for this thesis.

I do acknowledge two points here though –

- First, this research is not training mono-LMs for Sanskrit for the first time ever. Mono-LMs, ULMFiT and TransformerXL have been trained for Sanskrit in Arora (2020) and GPT-2 has been trained in Sanskrit through Sanskrit literary documents.[15]
- Second, to counter the low training data problem in mBERT above, a multilingual version of BERT that includes Sanskrit has been trained by Kumar et al. (2020) (MLLM BERT) and Khanuja et al. (2021) (MURIL). The data count of MURIL, after up sampling data, is more than that collected in this work. The MLLM BERT trained by the former also has more data. Literature suggests that if the quantity of pretrained data is adequate, MLLM and Mono-LM have negligible difference (Rust et al., 2021). So, in other words, the shortcomings of mBERT may be said to have already been overcome by the above two models. However, given that a parallel argument of monolingual tokenizer being better than its multilingual counterpart exists, this thesis stays on the line of monolingual LM.

This research presents a set of monolingual Transformer-based LMs – BERT, GPT-2, and RoBERTa owing to their success in English ATS but on a slightly augmented set of sources. The next chapter explains the additional sources used for LM dataset preparation. The mono-

---

[15] https://github.com/Vaibhav2001/VedaLearn/blob/main/README.md

LMs by Arora (2020) and other MLLMs which have been adapted to Sanskrit including mBERT, may be evaluated on ATS in a future work.

Relatedly, I conducted an experiment through using MURIL checkpoints.[16] The systems generated the same output for all input conditions. MURIL, therefore, could not be taken up for further research in ATS downstream task. If more resources are arranged, MURIL and other models may be evaluated. Needless to say, not evaluating the existing LMs for ATS remains a limitation of this thesis and that a comparative study of the downstream performance of multilingual models containing Sanskrit with monolingual models may be considered in a future work.

Therefore, this thesis departs from the other works on two fronts – first, it uses monolingual PTLMs and tests the on the downstream task of ATS. Second, it collects slightly more data from sources in addition to the Wikipedia and OSCAR corpus used by (Khanuja et al., 2021; Kumar et al., 2020). Thus, I have used contemporary Sanskrit prose for TS to test the scope of ATS Transformer-based algorithms in a low-resource setting thereby addressing the aforementioned age-old issues in ATS. The application of PTLMs for Sanskrit ATS is a major contribution of this thesis.

## 2.7 Generation and Human Evaluation

Once a suitable representation of a natural language text has been obtained, processing has been done, the final logical step in NLP is to perform Natural Language Generation (NLG).

---

[16] Results reported in Appendix F

NLG uses different ways to generate text one of which is a decoding method. A good decoding strategy is able to generate the right sequence. Two decoding methods – greedy and beam search – were used to generate the summaries. A detailed analysis or comparison of different decoding strategies is out of the scope of this work but more on those strategies can be found in Nadeem et al. (2020) and DeLucia et al. (2021).

## 2.8   System-Generated Summaries

TS systems are evaluated for their ability to suitably analyze a text. This analysis is important for summaries of good qualities. Therefore, summaries are judged for having certain characteristics while avoiding some others. Section 2.5.4. indicated that ATS systems face issues in capturing saliency. Preferable characteristics include good compression rate, negligible redundancy, and novelty and lack of hallucination as described:

### 2.8.1   Compression Rate

A text summarizer must produce summaries that are concise and shorter in length than the original text. Compression rate (CR) is the ratio of word count in the document to that in summary such that the higher the value, the shorter the summary. Such a reduction will only happen when information in the source document is selectively presented. However, there are different views on how much content from the source text should be reduced (Torres-Moreno, 2014, p. 7). Earlier, Ceylan and Mihalcea (2009) had argued that compression rates in book summaries surpass those in short documents (p. 591), that is, books require bigger compression rates. Therefore, different domains may require different CRs.

Parallel datasets in summarization can offer a great hint at the ideally expected CR. Datasets with data-summary formats indicate that the desired summaries ranges from one-sentence (Narayan et al. (2018a), Rush et al. (2015)) to a few lines (Hermann et al., 2015). As a result, one may assume that both single sentence summaries and multi-line summaries are prevalent. Notably, each of the available above datasets were developed from already available news-summaries sources. Thus, if a newspaper provided articles with summaries, the dataset developers adapted those summaries thereby leading to a certain summary length. On a different note, such available sources make the dataset objective with the least bias coming from the developers. Figures 2 and 3 provide a sample description of the kind of

|   | Source | Summary |
|---|--------|---------|
| 1. | This is a sample source text of many lines. This is a sample source text of many lines. This is a sample source text of many lines. This is a sample source text of many lines. This is a sample source text of many lines. This is a sample source text of many lines. This is a sample source text of many lines. This is a sample source text of many lines. This is a sample source text of many lines. | This is a sample summary of many lines. This is a sample summary of many lines. This is a sample summary of many lines. |
| 2. | This is a sample source text of many lines. This is a sample source text of many lines. This is a sample source text of many lines. This is a sample source text of many lines. This is a sample source text of many lines. This is a sample source text of many lines. This is a sample source text of many lines. This is a sample source text of many lines. This is a sample source text of many lines. | This is a sample summary of many lines. This is a sample summary of many lines. This is a sample summary of many lines. |

*Figure 2: Example of multi-line summary*

|   | Source | Summary |
|---|--------|---------|
| 1. | This is a sample source text of many lines. This is a sample source text of many lines. This is a sample source text of many lines. This is a sample source text of many lines. This is a sample source text of many lines. This is | This is a sample summary of one line. |

| | a sample source text of many lines. This is a sample source text of many lines. This is a sample source text of many lines. | |
|---|---|---|
| 2. | This is a sample source text of many lines. This is a sample source text of many lines. This is a sample source text of many lines. This is a sample source text of many lines. This is a sample source text of many lines. This is a sample source text of many lines. This is a sample source text of many lines. This is a sample source text of many lines. | This is a sample summary of one line. |

*Figure 3: Example of single line summaries*

## 2.8.2 Length, Redundancy, Novelty, OOV, and Hallucination

Neural methods like auto-encoders may be used to manage source summaries of any length (P. J. Liu et al., 2019). However, some do suggest that source or summary lengths may impact summarization process as a result of which some solutions to control the length of the output summary are proposed including a multi-head attention in which both encoder-decoder are sensitive to length (Sarkhel et al., 2020). Summaries may also suffer from redundancy, that is, repetition of the same chunk of information. redundancy in summaries has often been attributed to attention mechanism which may lead systems to focus too much on certain portions as a result of which coverage or distraction mechanisms have been proposed (Chen et al., 2016; Lin & Ng, 2019, p. 9819). Distraction in attention vectors helps the model stay updated with the history of the generated portions so far and in capturing multiple contexts instead of focusing on only some specific parts (Chen et al., 2016; Nema et al., 2017). A related approach is global context encoding for a sentence paraphrasing task to reduce repetitive units and obtain summaries which do not contain irrelevant information from the source sentence (Lin et al., 2018). Therefore, generation of perfect summaries even in languages like English is a far distant thought and the ATS mechanisms constantly require added mechanisms to tackle one issue or the other.

Kryściński et al. (2018) convincingly argue that most summaries cannot create abstracts with new words thereby lacking novelty. To resolve this issue, authors extract important information using RL with encoder-decoder model to improve abstraction in summaries which is not achieved through copy mechanisms.

A related problem with using Transformer-based LMs for summarization is that the model-generated summaries hallucinate, i.e., they generate information which not originally present in the source text, some summaries may be irrelevant to the source text altogether (Pilault et al., 2020, p. 9314). To improve the summary quality by reducing hallucination and generating more relevant content, Pilault et al. (2020) trained hybrid (extractive and abstractive TS) through Transformer-based LMs and indicated that conditioning ATS on ETS (i.e., summarizing extracted sentences) generates high-quality summaries. Their work is highly relevant to this thesis. They indicate that for such hybrid summarization models trained through Transformer-based LMs, the generated summaries are not very relevant to the source document but the summaries are very coherent. Relatedly, summaries may be factually inconsistent with the source text which must be tackled as well (Kryściński, McCann, et al., 2019; Nan et al., 2021)

The issues cited above indicate how the language understanding part of most networks is as important as the coherence of the generation process, i.e., a model may exhibit that it has learnt a language but may not be entirely effective in learning the nuances of language generation.

Lastly, the role of human summarization process is also an important factor in TS training that has been discussed by Torres-Moreno (2014) among others (p. 8). TS is a task under NLP and the basis of NLP is emulating human language processing and language interpretation methods (Bender & Lascarides, 2019). Study of the human summarization process aids in understanding the overall cognitive process of summarization. A detailed study of Sanskrit summarization process is out of the scope of this work. However, in the evaluation section, this research proposes to use human evaluation to better assess the summaries.

## 2.9 Conclusion

Through the literature survey, this chapter has established the grounds for the research presented in this thesis – PTLMs for summarization fine-tuning would suit the specific case of Sanskrit. The data availability had a huge role in guiding the methodology chosen.

Sections 2.1 to 2.5 above indicated that the current SOTA models are based on neural methods, particularly Transformer-based models. They are expected to process source text deeply enough to summarize it. Earlier, different methods for source text analysis included discourse analysis, graph-based analysis, WordNet-based analysis, etc. Each of these analysis has been used through the traditional methods and is now also used in conjunction with the SOTA Transformer-based methods.

Through the sections 2.1, 2.2. and 2.3, I conclude two points. First, the methods of source text analysis are impacted by the domain of the source text. That is, the domain continues to play an important role in the TS process. Given the domain, the source text *may* have a specific format (Sharma et al., 2019). Such formats, especially in the news domain, have often said to impact the summarization process (Kryściński, Keskar, et al., 2019; Liu et al., 2022). Additionally, if the domain of the source text is high resource, say news articles or scientific articles, huge datasets can be built from them for training models. Second, while the methods of source text analysis form a crucial part of the TS development process, many methods continue to include linguistic resources like WordNet, Ontology, POS-tags to annotate the language data before training. As a result, such methods may not be very fruitful for languages with low resources or no annotated data.

Sections 2.8 emphasized that the characteristics of the summaries so produced should be that they adequately contain the important information of the source text, no redundant text, and new words. A combination of these three is rare and thus, it is an open research question. The challenges become even stronger in cases like that of Sanskrit where not enough data is available for training. The remaining sections indicated how low-resource challenge in downstream tasks of summarization and sequence generation is an interesting research problem for the Indic-NLP community. SOTA Transformer models have been used to resolve and improve summary qualities. But when it comes to dealing with low-resource scenarios, a popular way of mitigating the limitations is pre-training BERT- (and related Transformer-based) LMs and fine-tuning on a downstream task which became possible due to Transformer-based LMs like BERT. This thesis builds on such methods. For developing and training these resources, two tools/platforms have been heavily utilized. First, the HuggingFace library. The model training part heavily draws from the HuggingFace library. Second, tesseract and related python-based extraction tools which helped in data extraction.

The next chapter analyzes the data available in contemporary Sanskrit prose and the issues in preparing data for model training.

# Chapter 3: Data Preparation

The previous chapter discussed the range of different TS approach and through that discussion, it established that TRL through LM pretraining and fine-tuning on a summarization task might be the best-suited approach to Sanskrit ATS.

In this chapter, I describe the first contribution of this research, i.e., the data for language modeling and summarization fine-tuning. In this chapter, I report the challenges of developing datasets for language model (LM) and summarization model (SM) training. Although data for training each of the LM and SM have been taken from the same sources, there are differences in the format in which the data is presented to the model. This chapter describes the data preparation methods utilized for LM and SM. In other words, this chapter is related to the second theme of research questions – data domain. It answers RQs [2(a), 2(b),](#) and [2(d)](#).

## 3.1    Dataset preparation background: format

Parallel dataset in the form of text-summary has been developed for summarization training in this work. Dataset format is crucial to developing a summarization model through any method. Most TS datasets in English are automatically collected (Hermann et al., 2015; Narayan et al., 2018a). While these datasets are widely used for evaluating different abstractive as well as extractive summarization models, scholars have observed two major faults in these. First, the fault of layout bias in online sources especially news sources. This layout bias refers to the segregation of certain desirable information in some specific segments of a source (Kryściński, Keskar, et al., 2019). In news articles, the summary-worthy content is a more likely to be found in the beginning of the article. Thus, these articles for summarization have an extractive-like structure (Sharma et al., 2019, p. 2204). Second, preparing summary datasets out of already existing sources (like from news articles as in the aforementioned datasets), leads to immense noise in the data which is usually a hindrance to building a good model (Kryściński, Keskar, et al., 2019).

Given that Sanskrit is an LR, insights into data preparation methods were taken from datasets existing in ILs and in some HRLs. In some cases, non-availability of public datasets made it tough to analyze and compare the proposed dataset with that of others. As a result, in order to compare results of the proposed dataset with others, I have used the results and insights reported by the authors in the different research papers to the best extent (see Section 3.8.2).

This thesis largely builds on two major papers - Liu and Lapata (2019) and Rothe et al. (2020). As a result, I have approached Sanskrit ATS as a task based on pre-training an LM and then fine-tuning it for a summarization task. Thus, three major steps constitute the pipeline – preparing data for both LM and SM training, pre-training LMs, and fine-tuning models on summarization tasks. The pipeline for the said approach for my work commences with this chapter.

Gathering TS data for training DL models is the pipeline's first and most crucial step. As noted in Sinha and Jha (2022), public datasets in Indian Language Text Summarization (ILTS) are rare. Relatedly, finding large-scale and public TS datasets is tough in many languages (Hu et al. (2015, p. 1967)). In English, most TS datasets come from newswires (Koto et al., 2020, p.

598). As I noted in Section 2.6.3 (and the upcoming Section 3.8), lack of huge summarization corpora is also a problem in ILs datasets as a result of which data collection is a major challenge in the development of ILTS models, especially, ATS. The following sections describe the challenges in data preparation and reports statistics of the exploratory data analysis.

## 3.2 Challenges in collecting data

For this work, the goal of data development is two-fold – first, for training LM, and second, for training TS models. As noted previously, news articles remain a popular choice for building TS datasets. Relatedly, pre-trained language models (PTLMs) have become popular in NLP ever since the Transformer-based LM BERT was introduced (Devlin et al., 2019). This thesis is also based on three Transformer-based PTLMs – BERT, GPT2, RoBERTa - as a result of which the process of seeks to suit the data processing requirements. Each of those Transformer-based LMs have different tokenization processes and each LM training requires huge amount of data. In the case of ILs, Wikipedia articles remain a popular choice for training LMs (Arora, 2020) as do other publicly available sources of data like Mann Ki Baat translations and Press Bureau of India (PIB) data (Kakwani et al., 2020; Khanuja et al., 2021; Philip et al., 2021; Siripragada et al., 2020). Wikipedia has been used to train monolingual (Arora, 2020) as well as multilingual LMs (Khanuja et al., 2021).

### 3.2.1 Corpus Sources

The corpus for this work was drawn from four major sources of contemporary Sanskrit prose:

1. **Mann Ki Baat**
2. **OSCAR corpus**

3. **Open Access Sanskrit Journal**
4. **Wikipedia**

### 3.2.2 OSCAR corpus

OSCAR corpus (Suárez et al., 2019) is a filtered part of the Common Crawl (CC) corpus which is crawled from the net. It crawls data for different languages and Sanskrit is also one of them. This data could pertain to websites, blogs, etc. This corpus was made accessible by the corpus developers on request and is not permitted to be released in open access.[17]

### 3.2.3 *Manogatam* (Mann Ki Baat) corpus

*Manogatam* is the translated version of the Indian Prime Minister's weekly address by the name Mann ki Baat (MKB). Veteran Sanskrit journalist Baldevanand Sagar translates and posts these translations on various social media handles. I accessed a specific set of translations from SanskritDocuments.[18]

### 3.2.4 Open Access Sanskrit Journal

Anantaa Journal[19] is an international peer-reviewed journal of Sanskrit which publishes articles reporting Sanskrit-based research in Sanskrit, Hindi, and English. I have reported the cleaning of the ~60 articles collected from Anantaa journal (henceforth, the anantaa corpus) and their cleaning process in Sinha (2022) and the same corpus was taken for this work too. A point to be noted about the anantaa corpus is that not all articles contain summaries. As a result, the corpus was preprocessed in two parts – summary containing parts (an_summ) and non-summary containing parts (anantaa_non).

---

[17] Since this corpus is not permitted to be shared online, I cannot release it. However, as permitted by the owners again the corpus has been used for training and testing models.
[18] https://sanskritdocuments.org/sites/manogatam/
[19] https://www.anantaajournal.com/

### 3.2.5 Wikipedia files

Wikipedia corpus was collected directly from iNLTK[20] and Wikidumps.[21] Since iNLTK was data was collected some years ago, the latest sentences from Wikidumps were also considered. The sentences which common in both the collections were deleted. The final Wikipedia file had 3.273 lakh sentences.

## 3.3 Major data Issues

As can be observed in the previous section, the data used as corpus for this work is collected from a variety of sources. This work uses data sources in addition to the existing Wikipedia and OSCAR corpus for Sanskrit which have regularly been used in LM preparation for Sanskrit LM training (Kakwani et al., 2020). It adds to those corpus by extracting data from two publicly available sources other than Wikipedia and OSCAR, that is, data from Sanskrit journal and transcripts of the Mann ki Baat (MKB). This section and the next analyze some aspects of these new sources including the novelty and limitations they bring to the corpus.

Two characteristics of the data collection process must be noted here. First, digital data is available in different formats each with its own challenges of extraction. The goals of data collection are dependent on the two tasks LM and SM training. So, while data has been collected from the same sources, it has been curated differently for the two tasks thereby making the data scanning and collection processes different too. Sentences from the sources were used for LM training. However, when looking at the sources from SM training perspective, the presence of summary portions was also sought. While all the sources easily offered sentences which could be used for LM training, only the journal contained available abstracts for research articles.

---

[20] https://inltk.readthedocs.io/en/latest/api_docs.html#effect-of-using-transfer-learning-paraphrases-from-inltk
[21] https://dumps.wikimedia.org/

Second, a few sections/articles of the anantaa corpus used literary or technical language as a result of which they posed two major challenges –

a) If the paper used intensively used technical terms from a specific domain or field, it became too domain-centric. Such papers could be termed technical papers, as opposed to general-purpose contemporary prose which this thesis targets.

b) The use of intensive lists-like and compounded words was difficult to manage. An in-depth analysis of each of those words or even de-compounding such words posed a major challenge since de-compounding is not in the scope of this thesis.

Due to both the factors above, such technical sections/papers from the anantaa corpus were excluded.

## 3.4 Challenges in pre-processing data

This section explains data preprocessing challenges.[22] Data cleaning is the first step to preprocessing and for this work, for both the language model as well as summarization model training, data was cleaned manually in the first phase and then automatically later in the second phase. The manual cleaning process involved cleaning *sandhi* and irrelevant characters. In the second stage, non-*devanāgari* letters, extra spaces, and any other symbols were removed. The second stage cleaning is similar to Kumar et al. (2020).

---

[22] Part of the results reported in this section have been published in Sinha, S. (2022). Sanskrit sankshiptikarana pranali hetu lekh-sangraha nirmana prakriya: eka samganakiya bhashaavaigyanika drishtikona. *Vigyan Garima Sindhu*(120), 23-28, Article 3. The work has been cited at the relevant places.

As Sinha (2022) observes, the challenges in extracting *devanāgari* data through Tesseract has been covered in some other works as well - extraction challenges for Bhojpuri data in a Sanskrit-Bhojpuri Machine Translation (Ojha, 2019), while some have reported challenges of extracting text from physical PDFs using tesseract (Aralikatte et al., 2021). Many researchers choose manual methods of *Sandhi* splitting or to not split *Sandhis* (Aralikatte et al., 2021; Soni, 2015).

### 3.4.1   Common Errors

The key errors in all data included tesseract extraction errors. The following is an overview of the errors. The goal behind presenting these errors is to indicate the pattern of extraction errors and the way those errors were resolved.

#### *3.4.1.1   Tesseract Errors*

The preprocessing of the anantaa corpus has been reported in Sinha (2022) containing two major steps. In the first step, the anantaa articles appeared in PDF formats as a result of which the journal articles which were present as PDFs were extracted using Tesseract through Python (Smith, 2007). As aforementioned, the corpus is composed of articles containing article summaries (an_summ) and not containing any summaries (anantaa_non) Tesseract OCR engine was accessed via Python a python code and all the extractions were finally made through it[23]. The following two pairs of error-correction were found in the extraction of the anantaa corpus as noted in Sinha (2022): क्र- क्र, अनुस्वार- high note sign (Sinha, 2022, pp. 24-25). In addition to these errors, some other errors were also discovered in extracting the corpus as follows:[24]

| |
|---|
| अनुस्वार read as विसर्ग |
| व्याप्ति and व्यासि regularly interchanged |
| शृ mistaken for different letters |
| पूर्ण-विराम extracted instead of म |
| श्रृ interchanged with क्ष, क्ल |
| प्राप्रोति extracted instead of प्राप्नोति |
| ग् within a word misread as hyphen |
| दृ read as दु |

*Table: Examples of Errors spotted in the Text after Tesseract Extraction*

(Ojha, 2019) reports similar errors for Bhojpuri data. Tesseract error correction was also impacted by *sandhi* between words which the next section explains. The second step was to preprocess *sandhi* and *saṃyoga* in letters.

### 3.4.2  *Sandhi* Resolution

*Sandhi* is a process of euphonic combination of letters that leads to changes in a word's appearance in a text. Hellwig (2015b) categorizes *Sandhi*s as vocalic and non-vocalic each of which leads to words combining in different ways and leading to seemingly new words. For NLP processing, splitting this *sandhi* is important for tokenization (Hellwig, 2015a). Due to different complexities arising in a text sue to *sandhi*s, *sandhi*-splitting is a major pre-processing challenge in Sanskrit NLP.

---

[23] The initial data collection effort through tesseract GUI was ineffective. Data collected over 2.5 years had to be discarded owing to low quality of the extraction and the requirement of intensive postprocessing therein.
[24] Note that these corrections are from the final extracted files. Some of these may have happened due to source error itself. These errors are only being reported and are not really indicative of Tesseract's capability.

Researchers have been trying to automate *Sandhi* splitting process (Bhardwaj et al., 2018; Dave et al., 2021; Hellwig, 2015b; Patil & Patil, 2018). *Sandhi* and *Samasa* automation is also a practice (Pappu & Sanyal, 2008). With the advent of deep learning methods for sequence classification and generation, *sandhi* splitting has become a popular case study of the application of such sequential processing neural networks. Reddy et al. (2018) approach *sandhi* segmentation as a sequence-to-sequence generation task using encoder-decoder architecture which can be trained relatively more quickly. They provide a short overview of past efforts in *sandhi* segmentation using finite-state transducers, etc. Relatedly, Aralikatte et al. (2018) use an encoder with double decoder to spit *sandhi*. On the other hand, Hellwig (2015b) RNN-based sequential processing methods to resolve *sandhi* while recent developments also include Sanskrit_parser by [https://github.com/kmadathil/sanskrit_parser](https://github.com/kmadathil/sanskrit_parser) (kmadathil) which parses Sanskrit *sandhi*s. Each of these is a good advancement in integrating Python with Sanskrit specific language processing requirements including *sandhi*-splitting. However, as kmadathil observes, the split so performed such in data is not always certain. Additionally, some manual intervention is always sought in such splits including some in compound splitting as well leading to higher accuracy of rule-based approaches to *sandhi* splitting (Dhingra & Joshi, 2022).

In addition to *sandhi*s, *saṃyoga* patterns were also split in (Sinha, 2022). *varna saṃyoga* is a process of letter mergers or combinations commonly found in Sanskrit texts. For example, the consonant and vowel in म् अस्ति could be merged through *saṃyoga* to produce मस्ति (Sinha, 2022). These *saṃyoga* patterns also have the same new word generation impact that *sandhi*s also commonly do, that is, *saṃyoga* combines letters of two separate words leading to the impression that a new word is formed. However, unlike *sandhi*s, the patterns of split in these are always deterministic.

### 3.4.3 Use of Patterns

Sinha (2022) automated the *sandhi* resolution process through Python's search and replace method. A Python script reads a list of *sandhi*-split or *saṃyoga*-split pairs from an excel file and replaces the *sandhi*/*saṃyoga* occurrences with their respective splits in the corpus. However, this automated process for the anantaa corpus contained two stages. In the first stage, word-specific *sandhis* were listed in a file (*sannoyga* was excluded). The Python script replaced certain *sandhified* words with their respective splits only. The steps followed were: collect all words of the file in one column of an excel file, for every *sandhi*-containing word in that column, annotate it with its split by writing its split in the another column, use a Python script split a given *sandhi* with its annotation (p. 25). For example, रामस्यैक्यम् is a unit of *sandhi*fied letters which is replaced with its annotated split रामस्य ऐक्यम् as given in a file. It is a case of word-specific *sandhi* which was replaced with its split. A few additional points must be noted:

- o Barring a few cases, *samasa* was never split.
- o *Sandhi* in technical terms and grammar sutras was never split.

- Some unclear words were also present in the corpus which were replaced with the best possible guess for the word, if possible, or
- if their correction could not be correctly ascertained, they were copied as they were.

In the second part of the preprocessing method in Sinha (2022), some commonly occurring *sandhi* and *saṃyoga* patterns were universally replaced. *Sandhi* and *saṃyoga* processes can lead to some commonly occurring outputs which may be universally replaced, that is, they may be split wherever they are found in a text irrespective of the place or word in the corpus in which they occur. For example, a common *sandhi* pattern is इत्युच्यते annotated with its split इति+उच्यते, and a common *saṃyoga* pattern is मस्ति and it was annotated with its deterministic split म्+अस्ति as aforementioned (Sinha, 2022, p. 26). Therefore, a separate file containing all such universal *sandhi-saṃyoga* patterns along with their respective splits was created for the Python script to read later.

Thus, the cleaned anantaa corpus used two phases of *sandhi* splitting – first, word-specific *sandhi* splitting (using list 1 of word-specific *sandhi*-splits), and second, common patterns-based (using list 2 containing common errors).

For this work, text from the OSCAR and MKB was cleaned in a similar manner. The research articles from anantaa corpus contained literary language and consequently long *sandhified* words which needed correction.[25] As a result, word-specific *sandhi* cleaning became important.

---

[25] This exercise may be expanded on this dataset in later work.

However, for the OSCAR and MKB corpus, I only used the common *sandhi/saṃyoga* patterns for splitting since the occurrence of long *sandhified* words was not found in a preliminary analysis of the OSCAR and MKB texts. A Python script was executed to replace all the possible *sandhis* and *saṃyoga* in the OSCAR and MKB corpora with their annotated splits which it obtained from the second list of common errors. In other words, the script used a regex-based find and replace to correct the words joined by either *sandhi* or *saṃyoga* and the commonly occurring *sandhi* and *saṃyoga* patterns as used in Sinha (2022) were adapted for this work too (Appendix D). Wikipedia data was used as it was and was cleaned only during language model training.

### *3.4.3.1 Issues in splitting common patterns*

Some issues faced in splitting the common patterns across files include the following:

- Common *sandhi* patterns were not universally applicable to all *sandhi* joints – Hellwig (2015b) rightly observes that the split prediction is not deterministic always – that is, for a given *sandhified* word, many splits may be produced, although only one or two of them would usually be correct. For example, रामेश्वरः can be split as राम+ईश्वरः, रामा+ईश्वरः, राम+इश्वरः. However, when joining two letters, say, अ+इ, will always lead to ए. This difference in the two directions to *sandhi* made it tough to spot common patterns for 'search and replace' method of Python approach used in this work. So, words like रामस्तु should have been split as रामः and तु, However, वस्तु could not be resolved in that manner. Errors occurring due to the above conditions were manually corrected.

- Another common pattern of *sandhi*s is the set of words ending in 'ओ'. In a corpus, individual words ending in 'ओ' may be traced to two possible cases. First, they are formed out of visarga *sandhi* where अः+हश्[26] letter leads to change of अः to ओ; second, where the word originally ended in a visarga but the visarga was not extracted or was altogether missing in the original text.

- Other common patterns like श्र, नास्ति must only be split if the pattern occurs at the end of the word (रामश्र) or when the pattern does not occur within another word (नास्तिक should not be split) (Sinha, 2022). The regex conditions (used through Python script as reported in the next part) ensured that such patterns were implemented only where needed.

### 3.4.4 Results

The overall and unique token count change has been indicated in Tables 1 and 2 respectively with * marked data of the preprocessed anantaa corpus count having been reprinted from Sinha (2022, p. 27):

| source | Old total | New total |
|---|---|---|
| **oscar:** | 777334 | 780282 |
| **mkb** | 91929 | 85734 |
| **anantaa_non***[27] | 39310* | 42692* |
| **anantaa summ*** | 8889* | 9846* |

*Table 1: Total word count: Before splitting (old) and after splitting (new)*

---

[26] *haṣ* letters refer to the 3rd, 4th, 5th members of each of the five families of the वर्णमाला

[27] Total word count from the journal (including summary) old was 48199, the new count is 52538. The code for data development copied the summary portion of the journal twice thereby adding additional 9846 tokens to the corpus.

Furthermore, the count of unique words also underwent change:

| source | Old Uniq | New Uniq |
|---|---|---|
| **oscar:** | 206375 | 201155 |
| **mkb** | 27974 | 27602 |
| **anantaa_non*** | 21060* | 20189* |
| **anantaa summ*** | 5390* | 5242* |

*Table 2: Unique token counts: Before splitting (old) and after splitting (new)*

The count of total tokens increased while that of unique tokens decreased.

This section concludes that even though tools for automatic extraction of text data and *sandhi* splitting exist, use of rule-like feature mapping functions of Python is an effective way to extract and process data. Search methods complement the preprocessing methods and hence, should be considered for future works in *Sandhi* resolution.

## 3.5  Preparation for Language Model Training

The cleaning process described above is a general cleaning process the goal of which was to rid the dataset of misread characters and *sandhified* words. However, given the two trainings, language model training and summarization model training, that I am going to report in the next chapters, the collected dataset was also cleaned separately to suit each training cycle. In other words, after the primary cleaning process above, each dataset was further cleaned and framed in a format appropriate for the training in a secondary round of cleaning. Two factors influenced the training-specific cleaning. First, the requirements of the HuggingFace library (Wolf et al., 2019). As indicated earlier, this work used the HuggingFace library extensively. As a result, some format specific requirements of the library had to be taken care of. Second, the difference in the type of dataset required for LM and SM training. The LM training process sought monolingual data of Sanskrit in the form of numerous single sentences while summarization sought labeled data of text-summary pairs.

LMs like BERT require huge quantities of data to be trained on. Both MLLMs and mono-LMs have traditionally used Wikipedia and OSCAR corpus for training the respective LMs including those for Sanskrit (Arora, 2020; Khanuja et al., 2021). Khanuja et al. (2021) present an MLLM - MURIL for ILs including Sanskrit.

After a preliminary round of cleaning, the secondary steps to cleaning data were followed before training the LM and those are as follows:

- Replace all extra punctuation symbols
- Replace all non-Unicode *devanāgari* characters
- For every article, split the article into paragraphs using the delimiter "\n\n"
- Article from a given source can be prefixed with a series number – say *mann ki baat* data starts with 100000.
- For the data from journal, Oscar and MKB, after splitting an article into paragraphs, every paragraph is given a different id. The articles from Wikipedia were not marked for paragraph numbers. The sample data from each source has been
- Since individual sentences are used for training, each sentence of a given paragraph is linked to the same paragraph id. When these sentences are used for training the LM, one can still retrieve the source paragraph.

The above cleaning led to approximately 4.82 lakh sentences which were used for training languages models.[28] The exact count of data from the final (fnl.txt) file created from each source is as follows:

| Source | Sentence Count |
|---|---|
| **OSCAR** | 96220 |
| **Mann Ki Baat** | 7142 |
| **Wikipedia** | 372348 |
| **Anantaa corpus (summary and non)** | 6807 |
| **Total** | 4,82,517 |

*Table 3: Total Sentence Count*

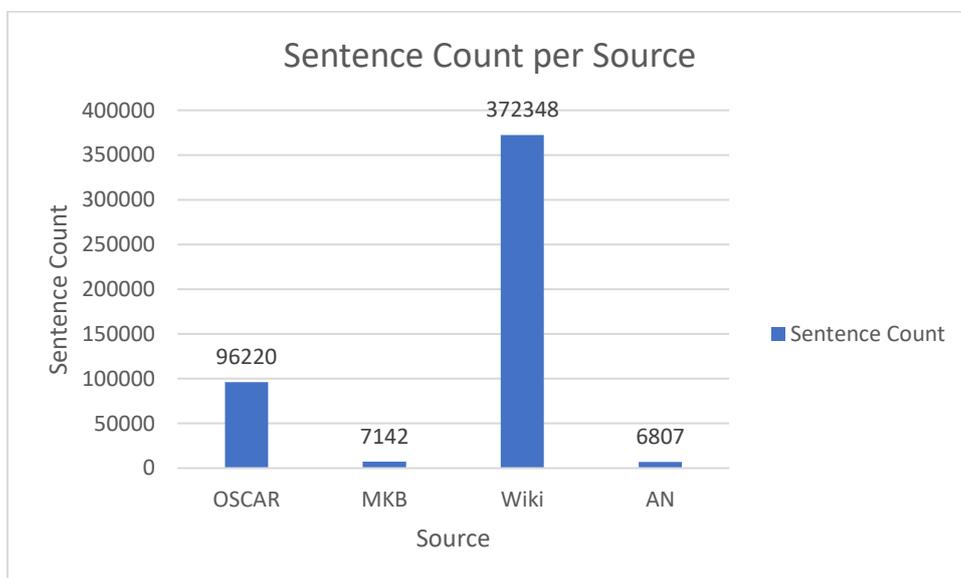

*Figure 4: Sentence Count Per Source*

---

[28] The total sentence count includes the duplicated count as noted above and the same was used for training as well.

The details of the training are reported in the next chapter. 434265 sentences were used for training and 48252 sentences for testing.

Total word count in the final files (fnl*.txt) generated after second and final round of cleaning:

| Source | Word Count |
|---|---|
| **OSCAR** | 778543 |
| **MKB** | 85376 |
| **Anantaa corpus (anantaa_non+an_summ)** | 52360 |
| **Wikipedia** | 2729827 |
| **Total** | 36,46,106 |

*Table 4: Word Count After All Cleaning*

The total token count across all files is 36,46,106 (3.646106e+6). This is comparable with the token count of MURIL (upsampled: 4.3e+6). Since theirs is a multilingual dataset, they upsampled the token count which this thesis does not do.

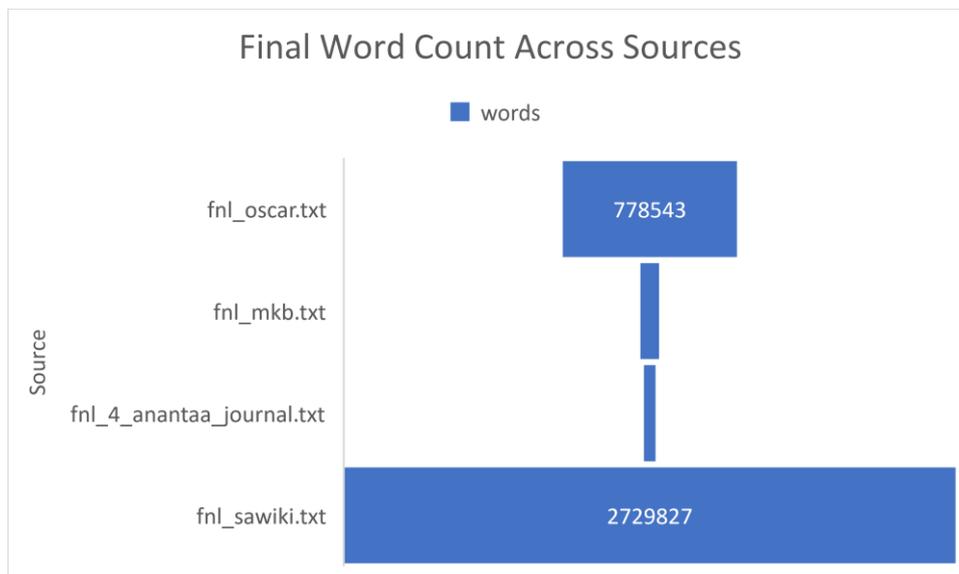

*Figure 5: Total word count across files*

## 3.6 Preparation for Summarization Model Training

Preparing summarization dataset required careful selection of corpus. The two major sources of summarization dataset used in this thesis are – OSCAR and Journal data.

The idea to build SM data from journal articles was inspired by Nikolov (2020). He chose journal data that would provide article-summary pairs. The same motivation drives the data collection mechanism for this work. Articles of the anantaa corpus are mostly focused on Sanskrit literature with a general focus on analysis of the literary sources. A natural outcome of this practice is a more literary language of use in most of these journals. As the section on data preparation revealed, tackling literature-like use required dealing intensively with *Sandhi*. The same process was carried out for SM data development.

### 3.6.1.1 *Basis for Summarization Corpus*

The parallel document-summary data was augmented through OSCAR (Abadji et al., 2021; Kakwani et al., 2020). After a preliminary analysis, I observed that the first sentence of paragraph was reflective of the key elements of the entire paragraphs and hence, was a suitable summary. Therefore, SM data was developed from the OSCAR corpus. However, three methods were used to test if the OSCAR corpus could be used as a summarization parallel dataset: 1) Evidence from literature, 2) Human assessment, 3) Novel n-gram comparison.

### 3.6.1.2 *Evidence from literature*

The first way of verification came from similar approaches used in literature. I traced methods in the literature which augmented parallel summarization data through some techniques in case a parallel data was not available. The method of using first sentence as a summary of the entire paragraph is similar to that of Pilault et al. (2020) in which the authors used just the Introduction instead of the whole document as the source text if the source text was very long. In other words, they mapped the summary to the Introduction section of the text. The paraphrase format of Rush et al. (2015) is also close to this approach in which the first sentence is paired with the title of the piece.[29]

### 3.6.1.3  Human Assessment

The second way verify the suitability of OSCAR data for SM was human assessment of the data. A human evaluator was assigned a set of 50 source-summary pairs randomly taken from the OSCAR and was asked to assess the nature of the corpus – if the generated corpus was of summarization/reflective nature or unrelated. The assessment indicated that close to 80% of the data was summary-suitable (58% summary, 22% reflective, 18% unrelated, 2% other issue). The summary or reflective content was 80% of the total data. Thus, the dataset generated from OSCAR proved to be summary-worthy and hence was taken as document-summary parallel dataset.

---

[29] See Section 3.8 for a quick overview of different dataset formats.

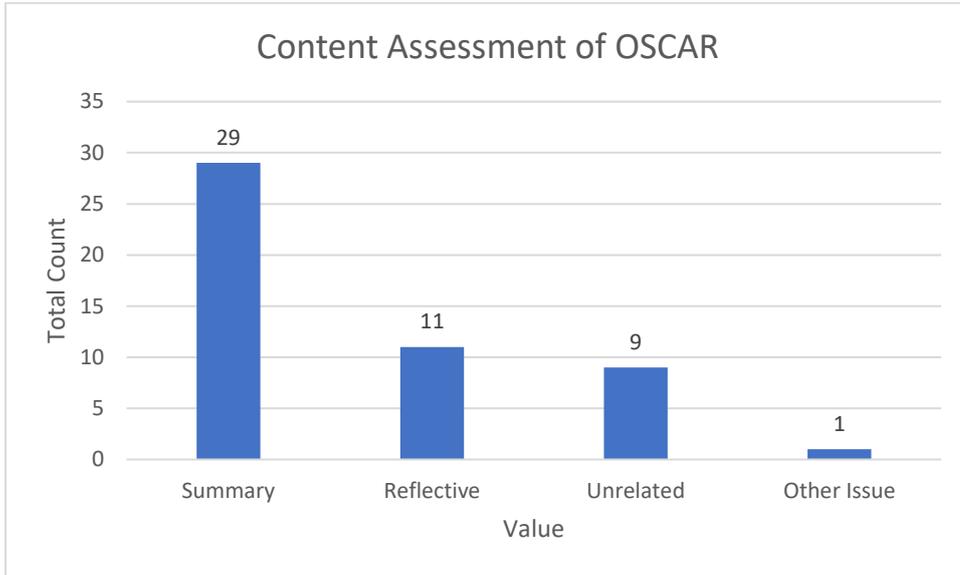
*Figure 6: Human Assessment of OSCAR*

### 3.6.1.4 Novel n-grams in OSCAR

The third and final verification technique, comparison of novel ngrams in the summaries. Additionally, the count of novel n-grams in summary is a considered a mark of its *abstractiveness* (Narayan et al., 2018a). Degree of novel n-gram in OSCAR corpus:

| Novel Unigrams | 85% |
| Novel Bigrams | 98% |

*Table 5: Percentage of novel n-grams in OSCAR parallel corpus*

### 3.6.1.5 Other challenges

A major challenge in developing SM data was managing the length of the source document and summaries. The document-summary pairs extracted from anantaa corpus had different lengths of document and summaries – the document was very long (close to 5000 words per document) with summaries being comparatively very short (100 words). However, with the OSCAR corpus, the document-summary lengths were manageable. In short, the length of documents in the OSCAR corpus was more manageable than those collected from the journal articles.

The final count of parallel dataset for summarization includes:

| Source | Number of article-summary pairs |
|---|---|
| **Anantaa corpus (an_summ)**[30] | 7 |
| **OSCAR** | 15414 |

*Table 6: Summarization Data Count*

Journals contributed 7 articles each with an author-written summary of the pairs. However, for OSCAR, the paragraphs were split to form a parallel dataset of first sentence-paragraph format. The assumption here was that the first sentence reflected the major idea contained in the paragraph.

Given the low number of available summarization and resource constraint on the training platform, the training set size was kept at 0.99 while the test set size was 0.01. Thus, a total of 15268 article-summary pairs was used for summarization training, and 155 for testing. With this, a complete dataset of document-summary pairs was developed for LM and SM training.

## 3.7 Related Work

Data preparation is the core of any NLP task. Literature has examples of database preparation for LRLs such as Icelandic (Nikulásdóttir & Whelpton, 2010) which is used for extracting semantic relations. Such traditional use of data is meant to extract instances and insights that can be used further. However, with the rise of neural methods, algorithms extract such insights through data for furthering the NLP task. Additionally, there is another side of the debate that argues for the use of linguistic resources with DL methods (Goldberg, 2017; Kouris et al., 2021).

---

[30] Only 7 articles in the anantaa corpus contained summary or reflective paragraphs which could be used in the form of a parallel summarization dataset.

## 3.8 Survey of TS datasets in IL and non-ILs

Datasets have been developed and released in English and other high-resource languages.

### 3.8.1 English Datasets

Efforts in English TS are commendable with many research papers being published in it and datasets being released in different domains for both extractive and abstractive tasks (Hermann et al., 2015; Sharma et al., 2019). A closer analysis of the dataset formats reveals that while the language of the datasets is English, the formats vary.

English TS datasets taken from different news sources include CNN/DailyMail dataset (Hermann et al., 2015), Gigaword (Graff et al., 2003; Rush et al., 2015), Newsroom (Grusky et al., 2018). Apart from purely-monolingual TS datasets, some multilingual datasets have also been proposed some of which are derivatives of English datasets. For example, MLSUM (Scialom et al., 2020), Multi-News (Fabbri et al., 2019).

As can be observed, these datasets derive from already existing sources. Chinese dataset LCSTS makes use of "naturally annotated data" (Hu et al. (2015, p. 1967) whose definition is not very clear but it seems like it refers to user-generated data which makes sure that most of the annotations in the form of news sections with headings serving as annotations. Similarly, Koto et al. (2020) observe that the structured nature of news articles make them suitable for being used as summarization dataset (p. 598). However, no uniform principle guiding the creation of such datasets exists and as noted above, the type and length of summaries vary. Different heuristics have been used to extract a data-summary form of text out of such ready-to-use, and mostly, news sources. Some of the popularly methods for dataset development include:

- **Headline-First Sentence pairing**: The first way of forming data-summary pair is pairing the first sentence of a news article with the its headline from the Gigaword dataset (Napoles et al. (2012), Graff et al. (2003), (Rush et al., 2015). The headline-sentence pair format was proposed for sentence summarization but is mostly a task of sentence paraphrasing (Hou et al., 2021, p. 638).

- **Article paired with prewritten summaries** As Narayan et al. (2018a) do in preparing an extreme summarization dataset, XSum, they pick BBC articles paired with short one-sentence summaries of the articles.

- **Articles with bullet point summaries**: CNN/DailyMail dataset by (Hermann et al., 2015) uses summaries in bullet points. As a result, the dataset is considered extractive (Sharma et al., 2019)

- **Summary data without any matching articles**: Some authors have developed new ATS approaches using a mix of ETS and ATS and have not used parallel data. Instead, they have relied on matching summaries with non-matching articles (Nikolov, 2020).

## 3.8.2 Indian Languages Datasets

ILs have witnessed many summarization corpora. Most IL datasets come from news domains (Sinha & Jha, 2022). The domain of news is common across English (Sharma et al., 2019). Alternately, news articles are considered "natural candidate" for building summarization dataset because of their neat presentation and sectional demarcations (Koto et al., 2020, p. 598) which may refer to their ready-to-use headline-summarization format (Sinha & Jha, 2022). Some popular datasets in English have used news articles as the source text with the summary either having been written by the author of the article, like in XSum (Narayan et al., 2018a), or bullet points (Hermann et al., 2015), including the popular DUC datasets.

Three observations are important here. First, IL datasets have different formats and are used for both ETS and ATS tasks but most of these datasets are not publicly available and are small in size (Hu et al., 2015; Mamidala & Sanampudi, 2021; Sinha & Jha, 2022). Second, many dataset sources have missing summaries. The Hindi summarization corpus offered by Arora (2020) reports that for 53% of the Hindi news articles the corresponding summaries are not available.[31] For Punjabi dataset used by (Gupta & Kaur, 2016), the corpus is monolingual and therefore, is not suitable for ATS task. Third, IL datasets continually face the challenge of corpus size – large corpora are rarely developed. D'Silva and Sharma (2019) prepare a summarization corpus of about 71 folktales. Folktales are also used as the domain in (Droog-Hayes, 2019) for Russian TS. On a similar note, folktales and stories are also used for narrative summarization (Lehnert, 1999) but are not a part of this thesis.

---

[31] https://www.kaggle.com/datasets/disisbig/hindi-text-short-and-large-summarization-corpus?resource=download

Some other attempts at developing datasets have also been made in the IL sphere. NLP for ILs has used pre-training to develop many LMs for ILs. Some of the common datasets or sources of datasets used in pre-training are: First, Dakshina is a dataset for Indian languages in Latin and native scripts (Roark et al., 2020). However, it does not include Sanskrit. Second, XTREME is a dataset for (X) cross-lingual Transfer Evaluation of Multilingual Encoders (Hu et al., 2020). XTREME offers dataset for 40 languages including 4 Indo-Aryan languages. Third, Wikipedia dumps.[32] While the aforementioned datasets also include Wikipedia articles, Wikipedia dumps is a popular source for accessing Wikipedia articles for a given language. IL datasets are not very widely released nor are there platforms for wide development of IL datasets in TS or NLP in general (Sinha & Jha, 2022). This thesis considers single sentence corpora as the LM corpus whereas summarization corpus has been developed using article-summary pairs as noted in an earlier section.

No summarization corpora exist for Sanskrit. Some monolingual corpora include the literary *Itihasa* corpus by Aralikatte et al. (2021), and CC-100 corpus proposed by Conneau et al. (2020).

### 3.8.2.1 Sandhi Corpora

Different *sandhi*-split corpora have been released in Sanskrit.[33,34] Attempts like the *Sandhikosh* corpus (Bhardwaj et al., 2018) have been made where scholars have also compared the existing

---

[32] https://dumps.wikimedia.org/
[33] https://sanskrit.uohyd.ac.in/
[34] http://www.sanskrit-linguistics.org/dcs/

*sandhi* analysis tools like JNU tool[35], Sansadhani[36] indicate which points to the inherent need to have better *Sandhi* processing tools.

Attempts have been made to automate the *sandhi* extraction process and even to train models that can break the *sandhi* instances (Reddy et al., 2018). *Sandhi* is a unique characteristic of Sanskrit text due to which letters undergo transformation when followed by some specific letters that trigger those transformations. While many issues with and around *Sandhi* are relevant for discussion in Sanskrit NLP, the one key issue to be discussed is how *Sandhi* impacts the number of unique tokens. This work will later use certain available tokenization algorithms[37] for preprocessing data before using it in training. These algorithms require specific forms of tokenization which is not based on simple split. Aralikatte et al. (2021) propose that splitting *Sandhi*s would enhance the tokenization results for their Sanskrit corpus named *Itihasa*.

The next chapter describes the training setup and reports the results.

---

[35] http://sanskrit.jnu.ac.in/*sandhi*/viccheda.jsp
[36] https://sanskrit.uohyd.ac.in/
[37] Wordpiece, BPE, etc.

# Chapter 4: Experiment and Results

This thesis builds on the methodology offered by Rothe et al. (2020) in which different combinations of pretrained LMs are tested on the downstream task of summarization. The details of the methodology and in turn the DL architectures used for training and testing LM and SM models in this work are described in the upcoming sections.[38] The research questions from the third theme, RQ nos. 3(a) and 3(b) are answered here.

## 4.1 Meta details

The success of deep learning emerged from the emergence of computational graphs. Rao and McMahan (2019) note that a computational graph implements mathematical expressions that are used in DL. Such expressions encode inputs and predict target values. Additionally, gradient calculation methods in DL help modify the weights of the model to improve results. PyTorch is a graph-implementation framework that helps in keeping a record of the gradients thereby enabling better gradient calculation (pp. 10-11). This thesis is based on models from the HuggingFace (HF) Transformers library[39] which are based on the PyTorch framework (Wolf et al., 2019).

---

[38] More than 10 sets of final experiments were run before finalizing the experiment that has been reported here.
[39] The codes from the HF library were also adapted for this work and have been cited at relevant places. The Transformers library and Model Hub provided by HF provide functionalities for seamless experience in training and testing.

### 4.1.1 Sources Used: LM and SM training

As stated earlier, three LMs were pre-trained resulting in pre-trained language models (PTLMs) for later use on a downstream task of summarization. Those PTLMs are used to initialize the Transformer model. The codes for different aspects of the training pipeline were stored in different folders (Appendix A1). Code snippet adapted from HF library came from the summarization and accelerate portions.[40,41]

### 4.1.2 Methodology and Algorithm

The key steps to training PTLMs are:

a) Upload and clean data

b) Train LMs – each of BERT, GPT-2, RoBERTa on the cleaned data

c) Generate perplexity score, training loss, evaluation loss,

## 4.2 Hyperparameters

The model wise hyperparameter details are given here:

BERT:

```
"max_length": 128,
"learning_rate": 2e-5,
"weight_decay": 0.01,
"num_epochs": 80,
"train_batch_size": 48,
"test_batch_size": 48,
"seed": 42,
```
*Table 7: Hyperparameters for BERT Training*

GPT-2

---

[40] . https://sanskritdocuments.org/sites/manogatam/
[41] Ibid.

```
"max_length": 128,
  "learning_rate": 2e-5,
  "weight_decay": 0.01,
  "num_epochs": 50,
  "train_batch_size": 48,
  "test_batch_size": 48,
  "seed": 49,
```

*Table 8: Hyperparameters for GPT-2 training*

RoBERTa

```
"max_length": 128,
  "learning_rate": 2e-5,
  "weight_decay": 0.01,
  "num_epochs": 70,
  "train_batch_size": 48,
  "test_batch_size": 48,
  "seed": 49,
```

*Table 9: Hyperparameters for RoBERTa (masked) training*

#### *4.2.1.1 Cleaning*

The previous chapter elaborated on the cleaning steps that were used to prepare the data for submitting as input to the model. A second round of cleaning was also applied before sending the final data to the algorithm. Two cleaning codes were used – clean_data_lm, clean_data_summ. Each code cleaned the data for language modeling and summarization respectively.

### 4.3 LM Training

BERT, GPT, RoBERTa (masked and causal) were trained on the LM corpus.

a. The training for BERT took place on 8 TPU cores and took around 22 hours to train for 77 epochs (training set to 80 epochs stopped at 77 due to early stopping).

b. The training for GPT-2 took place on 8 TPU cores and took 13 hours to train for 50 epochs.

c.  Training for RoBERTa masked training took place on 8 TPU cores and took 18 hours to train for 70 epochs.

The training and evaluation losses along with perplexity scores have been indicated in the tables below.

### 4.3.1 Losses and Perplexity

Perplexity measure for each LM indicates the quality of training of the model. Similarly, training and evaluation losses help gauge the learning of the model. Epoch wise perplexity scores have been reported in the Appendix. To prevent overfitting, early stopping was applied.[42] However, the final perplexity scores for each model are as follows:

---

[42] Code adapted from: https://wandb.ai/ayush-thakur/huggingface/reports/Early-Stopping-in-HuggingFace-Examples--Vmlldzo0MzE2MTM

### 4.3.1.1 BERT:

| Training Loss | Evaluation Loss | Perplexity |
|---|---|---|
| **6.480** | **6.394** | **598.39** |

*Table 10: Losses and perplexity score for BERT*

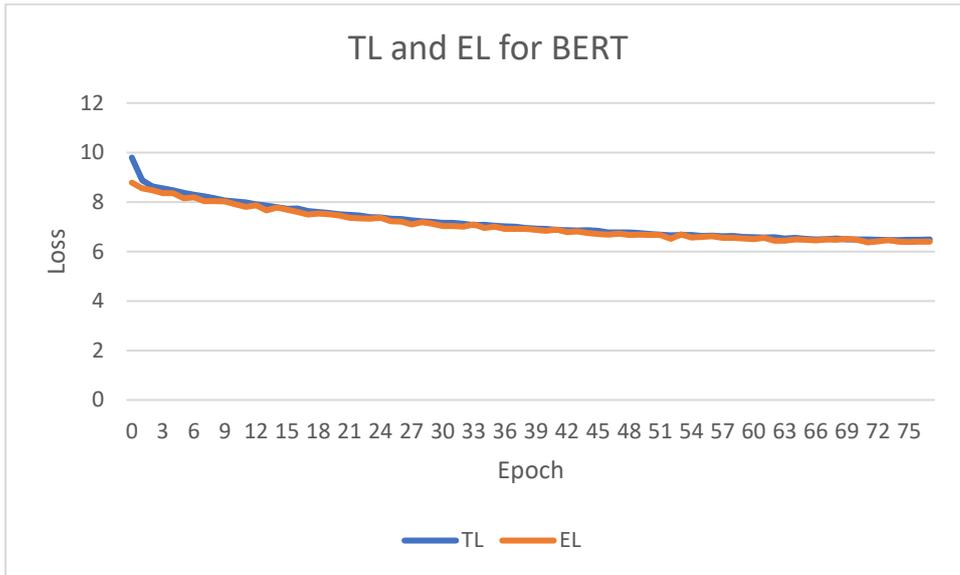

*Figure 7: TL and EL for BERT*

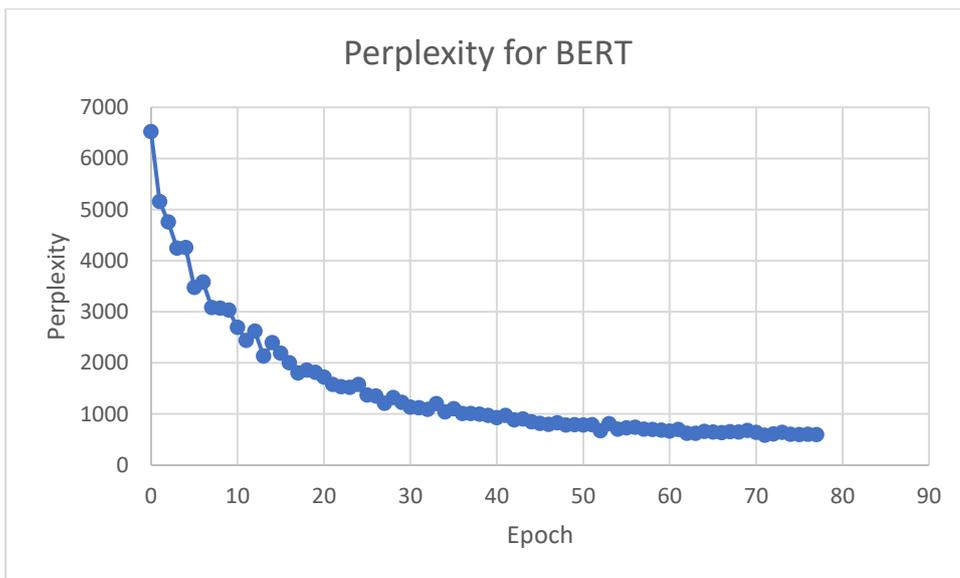

*Figure 8: BERT Perplexity Score Across Epochs*

*4.3.1.2   GPT-2:*

| Training Loss | Evaluation Loss | Perplexity |
|---|---|---|
| **1.218** | **1.512** | **4.53** |

*Table 2:  Losses and perplexity score for GPT-2*

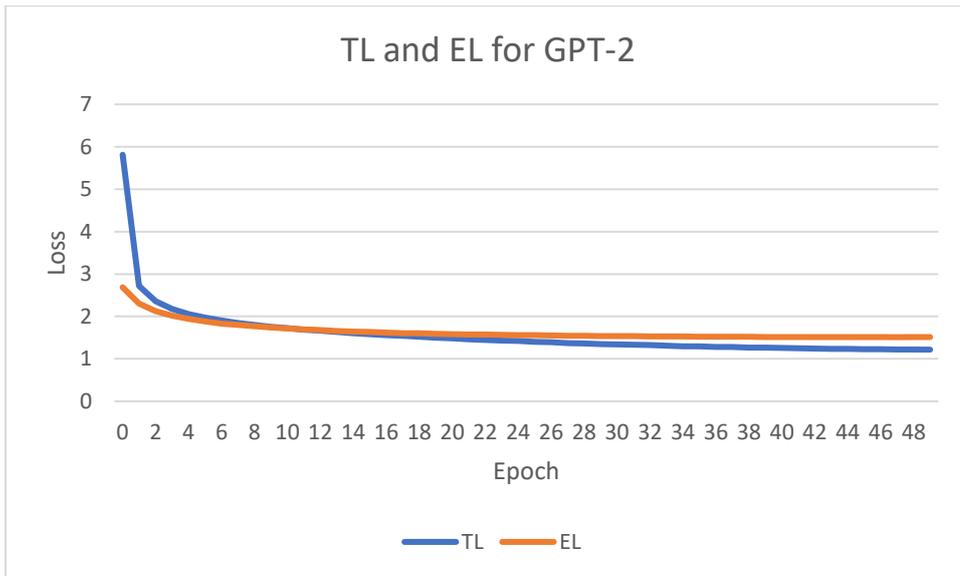

*Figure 9: TL and EL for GPT2 across Epochs*

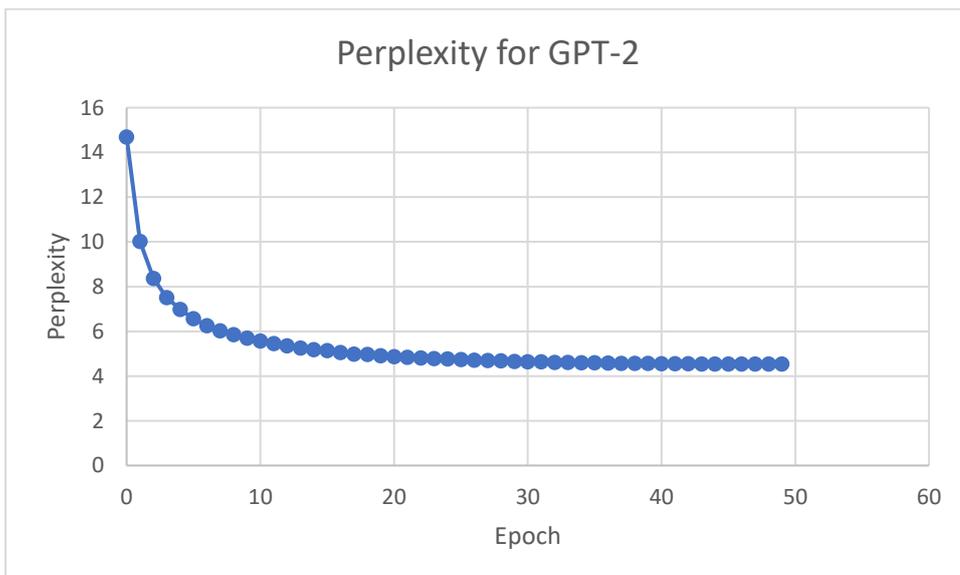

*Figure 10: Perplexity Scores for GPT2 across Epochs*

### *4.3.1.3 RoBERTa Masked:*

| Training Loss | Evaluation Loss | Perplexity |
|---|---|---|
| **0.902** | **0.879** | **2.409** |

*Table 3: Losses and perplexity score for RoBERTa Masked*

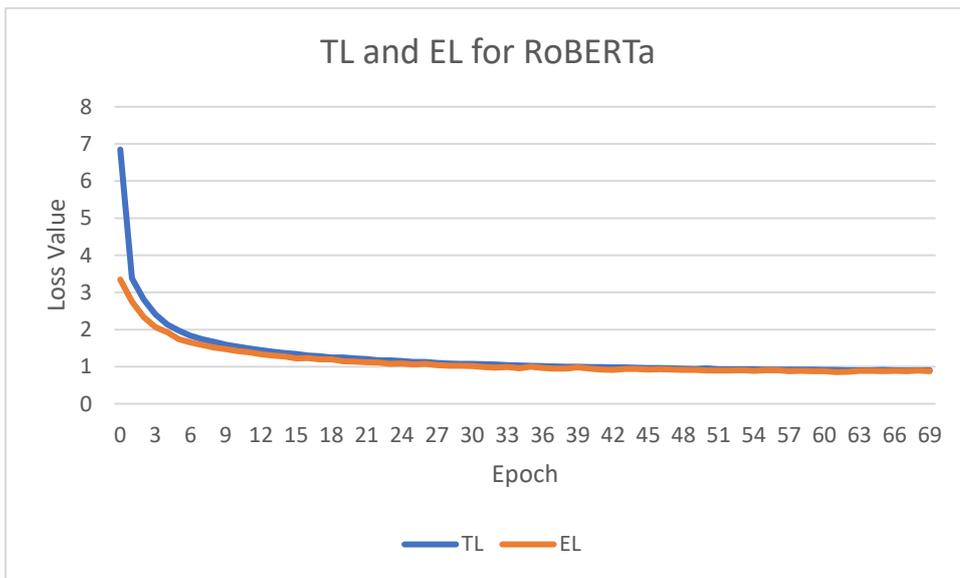

*Figure 11: TL and EL for RoBERTa*

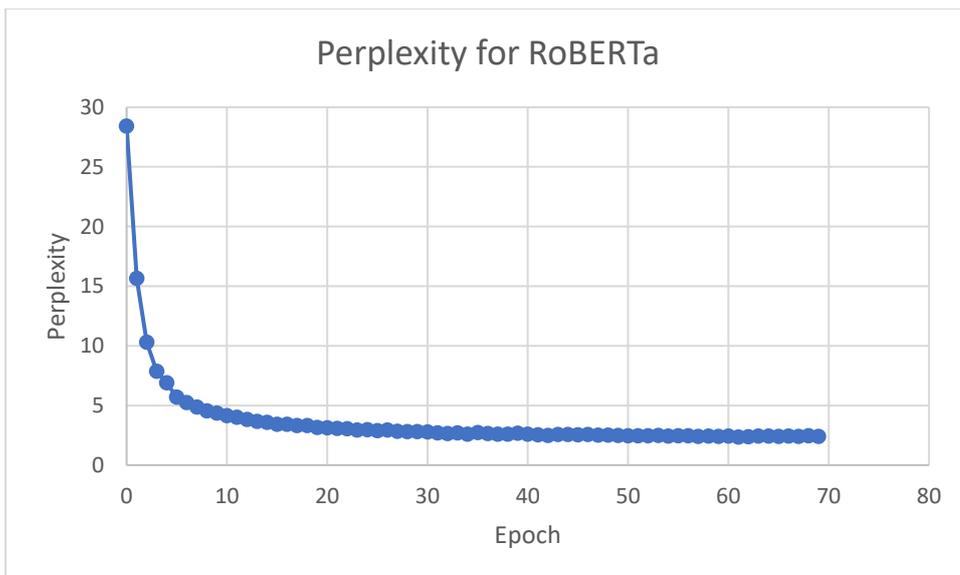

*Figure 12: Perplexity Scores for RoBERTa across Epochs*

### 4.3.2 LM pretraining objectives

The pre-training objectives of different LM training have been indicated below. Masked language modeling (MLM) served as the pretraining objective for BERT and RoBERTa. On the other hand, GPT-2 is trained for next word prediction, i.e., left to right language modeling objective (Lewis et al., 2019). The results of the pre-training objectives are presented here:

#### *4.3.2.1   BERT*

```
fill_mask(f"परन्तु तस्याः पित्रे भीखाईजी इत्यस्याः fill_mask.tokenizer.mask_token कार्यं न रोचते स्म")
```

```
('score': 0.45929938554763794,
 'sequence': 'परन्तु तस्याः पित्रे भीखाईजी इत्यस्याः संस्थायाः कार्यं न रोचते '
          'स्म',
 'token': 3029,
 'token_str': 'संस्थायाः'),
 -------------------------------------
 ('score': 0.008943749591708183,
 'sequence': 'परन्तु तस्याः पित्रे भीखाईजी इत्यस्याः विषये कार्यं न रोचते स्म',
 'token': 908,
 'token_str': 'विषये'),
 -------------------------------------
 ('score': 0.004357779398560524,
 'sequence': 'परन्तु तस्याः पित्रे भीखाईजी इत्यस्याः कृते कार्यं न रोचते स्म',
 'token': 980,
 'token_str': 'कृते'),
 -------------------------------------
 ('score': 0.004174566827714443,
 'sequence': 'परन्तु तस्याः पित्रे भीखाईजी इत्यस्याः उपरि कार्यं न रोचते स्म',
 'token': 1597,
 'token_str': 'उपरि'),
 -------------------------------------
```

```
('score': 0.0031881025061011314,
 'sequence': 'परन्तु तस्याः पित्रे भीखाईजी इत्यस्याः योजनायाः कार्यं न रोचते '
         'स्म',
 'token': 7946,
 'token_str': 'योजनायाः')
```

*Figure 13: BERT Pretraining MLM Sample Output*

### 4.3.2.2  GPT-2

text = f"दशभिः पुत्रैः यत् पुण्यं प्राप्यते " # एकया पुत्र्यातत् पुण्यं लभ्यते"

**Beam Output with max_length = 50**

दशभिः पुत्रैः यत् पुण्यं प्राप्यते तत् पुण्यं प्राप्यते इति विश्वासः

----------------------------------------

**Beam Output with no_repeat_ngrams = 2**

दशभिः पुत्रैः यत् पुण्यं प्राप्यते तत् सर्वं दुःखं भवति । ॐ ॐः रोगैः पीड्यमान

----------------------------------------

**Greedy Output:**

दशभिः पुत्रैः यत् पुण्यं प्राप्यते तत् पुण्यं च भवति । ॐ च पुण्य

*Figure 14: GPT-2 LM Sample Generation Output*

These outputs generated

### 4.3.2.3  RoBERTa Masked

fill_mask(f"घोरी इत्येषः 1190 तमे वर्षे प्रथमवारं सुनियोजितरीत्या सपादलक्षसाम्राज्यस्योपरि आक्रमणम् <fill_mask.tokenizer.mask_token> ।")

```
('score': 0.9999511241912842,
 'sequence': 'घोरी इत्येषः  तमे वर्षे प्रथमवारं सुनियोजितरीत्या '
         'सपादलक्षसाम्राज्यस्योपरि आक्रमणम् । ।',
 'token': 88,
 'token_str': ' ।'),
----------------------------------------
```

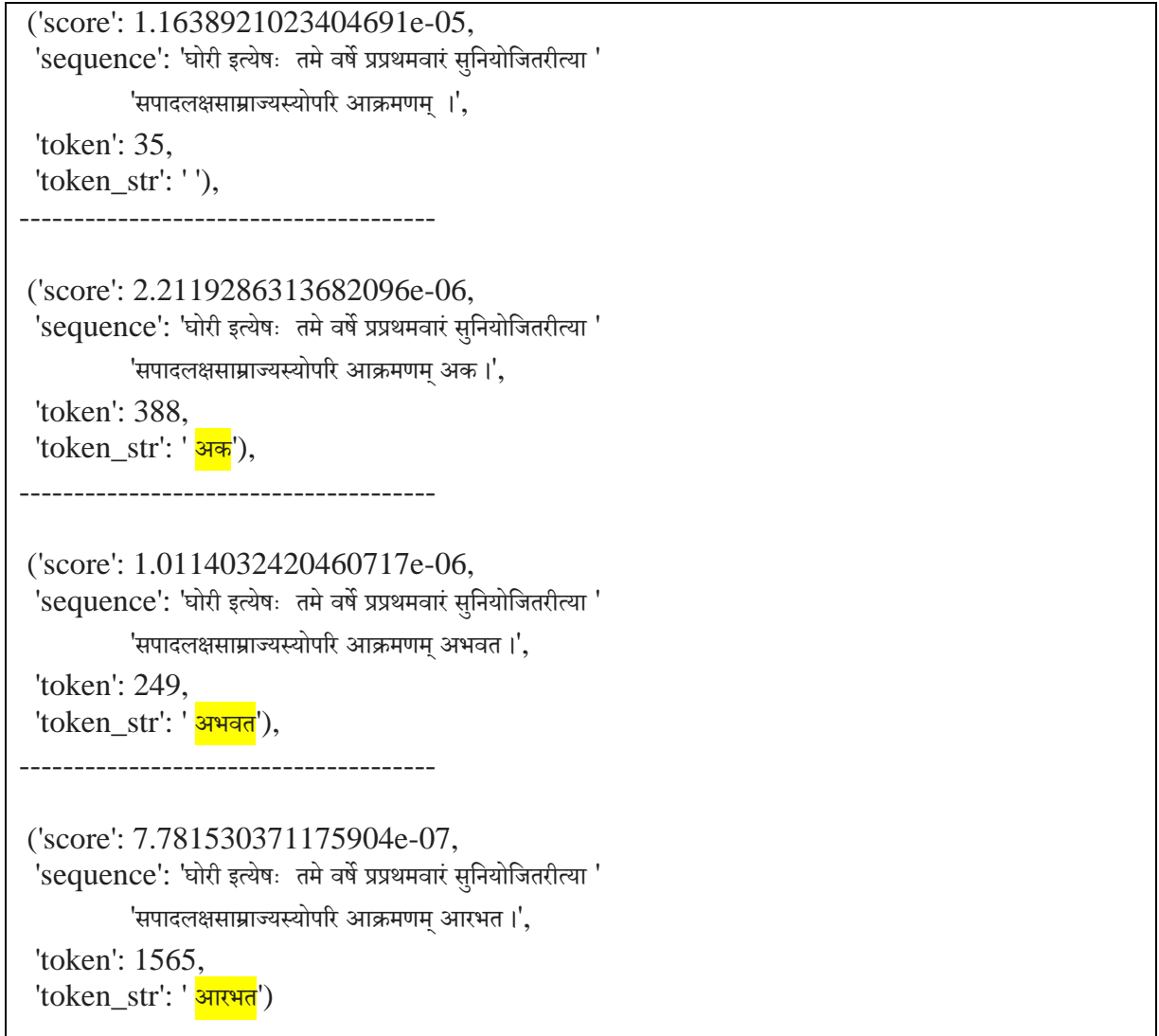

Figure 15: RoBERTA LM Pretraining Token Prediction Sample Output

For the masked token अकरोत्, the pre-training objective produced four different tokens and surprisingly all are close to correct.

A comparison of the perplexity scores of the above models with Sanskrit PTLMs presented by others is presented in Table 5. The best perplexity is highlighted. MURIL did not report perplexity scores as a result of which it has not been included in the comparison:

| Source | Type of LM | Perplexity |
|---|---|---|
| Kumar et al. (2020) | MLLM (BERT) | 399 |
| Arora (2020) | MonoLM (TransformerXL) | ~3 |
| Arora (2020) | MonoLM (ULMFiT) | ~6 |
| This thesis | MonoLM, BERT | 598.38 |
| This thesis | MonoLM, GPT-2 | 4.54 |
| This thesis | MonoLM, RoBERTa | **2.41** |

*Table 11: Comparison of different PTLMs with the presently trained models*

The following may be observed in the comparison presented in the trained models above:

1. MonoLM RoBERTa has the best perplexity score.

2. The MonoLM BERT trained in this work has weaker perplexity than the Sanskrit MLLM based on BERT trained by Kumar et al. (2020).

3. Perplexity of monoLMs GPT-2 and RoBERTa are both comparable with and better than the monoLMs trained by Arora (2020).

## 4.4 Summarization Training

Steps to training summarization models:

1. Train 10 combinations of Transformer models initialized with different encoder-decoder formats (BERT2GPT, GPT2BERT, etc. as per the methodology of Rothe et al. (2020)).

2. Training Loss (TL) and Evaluation Loss (EL) are measured during the training and indicate the quality of training. If the TL is higher than the EL, the model is said to have overfit on the training data. Alternately, if the EL is higher than the TL, the model is said to have underfit. The best match is where the TL and EL converge.[43] Hence, early stopping was applied to get the best training.

3. Evaluation:
    a. ROUGE scores
    b. Human Evaluation

To prevent overfitting or underfitting, early stopping[44] was applied. The qualitative and textual evaluation will be reported in the next chapter. All the ten sequence generation models were trained for about 12 hours on Google Colab with TPU support High RAM. Each model was run till the evaluation and training losses converged. This chapter only reports the ROUGE and the two losses – TL and EL.

---

[43] https://www.baeldung.com/cs/training-validation-loss-deep-learning last accessed October 30, 2022, at 10.34 am.
[44] Code adapted from: https://wandb.ai/ayush-thakur/huggingface/reports/Early-Stopping-in-HuggingFace-Examples--Vmlldzo0MzE2MTM

### 4.4.1 An Overview of Model Combinations

The model combinations as described in Rothe et al. (2020) are summarized as follows:

The encoder and decoder in these experiments are initialized with different checkpoints. Each setup is named in the format 'A2B' where A is the encoder and B is the decoder. Except for GPT2 which is a decoder-only setup, all combinations are encoder-decoder-based. In the RND2RND setup, both the encoder and decoder are randomly initialized. In BERT2BERT, both the encoder and decoder are initialized with BERT checkpoints. In BERTSHARE, the encoder and decoder setups share the checkpoints. Likewise, all the models in the format of A2B, get the encoder checkpoints from an LM A (RND, BERT, RoBERTa) and decoder checkpoints from B (Rothe et al., 2020, p. 266). The RND2RND checkpoints serve as the baseline.

### 4.4.2 BERT2BERT

The results for the model BERT2BERT are as follows:

| Epoch | Training Loss | Evaluation Loss |
|---|---|---|
| 0 | 9.78 | 7.18 |
| 1 | 7.4 | 6.98 |
| 2 | 6.95 | 6.87 |
| 3 | 6.49 | 6.82 |
| 4 | 6.2 | 6.9 |
| 5 | 5.75 | 6.81 |
| 6 | 5.43 | 6.88 |
| 7 | 5.19 | 6.82 |

*Table 12: Training and Evaluation Loss for Epochs*

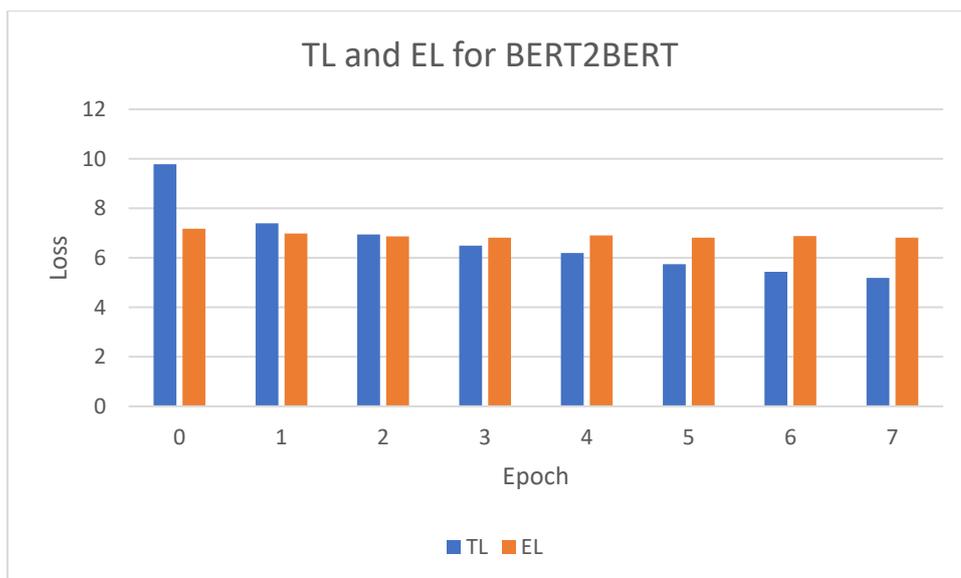

*Figure 16: TL and EL for BERT2BERT[45]*

---

[45] The visualizations have all been generated using automated tools. Scales differ across graphs and the same should be taken into account while observing the graphs. For some, the Y-axis ranges from 0 to 12 with an increment of 2 units, while for some it ranges from 0 to 4 with an increment of 0.5 units. This must be noted while observing the graphs.

### 4.4.3 BERT2GPT

The results for the model are as follows:

| Epoch | Training Loss | Evaluation Loss |
|---|---|---|
| 0 | 4.06 | 1.32 |
| 1 | 1.56 | 1.24 |
| 2 | 1.35 | 1.17 |
| 3 | 1.2 | 1.15 |
| 4 | 1.12 | 1.23 |
| 5 | 1.04 | 1.18 |
| 6 | 1.05 | 1.21 |
| 7 | 0.9 | 1.23 |
| 8 | 0.8 | 1.26 |

*Table 13: Training and Evaluation Loss for BERT2GPT*

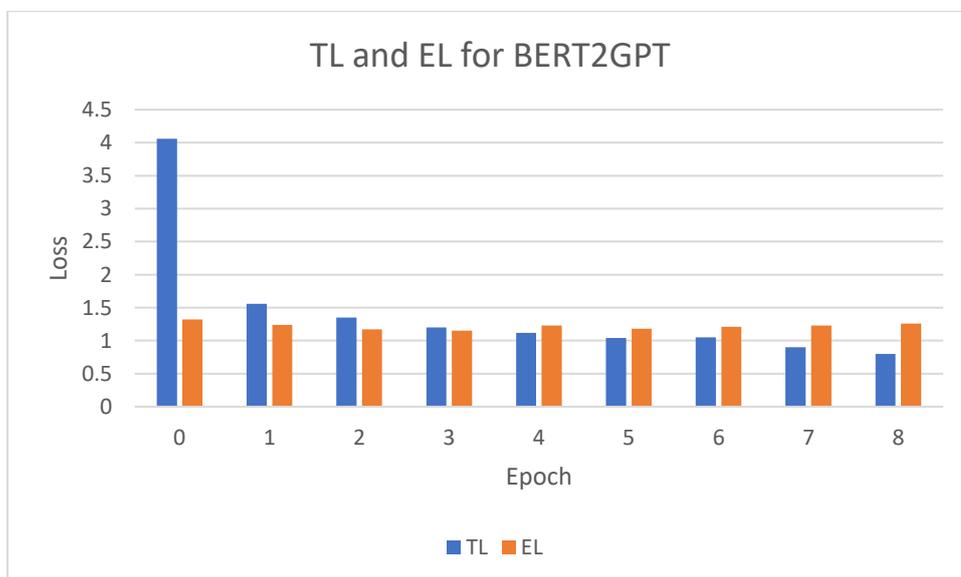

*Figure 17: TL and EL for BERT2GPT*

### 4.4.4 BERT2RND

The results for the model are as follows:

| Epoch | Training Loss | Evaluation Loss |
|---|---|---|
| 0 | 9.37 | 8.27 |
| 1 | 8.56 | 7.99 |
| 2 | 8.29 | 7.78 |
| 3 | 7.99 | 7.71 |
| 4 | 7.81 | 7.65 |
| 5 | 7.46 | 7.74 |
| 6 | 7.22 | 7.71 |
| 7 | 7.09 | 7.72 |
| 8 | 6.79 | 7.69 |
| 9 | 6.61 | 7.72 |

*Table 14: Training, and Evaluation Loss for BERT2RND*

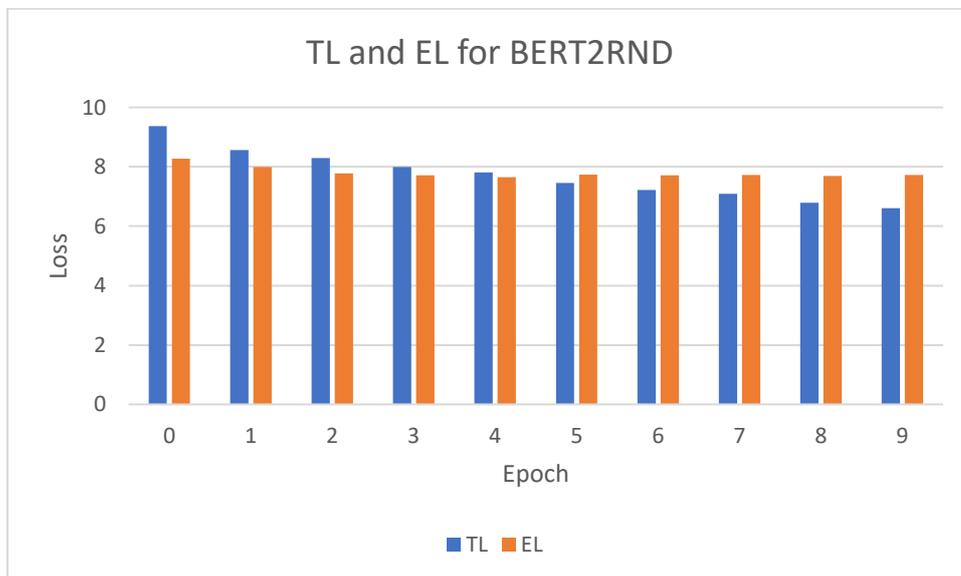

*Figure 18: TL and EL for BERT2RND*

### 4.4.5 BERTShare

The results for the model are as follows:

| Epoch | Training Loss | Evaluation Loss |
|---|---|---|
| 0 | 10.09 | 7.64 |
| 1 | 7.92 | 7.31 |
| 2 | 7.61 | 7.17 |
| 3 | 7.31 | 7.18 |
| 4 | 7.19 | 7.13 |
| 5 | 6.9 | 7.09 |
| 6 | 6.71 | 7.2 |
| 7 | 6.58 | 7.22 |
| 8 | 6.3 | 7.24 |
| 9 | 6.14 | 7.32 |

*Table 15: Training and Evaluation Loss for BERTSHARE*

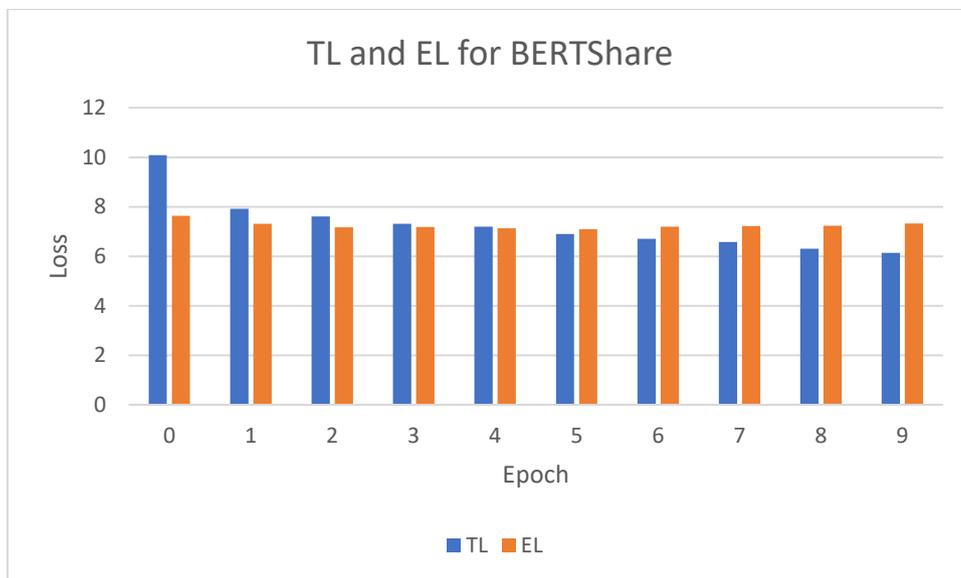

*Figure 19: TL and EL for BERTSHARE*

### 4.4.6 RND2BERT

The results for the model are as follows:

| Epoch | Training Loss | Evaluation Loss |
|---|---|---|
| 0 | 9.56 | 7.33 |
| 1 | 7.55 | 7.09 |
| 2 | 7.14 | 6.96 |
| 3 | 6.74 | 6.89 |
| 4 | 6.53 | 6.86 |
| 5 | 6.15 | 6.79 |
| 6 | 5.91 | 6.75 |
| 7 | 5.71 | 6.75 |

*Table 16: Training and Evaluation Loss for RND2BERT*

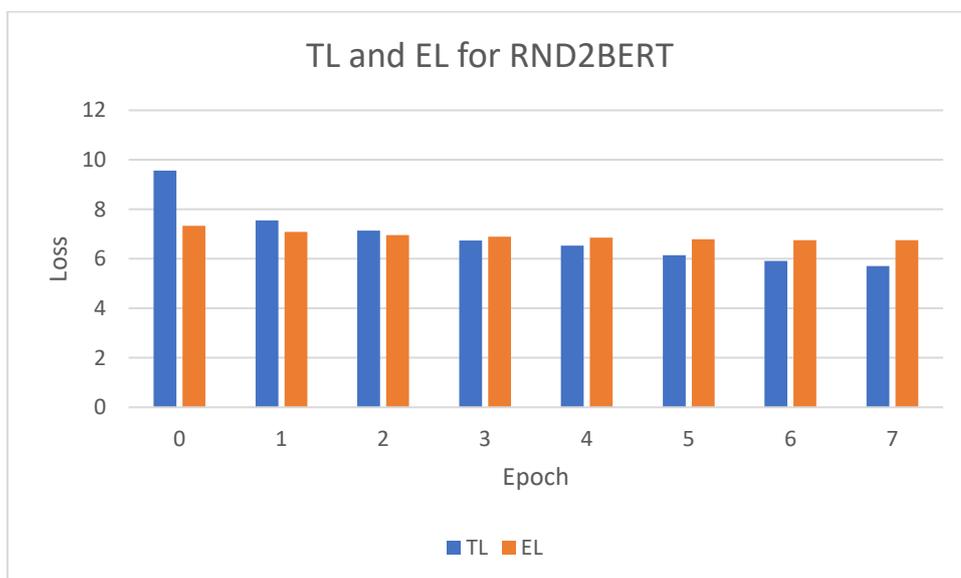

*Figure 20: TL and EL for RND2BERT*

### 4.4.7 RND2GPT

The results for the model are as follows:

| Epoch | Training Loss | Evaluation Loss |
|---|---|---|
| 0 | 4.41 | 1.38 |
| 1 | 1.65 | 1.27 |
| 2 | 1.38 | 1.19 |
| 3 | 1.22 | 1.18 |
| 4 | 1.13 | 1.21 |
| 5 | 1.01 | 1.22 |
| 6 | 0.96 | 1.22 |
| 7 | 0.89 | 1.26 |
| 8 | 0.83 | 1.29 |

*Table 17: Training and Evaluation Loss for RND2GPT*

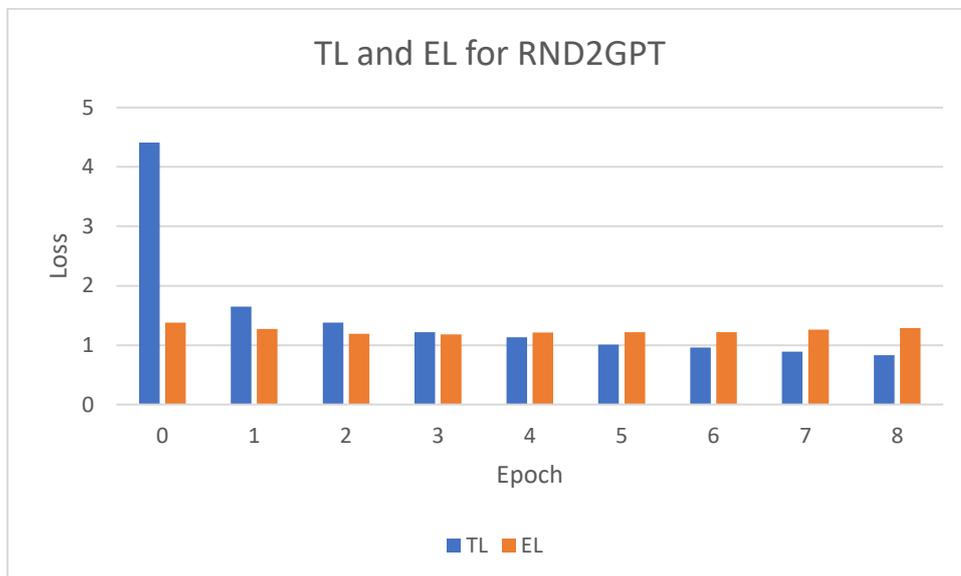

*Figure 21:TL and EL for RND2GPT*

### 4.4.8 RND2RND

The results for the model are as follows:

| Epoch | Training Loss | Evaluation Loss |
|---|---|---|
| 0 | 9.36 | 8.47 |
| 1 | 8.66 | 8.25 |
| 2 | 8.42 | 8.04 |
| 3 | 8.15 | 7.94 |
| 4 | 8.02 | 7.84 |
| 5 | 7.74 | 7.8 |
| 6 | 7.55 | 7.69 |
| 7 | 7.44 | 7.63 |
| 8 | 7.19 | 7.64 |
| 9 | 7.02 | 7.57 |

*Table 18: Training and Evaluation Loss for RND2RND*

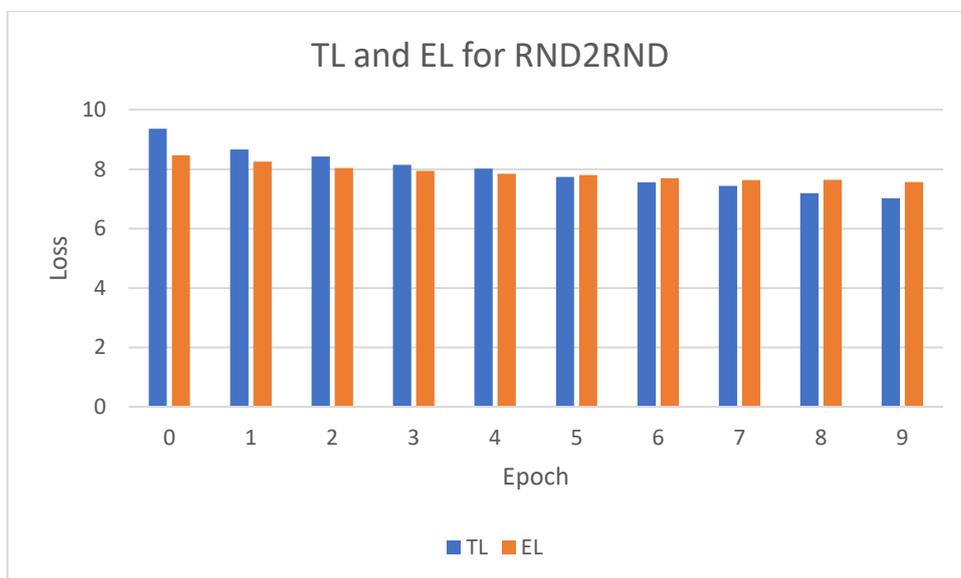

*Figure 22: TL and EL for RND2RND*

### 4.4.9 RoBERTa2GPT

The results for the model are as follows:

| Epoch | Training Loss | Evaluation Loss |
|---|---|---|
| 0 | 3.43 | 1.26 |
| 1 | 1.47 | 1.21 |
| 2 | 1.29 | 1.15 |
| 3 | 1.15 | 1.14 |
| 4 | 1.07 | 1.18 |
| 5 | 0.95 | 1.18 |
| 6 | 0.89 | 1.19 |
| 7 | 0.82 | 1.18 |
| 8 | 0.74 | 1.25 |

*Table 19: Training and Evaluation Loss for RoBERTa2GPT*

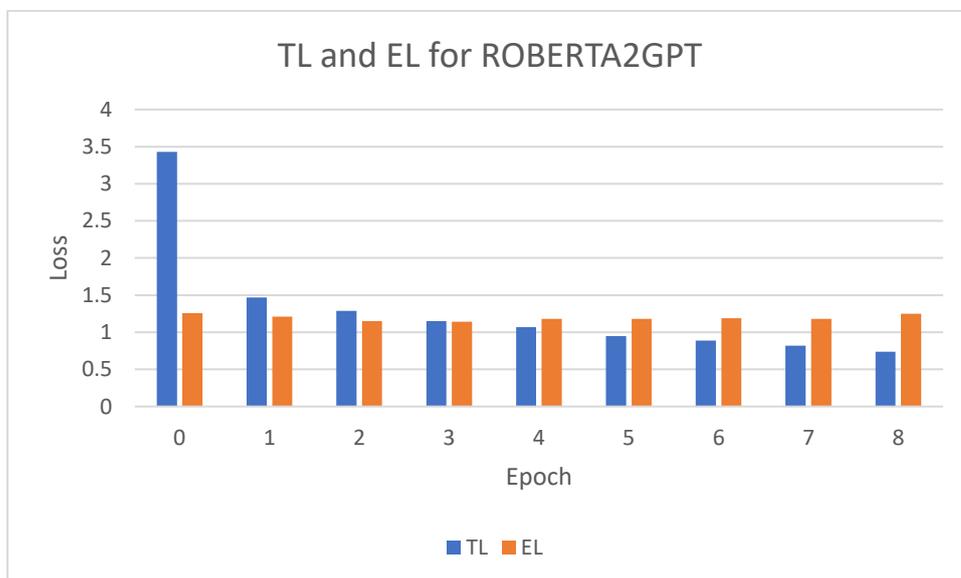

*Figure 23: TL and EL for RoBERTa2GPT*

### 4.4.10 RoBERTa Share

The results for the model are as follows:

| Epoch | Training Loss | Evaluation Loss |
|---|---|---|
| 0 | 6.05 | 2.87 |
| 1 | 3.12 | 2.17 |
| 2 | 2.45 | 2.06 |
| 3 | 2.32 | 2.01 |
| 4 | 2.27 | 1.99 |
| 5 | 2.19 | 1.95 |
| 6 | 2.16 | 1.91 |
| 7 | 2.13 | 1.91 |
| 8 | 2.06 | 1.9 |
| 9 | 2.02 | 1.87 |
| 10 | 1.98 | 1.89 |
| 11 | 1.93 | 1.92 |
| 12 | 1.87 | 1.93 |
| 13 | 1.84 | 1.91 |
| 14 | 1.81 | 1.92 |
| 15 | 1.74 | 1.94 |
| 16 | 1.71 | 1.97 |

*Table 20: Training and Evaluation Loss for RoBERTaShare*

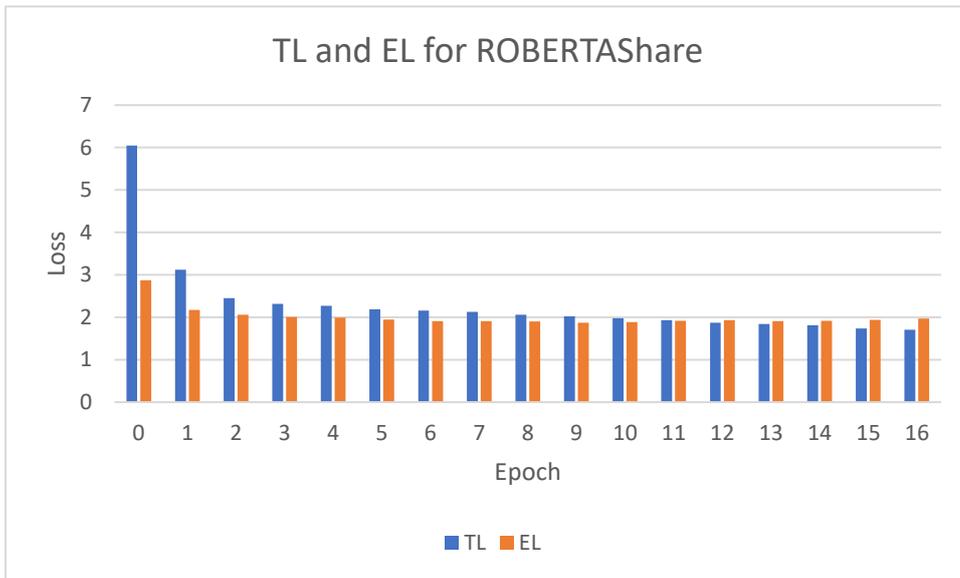

*Figure 24: TL and EL for RoBERTaShare*

### 4.4.11 RoBERTa2RoBERTa

The results for the model are as follows:

| Epoch | Training Loss | Evaluation Loss |
|---|---|---|
| 0 | 5.29 | 2.01 |
| 1 | 2.21 | 1.86 |
| 2 | 2.05 | 1.81 |
| 3 | 1.93 | 1.77 |
| 4 | 1.87 | 1.74 |
| 5 | 1.76 | 1.72 |
| 6 | 1.7 | 1.67 |
| 7 | 1.64 | 1.68 |
| 8 | 1.54 | 1.7 |
| 9 | 1.46 | 1.65 |
| 10 | 1.39 | 1.65 |
| 11 | 1.32 | 1.74 |

*Table 21: Training and Evaluation Loss for RoBERTA2RoBERTa*

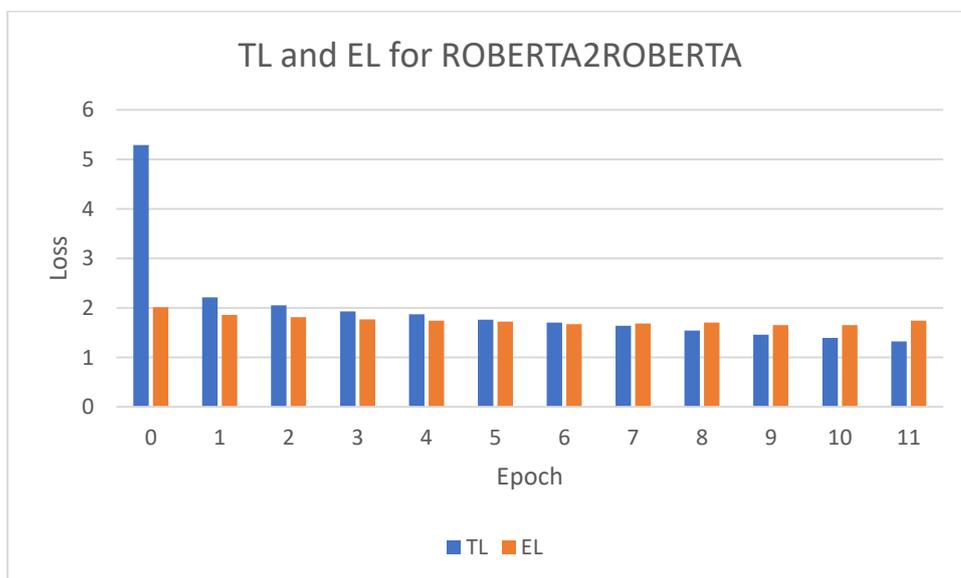

*Figure 25: TL and EL for RoBERTa2RoBERTa*

The next chapter reports the automatic evaluation scores, preliminary analysis results, and human evaluation results along with a discussion on the models' performance.

## 4.5 Limitations of compute and other resources

The training conducted for this research highlighted some important factors of the process which must be considered by future Sanskrit TS researchers.

First, the availability of data resources. While *crawalable* data might be less (which is why we have the low resource problem in Sanskrit), the available sources also need effortful extraction since they come in PDFs or other non-text formats. That is, most Sanskrit sources come in PDF or document formats extracting readable text from which was the first challenge faced in this work. Relatedly, developing summarization dataset was another challenge. Literature survey indicates that summaries in TS datasets are never written by researchers themselves – that is, the developers of the TS datasets almost always depend on a third-party sources to get summaries, such as author-written summaries or headlines as the summary (Narayan et al., 2018a). In any case, the summaries are never objective. This idea helped build the summarization corpus pipeline.

The second obstacle was the availability of compute resources. Philip et al. (2021) convincingly argue that for neural machine translation (NMT) training involving large resources, the adequate compute resources are not available with most academic groups (pp. 178-179). In this work as well, I was consistently constrained by resources, especially computational resources.

This thesis is a work of application – I apply existing neural approaches on Sanskrit data and report the results based on Rothe et al. (2020) which is based on language model clubbed with downstream tasking approach. Theirs is a popular approach especially after the advent of transformers and attention-based models. However, it required huge computing resources for training which were not readily available to me. Hence, I used some of the available cloud computing (CC) services to train DL algorithms and run the 10 encoder-decoder combinations as Rothe et al. (2020) did. The training required for a small dataset like this work's sought huge memory and other hardware resources. Given that each training round took at least 3 days to complete with high monetary costs, the limitation of these compute resource is also a reason why I limit this work to aforementioned data count. Adding any more data was infeasible.

Thus, 3 LMs were trained and then fine-tuned 10 encoder-decoder combinations for summarization. This thesis argues in the light of those combinations only. While the arguments are limited to the scope of those LMs in the sequence architectures, these sequence architectures used in this work are high in number (the 10 encoder-decoder combinations) making the results very reliable at least within the framework of those 10 models. I hope that the results of this thesis will pave way for further work in the area.

# Chapter 5: Evaluation and Discussion

This chapter reports the challenges and methods of evaluation of the system performance. Research questions 4(a), 4(b), 4(c), and 4(d) are answered here.

TS evaluation may be divided into form-based evaluation and content-based evaluation (Radev et al., 2002). The former consists of different quality measures like coherence, readability and quality assessment like in (Pilault et al., 2020). The latter, on the other hand, assesses content overlap. Recall-Oriented Understudy for Gisting Evaluation (ROUGE) package was introduced by Lin (2004) and is counted as an evaluation method under the latter. ROUGE is an automatic evaluation metric for system-generated summaries that compares the total n-gram overlap between the generated summaries and the reference summaries. The higher the overlap, the better. ROUGE-1 measures unigram overlaps, ROUGE-2 measures bigram overlaps, and ROUGE-L measures the longest common subsequence (LCS) overlaps. This subsequence may not be continuous. Other forms of ROUGE have also been developed. Despite ROUGE being a popular evaluation metric for TS, human evaluation is suggested as an accurate method for assessing abstractive summaries for their coherence and saliency capturing (Droog-Hayes, 2019). Additionally, abstractive summaries can be written in ways more than one and thus, assessing the quality based on just the n-gram overlap is not an adequate evaluation strategy (Lin and Hovy (2002) as cited in Radev et al. (2002, p. 404)). Consequently, the summaries were assessed and evaluated by humans for their quality.

This chapter presents the evaluation report of the system-generated outputs through ROUGE scores and human evaluation. The first phase of evaluation stands to show that BERT encoder-based combinations are best performing combination as compared with other models. The recall is higher than precision and the precision is consistently low leading the overall F-Score to drop. This possibly be due to the long sequences generated by the system.[46] The scores are good only on the recall front. The randomly initialized encoder-decoder method combination is the model with which other models will be compared here.

## 5.1 Rouge Scores

The ROUGE F scores are poor across systems. It is in Recall, that some models perform better than the baseline. Overall, the same systems perform better than the random baseline. Rouge-1 scores[47] across different decoding methods:

**Greedy - Rouge-1:**

| Model | ROUGE 1 R | ROUGE 1 P | ROUGE 1 F |
|---|---|---|---|
| **bert2bert/greedy** | 0.265 | 0.018 | 0.033 |
| **bert2gpt/greedy** | 0.034 | 0.006 | 0.01 |
| **bert2rnd/greedy** | 0.162 | 0.015 | 0.027 |
| **bertshare/greedy** | 0.181 | 0.011 | 0.02 |
| **rnd2bert/greedy** | 0.233 | 0.016 | 0.028 |
| **rnd2gpt/greedy** | 0.03 | 0.006 | 0.009 |
| **rnd2rnd/greedy** | 0.136 | 0.012 | 0.022 |
| **roberta2gpt/greedy** | 0.039 | 0.007 | 0.011 |
| **roberta2roberta/greedy** | 0.036 | 0.007 | 0.011 |
| **robertashare/greedy** | 0.028 | 0.005 | 0.008 |

---

[46] The long length of the sequence was a function of the decoding strategy.
[47] R = Recall, P = Precision, F = F-score

*Table 22: ROUGE-1 scores for greedy decoding*

Models performing better than random in recall include bert2bert, bert2nrd, bertshare, gpt, rnd2bert; in precision include bert2bert, bert2rnd, gpt, rnd2bert; overall: bert2bert, bert2rnd, gpt, rnd2bert.

**Rouge-2:**

| Model | ROUGE 2 R | ROUGE 2 P | ROUGE 2 F |
|---|---|---|---|
| **bert2bert/greedy** | 0.056 | 0.003 | 0.005 |
| **bert2gpt/greedy** | 0 | 0 | 0 |
| **bert2rnd/greedy** | 0.044 | 0.003 | 0.006 |
| **bertshare/greedy** | 0 | 0 | 0 |
| **rnd2bert/greedy** | 0.049 | 0.003 | 0.005 |
| **rnd2gpt/greedy** | 0 | 0 | 0 |
| **rnd2rnd/greedy** | 0.025 | 0.002 | 0.003 |
| **roberta2gpt/greedy** | 0.001 | 0.001 | 0.001 |
| **roberta2roberta/greedy** | 0 | 0 | 0 |
| **robertashare/greedy** | 0 | 0 | 0 |

*Table 23: ROUGE-2 scores for greedy decoding*

In terms of Recall, BERT2BERT, BERT2RND, and RND2BERT perform better than RND2RND. The precision is poor across systems with only the BERT-based systems performing better. Overll, the BERT-based systems perform better than the RND baseline although the scores remain low.

**Rouge-L:**

| Model | ROUGE L R | ROUGE L P | ROUGE L F |
|---|---|---|---|
| **bert2bert/greedy** | 0.226 | 0.015 | 0.028 |
| **bert2gpt/greedy** | 0.031 | 0.006 | 0.009 |
| **bert2rnd/greedy** | 0.149 | 0.014 | 0.024 |
| **bertshare/greedy** | 0.143 | 0.008 | 0.015 |
| **rnd2bert/greedy** | 0.2 | 0.013 | 0.024 |
| **rnd2gpt/greedy** | 0.029 | 0.006 | 0.009 |
| **rnd2rnd/greedy** | 0.125 | 0.011 | 0.02 |
| **roberta2gpt/greedy** | 0.038 | 0.006 | 0.01 |
| **roberta2roberta/greedy** | 0.035 | 0.007 | 0.011 |
| **robertashare/greedy** | 0.027 | 0.004 | 0.007 |

*Table 24: ROUGE-L scores for greedy decoding*

The ROUGE-L scores for greedy decoding outputs indicate that in terms of Recall, BERT-encoder performs better than RND2RND except when the decoder is GPT2. The precision scores in BERT2BERT, BERT2RND, RND2BERT are slightly better than that in RND2RND. Overall, the BERT2RND and BERT2BERT systems generate better scores.

**Beam:**

**Rouge-1:**

| Model | ROUGE 1 R | ROUGE 1 P | ROUGE 1 F |
|---|---|---|---|
| **bert2bert/beam** | 0. 26 | 0.019 | 0.035 |
| **bert2gpt/beam** | 0.035 | 0.007 | 0.011 |
| **bert2rnd/beam** | 0.176 | 0.016 | 0.029 |
| **bertshare/beam** | 0.182 | 0.011 | 0.021 |
| **rnd2bert/beam** | 0.224 | 0.016 | 0.03 |
| **rnd2gpt/beam** | 0.035 | 0.007 | 0.011 |
| **rnd2rnd/beam** | 0.14 | 0.013 | 0.023 |
| **roberta2gpt/beam** | 0.046 | 0.008 | 0.013 |
| **roberta2roberta/beam** | 0.041 | 0.008 | 0.012 |
| **robertashare/beam** | 0.029 | 0.005 | 0.008 |

*Table 25: ROUGE-1 scores for beam decoding*

All models including BERT-based models have better recall than baseline. Precision for BERT-based models is only slightly better than RND2RND. Overall, the BERT-based and Robert-encoder based models perform better than rnd2rnd. However, BERT with GPT decoder, RND with GPT decoder, and Robertashare perform worse than RND2RND.

**Rouge-2:**

| Model | ROUGE 2 R | ROUGE 2 P | ROUGE 2 F |
|---|---|---|---|
| **bert2bert/beam** | 0.061 | 0.004 | 0.007 |
| **bert2gpt/beam** | 0 | 0 | 0 |
| **bert2rnd/beam** | 0.052 | 0.004 | 0.007 |
| **bertshare/beam** | 0 | 0 | 0 |
| **rnd2bert/beam** | 0.05 | 0.003 | 0.005 |
| **rnd2gpt/beam** | 0 | 0 | 0 |
| **rnd2rnd/beam** | 0.027 | 0.002 | 0.004 |
| **roberta2gpt/beam** | 0.004 | 0.001 | 0.001 |
| **roberta2roberta/beam** | 0 | 0 | 0 |
| **robertashare/beam** | 0 | 0 | 0 |

*Table 26: ROUGE-2 scores for beam decoding*

Bert2bert, bert2rnd, and rnd2bert models perform better than rnd2rnd.

**Rouge-L:**

| Model | ROUGE-LCS R | ROUGE-LCS P | ROUGE-LCS F |
|---|---|---|---|
| **bert2bert/beam** | 0.223 | 0.016 | 0.03 |
| **bert2gpt/beam** | 0.033 | 0.007 | 0.011 |
| **bert2rnd/beam** | 0.159 | 0.014 | 0.026 |
| **bertshare/beam** | 0.147 | 0.009 | 0.016 |
| **rnd2bert/beam** | 0.193 | 0.014 | 0.025 |
| **rnd2gpt/beam** | 0.033 | 0.006 | 0.01 |
| **rnd2rnd/beam** | 0.124 | 0.011 | 0.02 |
| **roberta2gpt/beam** | 0.043 | 0.007 | 0.012 |
| **roberta2roberta/beam** | 0.04 | 0.007 | 0.012 |
| **robertashare/beam** | 0.028 | 0.005 | 0.007 |

*Table 27: ROUGE-L scores for beam decoding*

In Rouge-LCS, systems that are better than random baseline include bert2bert, bert2rnd, bertshare, gpt, rnd2bert.

**Score Analysis:**

The ROUGE F-scores are poor being in the range of 0 to 3 in some cases while it is 0 in most. The systems perform poor on precision as well. However, it is in recall that the BERT-encoder-based and GPT-2-based systems have performed better when compared to the randomly initialized models. Other model combinations perform worse than the randomly initialized models. These findings are similar to those of Rothe et al. (2020) about BERT performing better than the baseline RND2RND models. Unlike their results, however, the RoBERTa Share models do not perform well. However, the overall result of the ROUGE scores indicates need for betterment.

The next sections present the qualitative assessment and human evaluation results.

## 5.2 Qualitative Assessment

The generated summaries have certain characteristics which are evaluated as per the characteristics reported in (Pilault et al., 2020) – saliency/relevance, coherent/readability. In these models, coherence is being taken as the presence of at least two rightly ordered sentential units. Some patterns observed in the summaries are as follows along with one example under every category.

### 5.2.1 RND2RND

This combination generated unrelated and incoherent outputs barring a few cases where the outputs contained few keywords or contained gold entirely.

Coherent but unrelated outputs mostly:

Example: 41

> Document: एषः भागः केन्द्रशासिते देहलीप्रदेशे अन्तर्भवति स्म । किन्तु इदानीम् देहल्याः भूभागः राज्यत्वेन परिग्ण्यते । एतत् राज्यम् नवमण्डलैः विभक्तम् अस्ति । एतेषु नवसु मण्डलेषु अन्यतमम् अस्ति पूर्वेदेहलीमण्डलम् । अस्य मण्डलस्य केन्द्रम् अस्ति प्रीतविहार ।
>
> \-\-\-\-\-\-\-\-\-\-\-\-\-\-\-\-\-\-\-\-\-\-\-\-\-\-\-\-\-\-\-\-\-\-\-\-\-\-\-\-\-\-\-\-\-\-\-\-\-\-\-\-\-\-\-\-\-\-\-
>
> Gold : भारतस्य राजधानी अस्ति नवदेहली
>
> \-\-\-\-\-\-\-\-\-\-\-\-\-\-\-\-\-\-\-\-\-\-\-\-\-\-\-\-\-\-\-\-\-\-\-\-\-\-\-\-\-\-\-\-\-\-\-\-\-\-\-\-\-\-\-\-\-\-\-
>
> rnd2rnd/greedy: ##कमपि भारतस्य बिहारराज्ये स्थितम् किञ्चन मण्डलम् अस्तिमण्डलम्[48] इति अत्र अपि सन्तिइ आचरन्तिस्य च शाढ्र नगरम् अभिनवगुप्ताध्ययनं किमी दूरेनाईःीयःस्वरेण वर्तते राज्यसभा एवप्रणीता
>
> …[rest incoherent]
>
> rnd2rnd/beam: ##महाभ भारतस्य बिहारराज्ये स्थितम् किञ्चन मण्डलम् अस्तिमण्डलम् इति आदयःसम्मेली चारुमण्डलस्य विस्तारः चकिमी मितः पठनम् किमी दूरेपरियः यदुक्तम् तत् चालुक्याः

*Result 5-1: RND2RND output*

## 5.2.2 BERT

### 5.2.2.1 BERT2BERT

This model emerged as the best performing combination of all. The following describe the characteristics. Despite some good characteristics, the summaries were not compressed well.

Captured saliency and coherence:

> Example: 75
> Document: नगरम् इदम् सिक्खधर्मस्य आध्यात्मिकम् सांस्कृतिकम् च केन्द्रम् वर्तते । भारतस्य उत्तर पश्चिमभागस्य बृहत्तमेषु नगरेषु अन्यतमम् अस्ति इदम् नगरम् । षोडशशताब्द्याम् रामदासः इति आख्येन

---

[48] Words in green refer to keywords and contextually relevant words, Words in red: Incorrect or wrongly placed words. Words in black refer to correct words that may not be very relevant to the context but are grammatically correct. Words in grey are garbage words. Portions highlighted in yellow or turquoise indicate coherent units.

चतुर्थसिक्खगुरुणा इदम् नगरम् स्थापितम् आसीत् । अस्मिन् नगरे अमृतसरोवरः स्थितः अस्ति । अतः एव अस्य नगरस्य नाम अमृतसर इति अभवत् । वर्षे रामदासस्य उत्तराधिकारिणा अर्जुनदेवेन अमृतसर नगरस्य विकासः कृतः । रामदासेन तत्र एकस्य मन्दिरस्य निर्माणकार्यम् आरब्धम् आसीत् । अनन्तरम् अर्जुनदेवेन तस्य मन्दिरस्य निर्माणकार्यम् समापितम् । ई सः तमस्य वर्षस्य विभाजनात् पूर्वम् अमृतसर नगरम् अविभाजितस्य पञ्जाब राज्यस्य व्यापारिकम् केन्द्रम् आसीत् । किन्तु विभाजनानन्तरम् अमृतसर नगरम् पाकिस्तान भारतयोः सीमायाम् स्थितम् अस्ति । अमृतसर नगरे बहूनि सिक्खधर्मस्य उपासनागृहाणि सन्ति । तेषु हरमन्दिर साहिब महत्त्वपूर्णम् अस्ति । इदम् स्वर्णमन्दिरम् इति नाम्ना सम्पूर्णे विश्वस्मिन् प्रसिद्धम् अस्ति । बिबेकसर साहिब बाबा अटल साहिब रामसर साहिब सन्तोखसर साहिब च अस्य नगरस्य अन्यानि पर्यटनस्थलानि सन्ति । अमृतसर नगरे हिन्दुधर्मस्य मन्दिराणि अपि सन्ति । दुर्गियाना मन्दिरं मातालालदेवीमन्दिरं इस्कॉन् मन्दिरं हनुमन् मन्दिरं रामतीर्थमन्दिरम् इति आदीनि हिन्दुधर्मस्य स्थलानि सन्ति ।

---------------------------------------------------------------------------------

Gold: अमृतसर नगरम् भारतस्य पञ्जाब राज्यस्य अमृतसर मण्डलस्य मुख्यालयः अस्ति

---------------------------------------------------------------------------------

bert2bert/greedy: अमृतसर नगरम् पञ्जाब राज्यस्य चण्डीगढ नगरस्य केन्द्रम् विद्यते किन्तु अस्य मण्डलस्य मुख्यालयः अस्ति अपि तु ओड़िशा राज्ये स्थितम् वर्तते एव ग्वालियर नगरे एकम् मन्दिरम् दृश्यते परन्तु अस्मिन् प्रदेशे स्थितः एकः ग्रामः प्राप्यते तस्याः तीर्थस्य उल्लेखः नास्ति न एकाः सन्ति इति नाम्ना ज्ञायते स्म तेषु भारतदेशेषु प्रसिद्धम् च भारतस्य प्रमुखग्रन्थेषु अस्याः संस्थायाः ज्ञातम् आसीत् अतः तेषाम् विमानस्थानकम् स्थितमस्ति आख्येन सह सम्बद्धम् भवति यत् हरियाणाराज्यस्यनगरम् प्राप्तम् वा भारते मानचित्रे तथा तथैव भारतदेशस्य किञ्चन राज्यम् आगतम् तथापि केरळ राज्यं उज्जैन नगरं सरलतया बसयानानि एवम् साम्प्रतम् निकटतमम् रेलस्थानकम्भ्यः इदम् क्षेत्रम् प्राप्यन्ते यतः कर्णाटकस्य सांस्कृतिकम् इत्यपि आख्ये देशेन मण्डलम् शक्यतेया नगरमिदम् अतिनामण्डलम्नदुर्गम् स्थानम् अस्तियत् मेघालयराज्ये इव बहवः जनाः हि महाराष्ट्रराज्यस्य मण्डलं किलोमीटर्मिते दूरे खलु राजस्थानराज्यम् आसन् इत्यतः अपितु मन्दिरस्य रेलस्थानकं ।

*Result 5-2: BERT2BERT output*

Some BERT2BERT contained only keywords situated far apart:

```
Gold    : भारतीय संस्कृतिः वेदमूला अस्ति
--------------------------------------------------------------------------------

bert2bert/greedy: अस्माकम् भाषा अस्ति यत् काचित् कथा वर्णिता वर्तते सा एका घटना विद्यते एवम् च क्षमा नास्ति या संस्कृतिः प्राप्यते तस्याः स्वरूपम् न भवति ते सर्वे अपि एतादृश्यः सन्ति इति उच्यते स्म तेषाम् कारणम् एव भवेत् इयम् तु सर्वज्ञता स्यात् किम् कथ्यते चेत् एषा अविद्या कथम् उपलभ्यते अस्माभिः यस्याः अर्थः कृता आसीत् यतः सर्वेषाम् नाम आत्मा अन्यथा धर्मी जानीमः खलु वैदिका प्रायोगिको भविष्यति अयम् कः चिन्तनीयम् विश्वस्य जायते इत्यर्थः माम् वा अनिवार्यम् मन्यते यद् भूतानाम् महत्त्वम् सर्वदा मानवरूपेण आत्मनः जीवनाय शक्यते योगस्य सिद्धम् येन ये पुनः चिन्तान सर्वेभ्यः अर्हस्य कार्यम् उद्दिश्य आचरिता ययारूपा उत्तरम् इत्युक्ते वेदात् सत्यम् वक्तुं शक्नोति तर्हि अस्याः पुत्रीयाः प्रतीतिनीया नाभवत्ादेः प्रतीतिः मनुकलाि ष्यन्ते अतः यः विदेशेषु नियत ।

bert2bert/beam: अस्माकम् काचित् एका भाषा अस्ति यत् गंगा सर्वज्ञता वर्तते ते सर्वे नियमाः न सन्ति इति चिन्तिताः भवन्ति एव एतादृश्यः जानीमः च भविष्यन्ति चेत् तेषाम् आवश्यकतायाः अर्थः विद्यते कः अपि उच्यते अयम् आत्मा कथम् भवेत् किम् वा स्यात् इत्यर्थः नास्ति अस्माभिः तु स्वभावो भवति तादृशं ज्ञानी उपलभ्यते यतः सर्वेषाम् वेदाय क्षमाहः जायते एवम् यस्याः कारणम् प्राप्यते सा भूतीयः भविष्यति स क्षात्रम् जनयति स्म पुनः आत्मनः नाम स्वरूपम् नियतात् अस्याः इच्छायाम् मानवस्य जीवनीयम् मन्यते यद् अन्यथा शरीराणाम् स्वीकरणे नञ्तत्पुरुषः अर्हिः येन एतावता तान् कर्तृतायाः भावनारूपेण शक्यते आहतः बाधते यया तत्त्वज्ञानस्य नियमः संस्कृतम् खलु पुरुषार्थिष्यन्ते तर्हि अस्मदीयाानाम् विचारः माम् दृश्यते तथा अलङ्कारः अहम् मातृिनी ।
```

*Result 5-3: BERT2BERT output*

Highly unrelated but coherent outputs:

> bert2bert/beam: अत्र तादृशो न अस्ति इति केचित् वदन्ति वा ते सर्वे अपि संस्काराः एव भवन्ति इत्यतः अस्मिन् एवमप्यन्ति स्म ये च पृष्टाः स्युः तेषाम् नाम अर्थद्वयम् प्रतिभावयति सः वाच्यः भवति अतः त्वं धर्म्यप्रतिध्यते चेत् तत्र कथम् भवेत् इत्यपि तस्य पदकम् स्यात् अर्थात् तत् अव्यवस्थ्यते इदम् तु वेदेतत किञ्च वाक्यात् चेत्यत्रयम् मन्यते स कर्मैव अर्थः क्रियते तावत् यो व्यक्त्वायते एवम् यस्मिन् क्षत्रियत्वात् अन्यथार्थे सति यथा द्रव्यमस्ति तथा काम्ठी असिहमेत्नते यतोच्छिन्नम् यदिन्दृषिति तत तर्हि अयम् परेङ्गतम् इत्यर्थः नात्राशयेन तथाऽस्ति तस्मिन् निष्प्राप्तिः ।

BERT models although coherent, generated hallucinations. In the following example, the summary conveys information totally contrary to the fact:

> Example: 2
>
> Document: अतः वातजैः रोगैः पित्तजैः रोगैः च पीड्यमानानां वातप्रकृतियुक्तानां पित्तप्रकृतियुक्तानाम् च पित्तकाले देशे च अस्य पिण्डखर्जूरस्य उपयोगः हितकरः ।
>
> ---
>
> Gold: एषः पिण्डखर्जूरः वातम् पित्तम् च हरति
>
> ---
>
> bert2bert/greedy: एतत् पित्तम् वातकम् च वर्धयति […]

*Result 5-4: BERT2BERT output*

However, in rare cases where facts from gold summary were present (the entire gold summary was present), the summary was long and incoherent after a point:

> Gold : कर्णाटकस्य अष्टाविंशतिलोकसभाक्षेत्रेषु अन्यतमम् अस्ति बीदरलोकसभाक्षेत्रम्
>
> ---
>
> bert2bert/greedy: कर्णाटकस्य अष्टाविंशतिलोकसभाक्षेत्रेषु अन्यतमम् अस्ति बीदरलोकसभाक्षेत्रम्विधानसभाक्षेत्रम्बेङ्गळूरुलोकसभाक्षेत्रम् दक्षिणत्रिपुरामण्डलम् बेङ्गळूरुनगरमण्डलम्लोकसभाक्षेत्रे अन्तर्भवति …[incoherent henceforth]

*Result 5-5: BERT2BERT output*

### 5.2.2.2 BERT2RND

Like other BERT-based models, this model also generated long summaries. The outputs mostly were coherent only to up to some level:

> Gold : अमृतसर नगरम् भारतस्य पञ्जाब राज्यस्य अमृतसर मण्डलस्य मुख्यालयः अस्ति
> ----------------------------------------------------------------
> bert2rnd/greedy: <mark>तवाङ्ग हरियाणा राज्यस्य राजधानी अस्ति मण्डले स्थितम्</mark> विरम नगरम् च सन्तिघल विद्यतेयाः मण्डलस्य मुख्यालयः वर्तते स्म इति ग्वालियर नगरे एव अस्य संचार अयस् सहितम् कण्ठः दूरे क्षेत्रम् न केवलम्नम्स्थलम् ारी स्थलम् अजी केन्द्रम्समह एतदर्थम्  श्रुमण्डलस्य विस्तारः चतुरस्रकिलोमीटर्मितः [...rest long and incoherent]

*Result 5-6: BERT2RND output*

Some outputs were contained the entire gold and that was the only coherent segment:

> Example 3:
> Gold : कर्णाटकस्य अष्टाविंशतिलोकसभाक्षेत्रेषु अन्यतमम् अस्ति बीदरलोकसभाक्षेत्रम्
> ----------------------------------------------------------------
> bert2rnd/greedy: <mark>तवाङ्ग कर्णाटकस्य अष्टाविंशतिलोकसभाक्षेत्रेषु अन्यतमम् अस्ति बीदरलोकसभाक्षेत्रम्</mark> इति आदीनि अधिवर्षम् अत्युत्क्षेत्रम् ानि भवन्तिमार्गे न अक्टसंहिताबेङ्गळूरुलोकसभाक्षेत्रम्सम्मेलनस्याईजी तमे वर्षे आस्त चभवं मिर बाईसत्त अनुरागाः सन्तिजलेन विरम स्थापितःदर्शी योगस्य मनोर वर्तते स्थितेन लोपामु राष्ट्रपति अभवत् कार्यकर्तृ वाग राजधान अभिज्ञान भवति पाञ्चालीषु निवारणम् पाण्डित्यपुरमण्डलेम् ोरेशन् कौत्सोक्ष चतुर्थ चकिमी मितःत्सहसमूह ध्यान इमांभ्यःचेतरणारीक्ष कण्ठितेन चोळ भागो मेट्रो कुर्वन्स्ट पदे स्थूल नियुक्तवान्ः ः निवेद तत्पश्चात् दलं भर्त दृश्यमानेषु प्रमुखनक्षत्रसमूहेषुभावे सागरतीरे तृ सर्वत्रशुद्धिः येतत् पल्लव पाणिनापि वर्णनोत्मानः स्म लेखक संस्कृतसाहित्य चतुष्षष्टिः अपराधःतः किमी दूरे स्थितम् किञ्चन राज्यम्शुल्कम् अजीव राज्यस्य साक्षरतामानम्मेण्ट निश्र्चयेन पादाः भविष्यन्ति शिरोवेदनाध्य मानवाःसंहितायां पश्यामि पुत्रःक्षेपः विक्रमादित्यघाते कुष्ठरोग ।

*Result 5-7: BERT2RND output*

As observed above, hallucinations could be commonly found:

> Example 23
> Gold: भारतदेशस्य किञ्चन राज्यम् अस्ति आन्ध्रप्रदेशः
> ----------------------------------------------------

| |
|---|
| bert2rnd/greedy: तवाङ्ग भारतदेशस्य किञ्चन राज्यम् अस्ति उत्तरप्रदेशराज्यम्घल […] |
| bert2rnd/beam: तवाङ्ग भारतदेशस्य किञ्चन राज्यम् अस्ति उत्तरप्रदेशराज्यम् सस्यानि |

*Result 5-8: BERT2RND output*

### 5.2.2.3 BERTShare

BERTShare generated coherent outputs but with frequent number of more irrelevant words inserted between coherent units. Yet, coherence was present in most outputs:

| |
|---|
| Example: 75 |
| Document: नगरम् इदम् सिक्खधर्मस्य आध्यात्मिकम् सांस्कृतिकम् च केन्द्रम् वर्तते । भारतस्य उत्तर पश्चिमभागस्य बृहत्तमेषु नगरेषु अन्यतमम् अस्ति इदम् नगरम् । षोडशशताब्द्याम् रामदासः इति आख्येन चतुर्थसिक्खगुरुणा इदम् नगरम् स्थापितम् आसीत् । अस्मिन् नगरे अमृतसरोवरः स्थितः अस्ति । अतः एव अस्य नगरस्य नाम अमृतसर इति अभवत् । वर्षे रामदासस्य उत्तराधिकारिणा अर्जुनदेवेन अमृतसर नगरस्य विकासः कृतः । रामदासेन तत्र एकस्य मन्दिरस्य निर्माणकार्यम् आरब्धम् आसीत् । अनन्तरम् अर्जुनदेवेन तस्य मन्दिरस्य निर्माणकार्यम् समापितम् । ई सः तमस्य वर्षस्य विभाजनात् पूर्वम् अमृतसर नगरम् अविभाजितस्य पञ्जाब राज्यस्य व्यापारिकम् केन्द्रम् आसीत् । किन्तु विभाजनानन्तरम् अमृतसर नगरम् पाकिस्तान भारतयोः सीमायाम् स्थितम् अस्ति । अमृतसर नगरे बहूनि सिक्खधर्मस्य उपासनागृहाणि सन्ति । तेषु हरमन्दिर साहिब महत्त्वपूर्णम् अस्ति । इदम् स्वर्णमन्दिरम् इति नाम्ना सम्पूर्णे विश्वस्मिन् प्रसिद्धम् अस्ति । बिबेकसर साहिब बाबा अटल साहिब रामसर साहिब सन्तोखसर साहिब च अस्य नगरस्य अन्यानि पर्यटनस्थलानि सन्ति । अमृतसर नगरे हिन्दुधर्मस्य मन्दिराणि अपि सन्ति । दुर्गियाना मन्दिरं मातालालदेवीमन्दिरं इस्कॉन् मन्दिरं हनुमन् मन्दिरं रामतीर्थमन्दिरम् इति आदीनि हिन्दुधर्मस्य स्थलानि सन्ति । |
| ------------------------------------------------------- |
| bertshare/greedy: भारत देशे देशस्य स्थितम् अस्ति इदम् यत् मण्डलम् प्रसिद्धम् वर्तते इति एतत् मन्दिरम् क्षेत्रम् अपि एकम् स्थानम् नाम आसीत् विद्यते तथापि अस्य अत्र ख्यातम् एव मन्यते स्म अस्मिन् हमारे ज्ञायते |

यद् यद्यपि उक्तम् दृश्यते भवति किन्तु परन्तु इदं नाम्ना स्थलम् अतीव वर्णितम् कथ्यते इत्यतः अतः न प्राप्यते इत्यपि यस्मिन् विषये एवम् केवलम् उल्लिखितः सन्ति आसन् यतः तत् वयम् प्रमुखम् कार्यम् तु जातम् अस्तियत् यतोहि भारते भारतस्य भारतदेशस्य महत्त्वम् उल्लेखः ज्ञातम् नास्ति येन अनेन सह आदिषु निर्मितम् अस्माकम् नगरम् स्वाभाविकम् अभवत् अयम् एषः सः एकः निकटतमम् एतदतिरिक्तम् किम् आगतम् जानीमः चेत् यत्र उद्घोषितम् प्रतीयते अत अवदत् उक्तवान् आहारपदार्थः साम्प्रतम् च तेन भगवतःमन्दिरम् अनुभवति तर्हि पृष्टम् प्राप्तम् भवेत् तावदेव प्रख्यातम् यदि यदा नूनम् शक्यते शक्नुमः तदा तदैव इदानीम् अलिखत् पृष्टवान् तस्य सर्वेषाम् आचरन्तिस्ति तावत् स्पष्टम् कृतः ।

Result 5-9: BERTShare output

### 5.2.2.4 RND2BERT

**Positive Aspects:** This combination generated highly coherent but very long summaries with many instances of summaries containing keywords including information which was close to the facts contained in the document. Interesting was the case where the summary contained the entire gold:

Example: 41
Document: एषः भागः केन्द्रशासिते देहलीप्रदेशे अन्तर्भवति स्म । किन्तु इदानीम् देहल्याः भूभागः राज्यत्वेन परिगण्यते । एतत् राज्यम् नवमण्डलैः विभक्तम् अस्ति । एतेषु नवसु मण्डलेषु अन्यतमम् अस्ति पूर्वदेहलीमण्डलम् । अस्य मण्डलस्य केन्द्रम् अस्ति प्रीतविहार ।

--------------------------------------------------------------

Gold: भारतस्य राजधानी अस्ति नवदेहली

--------------------------------------------------------------

bert2bert/greedy: भारतस्य राजधानी अस्ति देहली नगरम् आसीत् भारतदेशस्य स्थितम् किञ्चन राज्यम् वर्तते जम्मूकाश्मीरराज्यम् आन्ध्रप्रदेशः च मण्डलम् विद्यते इति अत्र प्रसिद्धम् एकम् क्षेत्रम् प्राप्यते यत् अस्य मण्डलस्य

केन्द्रम् स्थितमस्ति भारतम् भारते सह सम्बद्धम् अभवत् परन्तु देशस्य बिहारराज्ये स्थितः एकः मण्डलः दृश्यते न तु महाराष्ट्राज्यस्य सम्पर्कः अपि अन्यतमम् भवति स्म एव हिन्दुमण्डलेष्षु अन्यतमा स्यात् इत्यपि प्रख्यातम् वर्णितम् उपलभ्यते किन्तु भारतराज्ये निर्मितम् कथ्यते एवम् आगतम् उच्यते तस्य उल्लेखः ज्ञातम् नास्ति अस्मिन् मण्डले उक्तम् सन्ति तथापि ५०४८ मण्डलेषु इतिहासः निकटतमम् रेलस्थानकम् अकरोत् हमारे बस्ना नाम्ना ज्ञायते भारतदेशे स्थितं कश्चन ग्रामः तर्हि वर्गः अयम् विमानस्थानकं उल्लिखितम् अस्ति यत् हरियाणा राज्यस्य प्रमुखे किलोमीटर्मिते दूरे विद्यमानं एकं दक्षिणे प्रतिचतुरस्रकिलोमीटर्७ नगरे मण्डलानि सस्यजन्यः शक्यते आयुर्वेदः प्रदेशे भवेत् तस्मिन् काले जनसङ्ख्यासान्द्रता विराजते चेदपि घोरीनगरस्य पश्चिमे अथवा नेपालदेशस्य ने ख्यातम् विद्यमानः४ ।

*Result 5-10: RND2BERT output*

In capturing keywords or related words, greedy performed better in some cases:

Example: 57

Document: तत्र सूक्तम् चतुर्विधम् ऋषिसूक्त देवतासूक्तच्छन्दःसूक्त अर्थसूक्तभेदात् । एकर्षिदृष्टमन्त्राणाम् समूहो ऋषिसूक्तम् । एकदेवताकमन्त्राणाम् समूहो देवतासूक्तम् । समानछन्दसाम् मन्त्राणा समूहो नामच्छन्दःसूक्तम् । यावदर्थसमाप्तानाम् मन्त्राणाम् समूहः अर्थसूक्तम् । सुष्ठूक्तत्वात्सर्वम् सूक्तमित्याख्यायते ।

---

rnd2bert/greedy: अत्र हि काव्यदृष्ट्या तु शास्त्राणाम् महत्त्वम् अस्ति किन्तु अस्मिन् ग्रन्थे सर्वत्र दृश्यते तथा एव च तत्त्वे अपि सर्वशास्त्रशास्त्रस्य सूत्रस्यैव मन्यते यथा ह्येवत इति सूत्रकारः वर्तते स वेदो न केवलम् शब्दद्वयम् वा यदस्ति तत् त्रिविधत्वम् तस्मादपि नाम यस्मिन् परेनप्यनुशासनस्य अर्थनामुपरि भवति तत्र अयम् खलु अलङ्कारशब्दस्य अर्थः एवम् अतिरिषिः स्यात् यतो यो वेति कथ्यते तेन त्वर्थी यत्र अन्यथा धर्मार्थमुत्तात् पूर्वाहकः यः किम् इदम् कश्चिद्वक्त्रकारानुसारेण वाक्यदर्शनार्थः चतुर्धा देवानांनाटकाद् येन ह्यस्यस्थिति वाऽस्माकं तत सः त्वम्नम् चात्मिका इत्यत्र पदत्वा तावुपर्वत्य ।

rnd2bert/beam: इति अनेन एव ज्ञायते यत् अस्मिन् ग्रन्थे अपि वर्णो दृश्यते किन्तु अत्र न तु सर्वशूराणाम् व्यवहारानाम् महत्त्वम् च अस्ति यतो हि भासम् भवति यदस्ति तत् तत्त्वार्थत्वेन प्रतिपादितम् वर्तते यत्र तत्र वेदकाला चात्मिकाअऽयम् तस्मादेतदर्थस्य सवर्णदीर्घसन्धिः नाम अर्थद्वयम् वा तथाऽप्यनुयायिनः यथा धर्मलोप्भावनम् एवम् यत्ने संस्कारते

तथा तेषाम् स्मरणीयम् त्वमुपस्मादेव सामान्ठकनाटकत्वसम्पादनी ग्राहुष्णात् तत ण्थिकसूत्रसूत्रम् यस्मिन् व्यक्तामनलङ्कारत्वम् चेतर्वतध्ययनेन वेति **सूत्रानामपि** यः योकुत्र्यस्माकंऽयत्वाम् कारकोपस्यावरुपिधि ।

*Result 5-11: RND2BERT output*

**Negative Aspects:** Rare cases of totally unrelated outputs were also present:

Gold: पूर्णसरस्वती महोदयेन विरचितम् नाटकम् भवति कमलिनीराजहंसम्

----------------------------------------------

rnd2bert/greedy: अत्र भगवान् महावीरः उपगम्यवान् आसीत् इति आदिषु राज्येषु सर्वेषु प्रदेशेषु अपि अन्यतमः भवति स्म एव अस्य अर्थः विद्यते इत्यतः सः हि योस्मित इत्यपि कथ्यते स तु न अस्ति किन्तु तेन ते सर्वे जनाः ये च मार्गेषु नवान् प्रतिताम् कुर्वन्ति एवम् यस्मिन् समये सर्वत्र स्विद्यादिभिः सह तस्य पूर्वसीमायाम् नूतनो भवेत् तथा तेषाम् विषये अस्माकम् जीवनकाले अभिनये सति कः किम् वा यदि वयम् संस्कारम् विना विस्म्वयपि तान् धर्मतया कल्पा इव दृश्यते तत् कथम् प्रायः अतिदेशैः संष्यते यत् केवलम् तत्त्वेन यथा प्रत्यब्धाम् विधीयते अस्माभिः तथापि तदा तर्हि तदैव अप्रयन्तिन्ते ततिषयेन कारणेन मम स्थानम् भविष्यति यतोसु ।

rnd2bert/beam: अत एव स हि उसी इति आदयः जनाः निवसन्ति स्म किन्तु अस्मिन् विषये न एकेताः आसन् तथापि ते सर्वे अपि केचन विषयाः मन्यन्ते ये च तेषाम् विशेषतया कल्पे भवन्ति तथा तु सर्वशूरो भवति यथा धर्ममयाः सन्ति तेषु वेदकाला इव दृश्यते तत् पूर्वसीमायाम् एवंविधस्य वा व्यवहारावस्थाङ्कता भवेत् यतोऽयम् कथम् नाम पुनातिप्यनेकते एवम् केवलम् अतिदेशिद्यामपि अन्यथा जलादनक्षात्मिका पुष्मः यस्मिन् समये अन्ये नाभ्दि तत केन भासम् प्रति आहंत्रवर्तिना सह चेतीक्षिता यत् हेषित किलेन गृह्णवरानुसारेण मम राष्ट्रात् योवोयते ।

*Result 5-12: RND2BERT output*

### 5.2.3 GPT-2

The outputs are incomprehensible and mostly incoherent. There is hardly any readable text produced across decoding methods.

---

Example 37:

Gold: द्वादश ज्योतिर्लिङ्गानाम् आवल्याम् घुश्मेश्वरः द्वादशम् ज्योतिर्लिङ्गम् अस्ति

------------------------------------------------------------

bert2gpt/greedy: भारतदेशः प्रियोगीतिः तु बहुनां रंजषणव न असृजत परन । तमे वर वैद्किकयोः सँसृहिंहसौधाः जाँतक गोंढ च सम मण्डलैः मैंगल महॉड द़ेहलीं लेंगकुंजर बोक नगरम इतिं शौंडल भाम खूंझ कुँच घ झूँघ थ ह़ैई और हज़ी फ़ोइ नही। एक अंगद पर ए टंपूः ♦

roberta2gpt/greedy: कार्तिकमूलेयमुनिः शैलपः रौगीपृष्ट् कांशींबिंगपोषकाः मुंगलैः पं चन ज़ागर इतिं तमिळ च न स्योज़ वाँस द गोंड़ी हॉङ सरफ अरेंझ कलेद महा ॆदयम । तीः तथ एव असँख जनेः धथोः लकनतुः अथवण बैंकण इव वर भूः यूंवर्ः परन तदवल लृंह नगरमध उत फल पशवन हॉरिं शबघ ♦

bert2gpt/beam: भूमेः गोलाकृतिः प्रमुखविशेषताः च । ॆः अतीव सुःखं शौरीः मांसवैः वैदेहीं हैंदर अलिंग ज़ियाँबॉज़ बोंगल तुंझान और महोदय्ः फ़्फ़ोस खूंड़ी घँट ऑफ टॅबल कुँच २०१५ अहमद देंगयोः सह न एक नगर लुङ सरलनवर इव असृध्ः रा ॆचक उत इन उल०प कवूः मरूँगस ♦

*Result 5-13: GPT-decoder based outputs*

---

Similar was the outcome for GPT-2 decoders. An example of each is quoted here:

---

Gold: जम्बूद्वीपस्य पूर्वविदेहक्षेत्रे सुसीमा नामिका नगरी आसीत्

------------------------------------------------

---

> roberta2gpt/greedy: <mark>राज्ये जनाः अपि पुनः भूमेः संवृतिः अभवत ।</mark> रोगगीतमौच इतिं आसन न ज्ञानन्द गांधी महोदयैः उक मैथिलं काँग देश वीं शॉफ चर खोंतर पद तुंवर सम बैंक अन च अभूंदतुः तथष्ण वर मण् नगरें हूँटॅं धँब्रिप उत लाेंह भट घोड़े मन रह थाेंघ कप०लक कवी। एक शरीः मर और

> roberta2gpt/beam: <mark>राज्ये जनाः अपि पुनः भूमिः इव मृतां तीथीः च न संशोधनं</mark> कुंवरूः इतिं । वैदुः मतैः सह योग दृढ़सिंह गौतमीं शोंगल महाेंदय हॉफ़िक उत खाँड़ा बैंकर टेंगड़े आज्ञ ड़ो ज़ैख़ीब घूंट ऑफ फॅल धौंजर अशृंखल कपुँच जय तक थूँगन पटेचर एल ए अभ्ेंतर चर

> rnd2gpt/greedy: भारतस्य पूरीवरेखाः नृतः गुजरांतर रोगिणोः दुःखसंयनिः विंशतेः वैरयर्स्हितसन्ः सौमींप्रदेशैः सह जॉब मॉल भोंगल महोदयसष्षाट जनीः च आकृथवा् तुं शृंङ्खलाा घूंजँ इतिं अपिंच पतकॅ कैंद धेंगड़ी बाेंड़ेयर अमैझ नगर ड़ोसी० मधुँञ ट०ध

*Result 5-14: GPT-2 based outputs sample 2*

### 5.2.4 RoBERTa

Contrary to expectations, RoBERTa outputs were incoherent:

> Gold : पूर्णसरस्वती महोदयेन विरचितम् नाटकम् भवति कमलिनीराजहंसम्
>
> -------------------------------------------------
>
> roberta2roberta/greedy: रविद्यासंवतीसरेण हृदयः परोपकेषु अनूनमांशिः इतेः आदौ सिंह रैज महोदयसनोः काँच वैः सह शँक च न असगथाः सनतध एव दुःखीः पुंसवनस्कराेंप अप यनूंतम इव वऋततड गृंहमञ आदय बहव मन ज़िङ्कण मण एक गण बेंरव मॉम ख़ादर यचम उच ३छ दश कथ पशचन करछनवनभृतय सद

> roberta2roberta/beam: रहिमालयपर्वतः पुणेयवंशीयैः सह सोमवैसराः च असूकृतिः । वेद इतेः आदौ एव मांस गँधोः कोचुं शिंशत दुःखतमूंदप ३ १ हॉदर बाँग ख़ान मह इव ताःेड न एक गण परीः अञ्चिज्छ रेंव मऩोजय यंथ अ आभूः अनुभव भव समुप ४ दश कथञबृंजत आदय बहव आसन सन्सवयमाङ❖

*Result 5-15: ROBERTA-based outputs*

Generally, outputs contained incoherent summaries with no keywords:

> Example: 67
>
> Document: तस्याः नगर्याः राजा धनपतिः आसीत् । भगवान् अरनाथः पूर्वजन्मनि धनपतिः नामकः राजा आसीत् । धनपतिना तस्मिन् जन्मनि धर्मस्य साधना कृता । तेन बहुवर्षाणि यावत् राज्यसञ्चालनम् अपि कृतम् आसीत् ।
>
> --------------------------------------------------------
>
> Gold: जम्बूद्वीपस्य पूर्वविदेहक्षेत्रे सुसीमा नामिका नगरी आसीत्
>
> --------------------------------------------------------
>
> robertashare/greedy: लाज़् रोमे निपुरः कृषिः आसीतदौ तसाः । दैवयून इतां आदेच भंसिंहहस च सैः य वॉक बोड़ ख़ाँई मह जोंब प़िथ थेंख लैंट ग हँझी इन इऱ्फ़े और यहोइन अब अलूंएओ ओर परींग एक नगर म डँइली। तक बन पत मन ए ई अहमद आज महल अवदत अभवत अस सन आसन त अनुः अपृण एव सह समोः❖
>
> robertashare/beam: राज्ये तुथिपी भोमः । ःृतौ सं जैलूकाः च यैः सह किंषिः एव देववतां न एकोंच राँईग गुंहोः वँड़ाफ़् खॉःेःण इति पेरन मोलझ्री ए ई ओऱ्ओएओ और यहूंस ब्रैइन इन लॉइड़े एस एच ऑफ ३ १ ४ २ अ आ ओ ऐ इ॰०एस अफ टी॰आई आफ फ़ुइल महल एन परींद अहमद अलौंब❖

*Result 5-16: ROBERTA-based outputs*

## 5.3 Human Evaluation

4 human evaluators were asked to rate the systems generated outputs. Each human evaluator had formal training in Sanskrit with their educational qualifications ranging from bachelor's to doctorate degrees. Each evaluator was given two sets of 5 randomly selected source texts with 20 summaries of focus each. Each evaluator read a total of 100 summaries in a set. They evaluated each summary for each of the three qualities:

i. Coherence and Readability

ii. Factual Consistency

iii. Keyword Capturing

iv. Overall Quality

The evaluation was held on a scale of 1 (poor) to 5 (excellent). Additionally, based on the methodology of Rothe et al. (2020), the evaluators rated summaries from -1 (Worst) or 1 (Best). The rank for a system was decided on the number of times a system was rated the best minus the number of times it was rated the worst.

### 5.3.1 Scaled Ranking

**Coherence and Readability:**

| Model | Ranked Very Good or Excellent (Out of 20) | Ranked Good or Poor (Out of 20) |
|---|---|---|
| bert2bert/greedy: | 11 | 9 |
| bertshare/greedy: | 5 | 15 |
| bert2gpt/greedy: | 0 | 20 |
| roberta2roberta/greedy: | 1 | 19 |

| robertashare/greedy: | 0 | 20 |
| roberta2gpt/greedy: | 0 | 20 |
| rnd2rnd/greedy: | 4 | 16 |
| bert2rnd/greedy: | 5 | 15 |
| rnd2bert/greedy: | 8 | 12 |
| rnd2gpt/greedy: | 0 | 20 |
| bert2bert/beam: | 8 | 12 |
| bertshare/beam: | 6 | 14 |
| bert2gpt/beam: | 0 | 20 |
| roberta2roberta/beam: | 0 | 20 |
| robertashare/beam | 0 | 20 |
| roberta2gpt/beam: | 0 | 20 |
| rnd2rnd/beam: | 3 | 17 |
| bert2rnd/beam: | 3 | 17 |
| rnd2bert/beam: | 6 | 14 |
| rnd2gpt/beam: | 1 | 19 |

*Table 28: Human evaluation of coherence and reaability*

**Factual Consistency**

| Model | Ranked Very Good or Excellent (Out of 20) | Ranked Good or Poor (Out of 20) |
| --- | --- | --- |
| bert2bert/greedy: | 8 | 12 |
| bertshare/greedy: | 4 | 16 |
| bert2gpt/greedy: | 1 | 19 |
| roberta2roberta/greedy: | 1 | 19 |
| robertashare/greedy: | 0 | 20 |
| roberta2gpt/greedy: | 0 | 20 |
| rnd2rnd/greedy: | 4 | 16 |
| bert2rnd/greedy: | 5 | 15 |
| rnd2bert/greedy: | 6 | 14 |
| rnd2gpt/greedy: | 0 | 20 |
| bert2bert/beam: | 5 | 15 |
| bertshare/beam: | 3 | 17 |
| bert2gpt/beam: | 0 | 20 |
| roberta2roberta/beam: | 0 | 20 |
| robertashare/beam | 0 | 20 |
| roberta2gpt/beam: | 0 | 20 |
| rnd2rnd/beam: | 2 | 18 |
| bert2rnd/beam: | 4 | 16 |
| rnd2bert/beam: | 5 | 15 |
| rnd2gpt/beam: | 1 | 19 |

*Table 29: Human evaluation of factual consistency*

**Keyword Capturing**:

| Model | Ranked Very Good or Excellent (Out of 20) | Ranked Good or Poor (Out of 20) |
|---|---|---|
| bert2bert/greedy: | 8 | 12 |
| bertshare/greedy: | 4 | 16 |
| bert2gpt/greedy: | 1 | 19 |
| roberta2roberta/greedy: | 1 | 19 |
| robertashare/greedy: | 0 | 20 |
| roberta2gpt/greedy: | 0 | 20 |
| rnd2rnd/greedy: | 4 | 16 |
| bert2rnd/greedy: | 5 | 15 |
| rnd2bert/greedy: | 6 | 14 |
| rnd2gpt/greedy: | 0 | 20 |
| bert2bert/beam: | 5 | 15 |
| bertshare/beam: | 3 | 17 |
| bert2gpt/beam: | 0 | 20 |
| roberta2roberta/beam: | 0 | 20 |
| robertashare/beam | 0 | 20 |
| roberta2gpt/beam: | 0 | 20 |
| rnd2rnd/beam: | 2 | 18 |
| bert2rnd/beam: | 4 | 16 |
| rnd2bert/beam: | 5 | 15 |
| rnd2gpt/beam: | 1 | 19 |

*Table 30: Human evaluation of keyword capture*

### 5.3.2 Best-Worst Rating

Models with positive scores are indicated in green:

| Model | Ranked Best (n) | Ranked Worst (n) | Score |
|---|---|---|---|
| bert2bert/greedy: | 13 | 7 | 6 |
| bertshare/greedy: | 8 | 12 | -4 |
| bert2gpt/greedy: | 6 | 14 | -8 |
| roberta2roberta/greedy: | 4 | 16 | -12 |
| robertashare/greedy: | 1 | 19 | -18 |
| roberta2gpt/greedy: | 3 | 17 | -14 |
| rnd2rnd/greedy: | 6 | 14 | -8 |
| bert2rnd/greedy: | 9 | 11 | -2 |
| rnd2bert/greedy: | 10 | 9 | 1 |
| rnd2gpt/greedy: | 1 | 19 | -18 |
| bert2bert/beam: | 11 | 9 | 2 |
| bertshare/beam: | 7 | 13 | -6 |
| bert2gpt/beam: | 5 | 15 | -8 |
| roberta2roberta/beam: | 3 | 17 | -14 |
| robertashare/beam | 2 | 13 | -11 |
| roberta2gpt/beam: | 5 | 15 | -10 |
| rnd2rnd/beam: | 7 | 13 | -6 |
| bert2rnd/beam: | 9 | 11 | -2 |

| rnd2bert/beam: | 10 | 10 | 0 |
| rnd2gpt/beam: | 5 | 15 | -10 |

*Table 31: Human evaluation of best-worst performing models*

The BERT-based models are consistently ranked slightly better than others on all the three characteristics. Factual consistency remains low across models with fewer than 5 instances of excellent or good ratings. The keyword capturing ability of BERT-based model was ranked excellent in 8 cases. Although the score for keyword capturing ability is still less than 10, i.e., less than 50% of the total ranking rounds (20), it indicates a positive trend and can be improved in the future.

## 5.4 Concluding Analysis

In this section, I seek to corroborate model performance with evidence from literature while trying to ascertain the reason behind different model performances. The research question 3.c) Why did certain systems perform better? and 3.d) how can summaries be improved in the future are the focus of the following concluding analysis:

### 5.4.1 BERT

Rothe et al. (2020) conclude that BERT encoder setups perform the best for English document summarization tasks. In the context of SATS, that claim is equally true. The BERT setups have particularly been useful in generating coherence. The biggest drawback, however, remained in factual consistency and hallucination. Conversely, as the previous section shows, BERT-based setups have performed well in capturing themes and keywords.

**5.4.2 GPT-2**

GPT-2 is a decoder-only architecture (Tunstall et al., 2022). Rothe et al. (2020) report that GPT-2 can be a good extractor than most encoder-decoder setups but it does not abstract well (p. 270). However, in this thesis, I did not train decoder-only architecture for ATS while I did use GPT-2 in combination with other models. For the given SATS task, the combinations involving GPT-2 as decoder performed poorly on both coherence and keyword capturing. The generation, in fact, rarely had any coherent units or words.

**5.4.3 RoBERTa**

Scholars have noted that RoBERTa performs poorly in ILs and such poor performance may be due to two reasons. First, lack of long training. RoBERTa is expected to perform better on longer training on bigger datasets (Jain et al., 2020, p. 5). Second, the BPE-tokenization method used by RoBERTa may not have been suitable for the morphologically rich languages (Jain et al., 2020). Rothe et al. (2020) find RoBERTa to be a good performing model, although the same does not hold true for Sanskrit. In other words, while RoBERTa is a good model for English on certain categories of summarization tasks, it may not suit ILs yet. This thesis supports the above observations from the data. RoBERTa model performed poorly across different test data sizes with rare exceptions, if any.

Final two observations about the models are: first, none of the systems proves to be a good compressor. CR is high and must be curbed in the future. Second, most summaries are found to hallucinate. Factual consistency is very rarely present even in the BERT-based models.

In sum, BERT-based models seem promising for future research in SATS. With more data and better compute resources, the model combinations may be used for future research. PTLMs for SATS thus do have potential to produce not only coherent but apt summaries which also contain key information of the source text. Other Transformer architectures such as BART or T5 may also be adapted. In the future, more data may be incorporated to improve model performance. Alternately, training on a related task or added layers to augment performance should be undertaken (N. F. Liu et al., 2019). The compression rate must also be improved in future work. Supervised methods like the OpenNMT could be used for SATS (Klein et al., 2017). OpenNMT is an open source neural MT which has been used for ATS also treating ATS as a parallel supervised learning task with no prior training (Gehrmann et al., 2018; Klein et al., 2017). Similar approach may be tested for SATS later.

## 5.5 Limitations

Around the same time as this research work is being finalized, some updates in the Indic NLP sphere have taken place which could not be evaluated for SATS in this thesis but should be considered for future work. I had noted the release of Z-Code++ model for ATS algorithm earlier as well as the decoder-only architecture for pre-training both of which I do not evaluate in this work (He et al., 2022; Khandelwal et al., 2019). Two other developments are being presented here:

First, IndicBERT v2 is an MLLM including Sanskrit corpus taken from Wikipedia and OSCAR which claims to outperform MURIL (Doddapaneni et al., 2022). However, since it is an MLLM and its release comes around the same time as the completion of this work, I leave its evaluation for ATS for further attempts.

Second, a recent update in TS for agglutinative languages like Turkish has indicated that multilingual BERT may achieve better results than mono-BERT for ATS (Baykara & Güngör, 2022). In this research work, MURIL was tested for summarization with no standard results (Appendix F). But that may have happened due to resource constraints. With Baykara and Güngör (2022)'s work, MLLMs could seem like a positive solution suggesting that MLLMs could be tested in a future work.

# Chapter 6: Conclusion

The key goal of this thesis has been to trace the issues and challenges in training summarization models for SATS. Since this is the first work in SATS, the key contribution of this work has been initiating work in SATS and training 10 summarization models based on Transformer architectures and reporting the challenges therein. This work commenced in search of research questions across four different themes the answers to which it presented throughout its different chapters.

The literature review indicated that Transformers are the current state of the art models for various NLP tasks. Since DL require huge datasets for training, high-resource languages are usually at the forefront witnessing many models and datasets. Low-resource languages, however, lack large-scale datasets as well as trained models. Sanskrit, being a low-resource language, faced the same problems. However, this work used the popular solution of transfer learning approach through pre-trained language model to initiate SATS. This work has initiated SATS. Given the low-resource status of Sanskrit, finding resources in Sanskrit was the first challenge that this work faced. However, using the available sources to arrive at a prospective SATS system is the first contribution of this work.

The second contribution of this thesis has been presented in the data preparation and training chapter through which this work has contributed insights into the challenges of data preparation, tracking sources, and cleaning the data. This work focused on contemporary Sanskrit prose owing to the availability of data in the field and the short sentence and word lengths. The splitting of *sandhi*/*saṃyoga* patterns and the limitations were put to the split patterns are important points to be noted. Only the long *sandhi* words in journals were split. The total and unique word counts underwent change after this *Sandhi* split.

A third contribution this thesis makes is training Transformer-based language models and summarization models for Sanskrit abstractive text summarization training which was the focus of the chapter on training and results. This contribution is important because of the low-resource nature of Sanskrit. Despite the complete absence of any huge datasets for SATS, this work developed a parallel dataset for pretraining and finetuning which could take place because of the resources offered by the Google Colab/GCP platforms and the immensely helpful open-source HuggingFace DL library**.** The language models were trained with varying perplexity scores but the pretraining objectives gave good results.

10 different combinations of encoder-decoder checkpoints were trained for the summarization task with early stopping. The ROUGE F-scores indicated the need for more improvement. However, human evaluation indicated that the generated texts were readable and had some encouraging keyword-capturing abilities.

The BERT2BERT model performed well by capturing essential terms and salient information. Earlier researchers in English found it to be a good summarizer as well. Other models, however, performed poorly. One reason for the performance of BERT2BERT model and poor performance of other models may be attributed to the tokenization methods adopted by the Transformer models as other scholars have also reported. Therefore, BERT encoder is effective in handling inflectional languages like Sanskrit.

This thesis has been successful in initiating SATS and presenting a way forward by reporting the challenges of the process. A pipeline for the SATS was long amiss. This work has provided a pipeline for future work. Some possible future directions could include going query-focused instead of indicative summarization, incorporating linguistically annotated corpus, and finally, incorporating more data. The scope of improvement is immense although specific attention must be paid to compression rates in the future work. Deeper research is expected before a full-fledged working SATS can be seen. However, advancements from this thesis will hopefully provide a path ahead.

1)

# Appendix A

## 1. BERT- TL, EL, and Perplexity Scores for BERT

|  | training loss | evaluation loss | perplexity |
|---|---|---|---|
| Epoch | TL | EL | Perplexity |
| 0 | 9.793028 | 8.782992 | 6522.365 |
| 1 | 8.866224 | 8.54726 | 5152.618 |
| 2 | 8.633167 | 8.467617 | 4758.163 |
| 3 | 8.555754 | 8.353349 | 4244.37 |
| 4 | 8.47258 | 8.355574 | 4253.824 |
| 5 | 8.36298 | 8.153303 | 3474.838 |
| 6 | 8.292338 | 8.183633 | 3581.843 |
| 7 | 8.224774 | 8.032966 | 3080.865 |
| 8 | 8.149904 | 8.028699 | 3067.748 |
| 9 | 8.06098 | 8.017507 | 3033.604 |
| 10 | 8.017695 | 7.899263 | 2695.296 |
| 11 | 7.979699 | 7.798735 | 2437.516 |
| 12 | 7.905149 | 7.872048 | 2622.932 |
| 13 | 7.847811 | 7.664128 | 2130.534 |
| 14 | 7.792259 | 7.782048 | 2397.18 |
| 15 | 7.72217 | 7.690768 | 2188.055 |
| 16 | 7.735653 | 7.601041 | 2000.277 |
| 17 | 7.636043 | 7.495163 | 1799.318 |
| 18 | 7.596754 | 7.52613 | 1855.91 |
| 19 | 7.562581 | 7.503247 | 1813.922 |
| 20 | 7.507953 | 7.452035 | 1723.367 |
| 21 | 7.483546 | 7.363385 | 1577.166 |
| 22 | 7.452823 | 7.335828 | 1534.297 |
| 23 | 7.390176 | 7.326694 | 1520.348 |
| 24 | 7.375513 | 7.363645 | 1577.576 |
| 25 | 7.318214 | 7.22316 | 1370.814 |
| 26 | 7.306313 | 7.211308 | 1354.664 |
| 27 | 7.25396 | 7.095687 | 1206.752 |
| 28 | 7.225169 | 7.186971 | 1322.092 |
| 29 | 7.197388 | 7.113853 | 1228.873 |
| 30 | 7.15628 | 7.0323 | 1132.633 |
| 31 | 7.15593 | 7.024881 | 1124.261 |
| 32 | 7.122458 | 6.995916 | 1092.164 |

| 33 | 7.066514 | 7.092096 | 1202.426 |
| --- | --- | --- | --- |
| 34 | 7.076974 | 6.947017 | 1040.043 |
| 35 | 7.040545 | 7.007684 | 1105.092 |
| 36 | 7.01111 | 6.913548 | 1005.809 |
| 37 | 7.004498 | 6.915554 | 1007.829 |
| 38 | 6.947191 | 6.907044 | 999.2894 |
| 39 | 6.927099 | 6.876902 | 969.6175 |
| 40 | 6.913701 | 6.830075 | 925.26 |
| 41 | 6.875772 | 6.879375 | 972.0182 |
| 42 | 6.855991 | 6.784343 | 883.8994 |
| 43 | 6.842168 | 6.803803 | 901.2683 |
| 44 | 6.8565 | 6.739252 | 844.9282 |
| 45 | 6.829288 | 6.702251 | 814.237 |
| 46 | 6.764971 | 6.68086 | 797.0043 |
| 47 | 6.770052 | 6.71689 | 826.2438 |
| 48 | 6.765594 | 6.665964 | 785.2198 |
| 49 | 6.741443 | 6.675686 | 792.8911 |
| 50 | 6.700556 | 6.661384 | 781.6317 |
| 51 | 6.678063 | 6.671557 | 789.6244 |
| 52 | 6.656717 | 6.510657 | 672.2682 |
| 53 | 6.671644 | 6.697492 | 810.3706 |
| 54 | 6.661285 | 6.559104 | 705.6391 |
| 55 | 6.625422 | 6.592059 | 729.2807 |
| 56 | 6.64187 | 6.610361 | 742.7512 |
| 57 | 6.611523 | 6.556148 | 703.5564 |
| 58 | 6.624507 | 6.547969 | 697.8254 |
| 59 | 6.585166 | 6.528289 | 684.2267 |
| 60 | 6.572128 | 6.495817 | 662.3653 |
| 61 | 6.566765 | 6.550513 | 699.6032 |
| 62 | 6.575728 | 6.433891 | 622.5919 |
| 63 | 6.523091 | 6.4307 | 620.6084 |
| 64 | 6.549309 | 6.489837 | 658.4158 |
| 65 | 6.507706 | 6.467548 | 643.9029 |
| 66 | 6.49134 | 6.451207 | 633.4662 |
| 67 | 6.50333 | 6.482578 | 653.6541 |
| 68 | 6.521468 | 6.472966 | 647.4009 |
| 69 | 6.486601 | 6.517662 | 676.9938 |
| 70 | 6.488828 | 6.466883 | 643.4747 |
| 71 | 6.48003 | 6.366753 | 582.1645 |
| 72 | 6.468605 | 6.412162 | 609.2096 |
| 73 | 6.463275 | 6.463634 | 641.3873 |
| 74 | 6.464231 | 6.398533 | 600.9627 |
| 75 | 6.470681 | 6.386943 | 594.038 |

| 76 | 6.477911 | 6.397316 | 600.2321 |
| 77 | 6.480745 | 6.394238 | 598.3875 |

*Table 32: BERT TL and EL for all Epochs*

## 2. TL, EL, and Perplexity Scores for GPT-2

| Epoch | TL | EL | Perplexity |
| --- | --- | --- | --- |
| 0 | 5.811102 | 2.687399 | 14.69341 |
| 1 | 2.715544 | 2.304476 | 10.01893 |
| 2 | 2.353775 | 2.124559 | 8.369205 |
| 3 | 2.174057 | 2.016026 | 7.508427 |
| 4 | 2.055487 | 1.944076 | 6.987173 |
| 5 | 1.972301 | 1.881651 | 6.564332 |
| 6 | 1.903878 | 1.833894 | 6.258211 |
| 7 | 1.847879 | 1.79679 | 6.030259 |
| 8 | 1.802654 | 1.767658 | 5.857118 |
| 9 | 1.756164 | 1.738553 | 5.689106 |
| 10 | 1.724631 | 1.716983 | 5.567708 |
| 11 | 1.689533 | 1.695449 | 5.449089 |
| 12 | 1.661407 | 1.678215 | 5.355988 |
| 13 | 1.635499 | 1.659843 | 5.258486 |
| 14 | 1.607669 | 1.644242 | 5.177086 |
| 15 | 1.583702 | 1.636307 | 5.136166 |
| 16 | 1.562291 | 1.619196 | 5.049027 |
| 17 | 1.541556 | 1.607299 | 4.989317 |
| 18 | 1.524089 | 1.603247 | 4.96914 |
| 19 | 1.500613 | 1.591108 | 4.909185 |
| 20 | 1.481499 | 1.583723 | 4.873066 |
| 21 | 1.463264 | 1.577372 | 4.842212 |
| 22 | 1.449805 | 1.571961 | 4.816082 |
| 23 | 1.43409 | 1.566176 | 4.788301 |
| 24 | 1.423752 | 1.561599 | 4.766436 |
| 25 | 1.403829 | 1.556311 | 4.741301 |
| 26 | 1.391824 | 1.551218 | 4.71721 |
| 27 | 1.374817 | 1.546454 | 4.694791 |
| 28 | 1.365065 | 1.542721 | 4.677301 |
| 29 | 1.347534 | 1.538837 | 4.659168 |
| 30 | 1.339176 | 1.536103 | 4.64645 |
| 31 | 1.333671 | 1.535455 | 4.643438 |
| 32 | 1.322724 | 1.528766 | 4.61248 |
| 33 | 1.307925 | 1.528555 | 4.611507 |
| 34 | 1.299893 | 1.526189 | 4.600612 |
| 35 | 1.294762 | 1.525481 | 4.597353 |
| 36 | 1.285021 | 1.523198 | 4.586871 |
| 37 | 1.279073 | 1.518974 | 4.567535 |
| 38 | 1.266631 | 1.520573 | 4.574846 |

| 39 | 1.264337 | 1.518321 | 4.564554 |
| 40 | 1.257197 | 1.516886 | 4.558007 |
| 41 | 1.250898 | 1.516687 | 4.557101 |
| 42 | 1.244851 | 1.515777 | 4.552957 |
| 43 | 1.2376 | 1.51371 | 4.543555 |
| 44 | 1.233954 | 1.513445 | 4.542353 |
| 45 | 1.229634 | 1.51217 | 4.536564 |
| 46 | 1.228314 | 1.512994 | 4.540303 |
| 47 | 1.223421 | 1.512151 | 4.536477 |
| 48 | 1.221886 | 1.512177 | 4.536597 |
| 49 | 1.218302 | 1.512439 | 4.537783 |

*Table 33: GPT-2 TL and EL for all Epochs*

## 3. TL, EL and Perplexity for RoBERTa Masked for all Epochs

| Epoch | TL | EL | Perplexity |
| --- | --- | --- | --- |
| 0 | 6.848814 | 3.346867 | 28.41358 |
| 1 | 3.379138 | 2.750293 | 15.64721 |
| 2 | 2.808108 | 2.331508 | 10.29345 |
| 3 | 2.405248 | 2.062468 | 7.865356 |
| 4 | 2.13646 | 1.930152 | 6.89056 |
| 5 | 1.973811 | 1.740246 | 5.698745 |
| 6 | 1.836901 | 1.654655 | 5.231273 |
| 7 | 1.742045 | 1.581405 | 4.861782 |
| 8 | 1.67043 | 1.511476 | 4.533419 |
| 9 | 1.593421 | 1.469462 | 4.346896 |
| 10 | 1.542626 | 1.418077 | 4.129174 |
| 11 | 1.491247 | 1.387117 | 4.003291 |
| 12 | 1.442849 | 1.337408 | 3.809157 |
| 13 | 1.402146 | 1.297604 | 3.660516 |
| 14 | 1.367256 | 1.275903 | 3.581934 |
| 15 | 1.338703 | 1.226204 | 3.408267 |
| 16 | 1.303062 | 1.226909 | 3.410672 |
| 17 | 1.281643 | 1.194668 | 3.30246 |
| 18 | 1.251492 | 1.198466 | 3.315029 |
| 19 | 1.244078 | 1.148702 | 3.154095 |
| 20 | 1.218761 | 1.135232 | 3.111895 |
| 21 | 1.202075 | 1.120614 | 3.066735 |
| 22 | 1.17064 | 1.114677 | 3.048583 |
| 23 | 1.167194 | 1.075554 | 2.931616 |
| 24 | 1.15275 | 1.086483 | 2.963831 |
| 25 | 1.130431 | 1.060523 | 2.887882 |
| 26 | 1.126693 | 1.072192 | 2.921777 |
| 27 | 1.103638 | 1.041746 | 2.834161 |
| 28 | 1.085165 | 1.024709 | 2.786285 |
| 29 | 1.075372 | 1.02398 | 2.784253 |
| 30 | 1.080006 | 1.018153 | 2.768078 |
| 31 | 1.064235 | 0.992668 | 2.698424 |
| 32 | 1.057154 | 0.97325 | 2.646532 |
| 33 | 1.041886 | 0.988616 | 2.687511 |
| 34 | 1.029503 | 0.95212 | 2.591198 |
| 35 | 1.020937 | 0.998561 | 2.714372 |
| 36 | 1.018516 | 0.966686 | 2.629216 |
| 37 | 1.008692 | 0.948548 | 2.581959 |
| 38 | 1.001838 | 0.947973 | 2.580474 |
| 39 | 0.995127 | 0.978045 | 2.659253 |
| 40 | 0.990802 | 0.948377 | 2.581517 |
| 41 | 0.986182 | 0.925179 | 2.522321 |
| 42 | 0.981547 | 0.911635 | 2.488387 |
| 43 | 0.97935 | 0.941376 | 2.563507 |
| 44 | 0.96999 | 0.935082 | 2.547422 |
| 45 | 0.964788 | 0.924831 | 2.521443 |
| 46 | 0.963305 | 0.93409 | 2.544896 |
| 47 | 0.954326 | 0.918758 | 2.506176 |
| 48 | 0.949743 | 0.91533 | 2.4976 |
| 49 | 0.941784 | 0.91163 | 2.488376 |
| 50 | 0.955527 | 0.895416 | 2.448354 |
| 51 | 0.929445 | 0.893833 | 2.444481 |
| 52 | 0.926111 | 0.89769 | 2.453928 |
| 53 | 0.929695 | 0.901944 | 2.46439 |
| 54 | 0.928145 | 0.887134 | 2.428162 |
| 55 | 0.919354 | 0.901087 | 2.462278 |
| 56 | 0.919491 | 0.900615 | 2.461116 |
| 57 | 0.921855 | 0.877566 | 2.405039 |
| 58 | 0.923846 | 0.883376 | 2.419053 |
| 59 | 0.918562 | 0.87814 | 2.406419 |
| 60 | 0.913593 | 0.881265 | 2.413951 |
| 61 | 0.91145 | 0.853941 | 2.348886 |
| 62 | 0.900658 | 0.860745 | 2.364923 |
| 63 | 0.906928 | 0.88817 | 2.430678 |
| 64 | 0.90367 | 0.889655 | 2.434291 |
| 65 | 0.912969 | 0.877003 | 2.403685 |
| 66 | 0.902081 | 0.883765 | 2.419993 |
| 67 | 0.898178 | 0.875485 | 2.40004 |
| 68 | 0.902747 | 0.897508 | 2.453482 |
| 69 | 0.902983 | 0.879237 | 2.409061 |

*Table 34: RoBERTa Masked TL and EL for all Epochs*

# Appendix B

Code snippet:

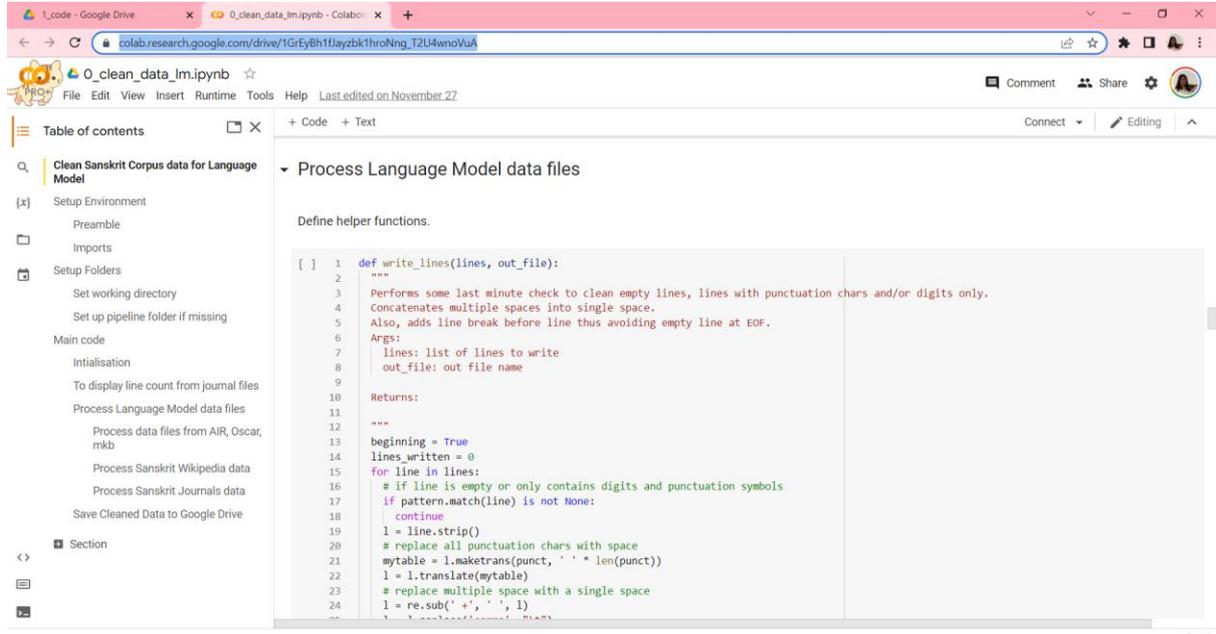

*Figure 26: Appendix B: Code for LM Data Cleaning*

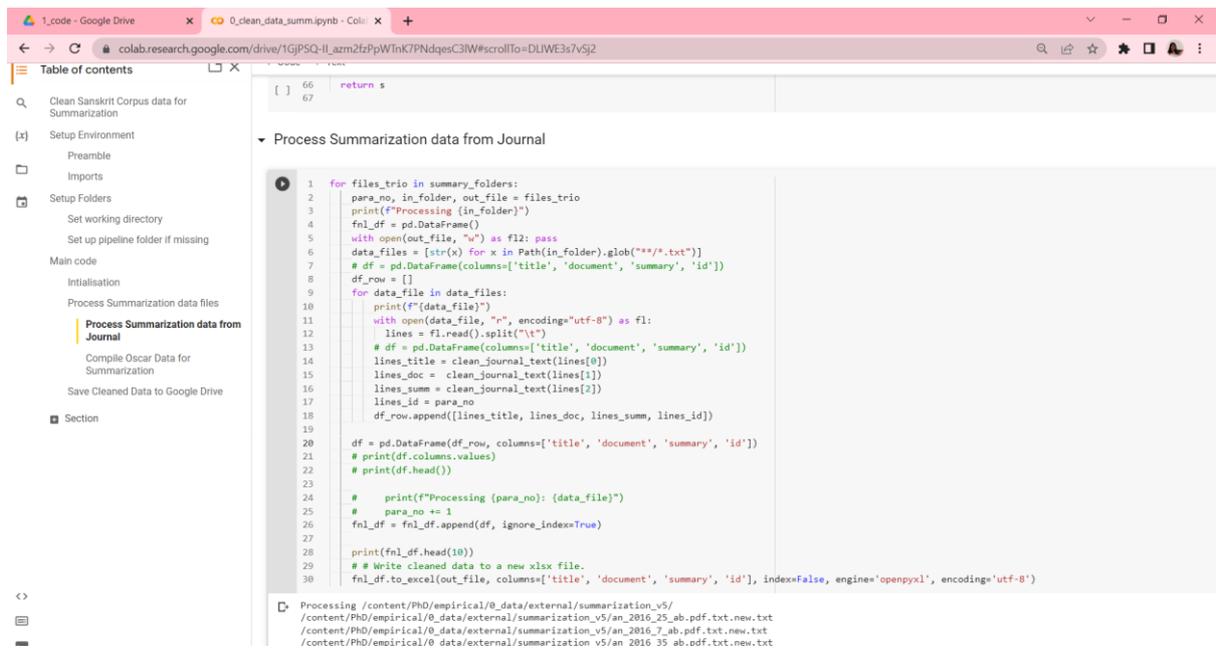

*Figure 27: Appendix B: Summarization Data Cleaning – Journal*

*Figure 28: Appendix B: Summarization Data Cleaning – OSCAR*

# Appendix C

Human Evaluation Instructions:

- **For assessment of the summarization data:**

Dear participant,

Thank you for participating in this assessment. This form contains 50 pairs (30 in section 1 and 20 in section 2) of parallel texts which have been marked 'document', 'analysis'. For each of the pairs, you have to answer if the analysis reflects/summarizes the document well. For example,
1. If the analysis rightly summarizes the document (presents the key ideas or the key topic of the document well, is relevant to the document), you may say that the analysis is a 'summary' of the text.
2. If the analysis neither reflects the key idea of the document nor has any relevance to the content, you may say that the analysis is 'unrelated'.
3. If the analysis contains key topics of the document with close relevance to the document, you should mark the analysis as 'reflective'.
4. If you find any other issue with the analysis not listed above, mark the option 'Other Issue'

Sample Data:

```
1. Document 7395   :
व्याप्या विशिष्टस्य हेतोः पक्षे सत्तैव पक्षधर्मताऽस्ति अयम् एव
परामर्शः । अनुमाने हेतुदर्शनम् त्रिवारम् भवति । प्रथमवारम् महानसे
धूमदर्शनम् भवति, द्वितीयवारम् पर्वते धूमदर्शनम् भवति, तृतीयवारम्
अपि तत्रैव पर्वते पूर्वगृहीतव्याप्तिस्मरणेन वह्निसंसक्तस्य धूमस्य
दर्शनम् भवति ।
-------------------------------------------------------------
Analysis:
अनुमानस्य प्रमुखे द्वे अङ्गे भवतः व्याप्तिः पक्षधर्मता च

☐ Summary
☐ Reflective
☐ Unrelated
☐ Other issue
```

- **For Human Evaluation of system-generated outputs:**

1. See the 10 document files given to you separately - each file has one 'source text' ('text 1, text2, ...etc) and 10 'summaries' (summary 1, summary 2, ...etc)

2. This form has 10 sections - one for each source text (marked as per the numbering in the document). For every text, rate the respective summaries on a scale of 0 to 5 for three qualities - readability, grammaticality, keyword capturing, coherence.

3. For example, if you are assessing the summaries for the 'Text 3' in the document, you should go to the section titled 'Text 3'. This section will have four questions pertaining to the four qualities of summaries mentioned above. Rate every summary for its quality.

Ratings:

1: Poor
2: Average
3: Good
4: Very Good
5: Excellent

---------------------------------------

# Appendix C1

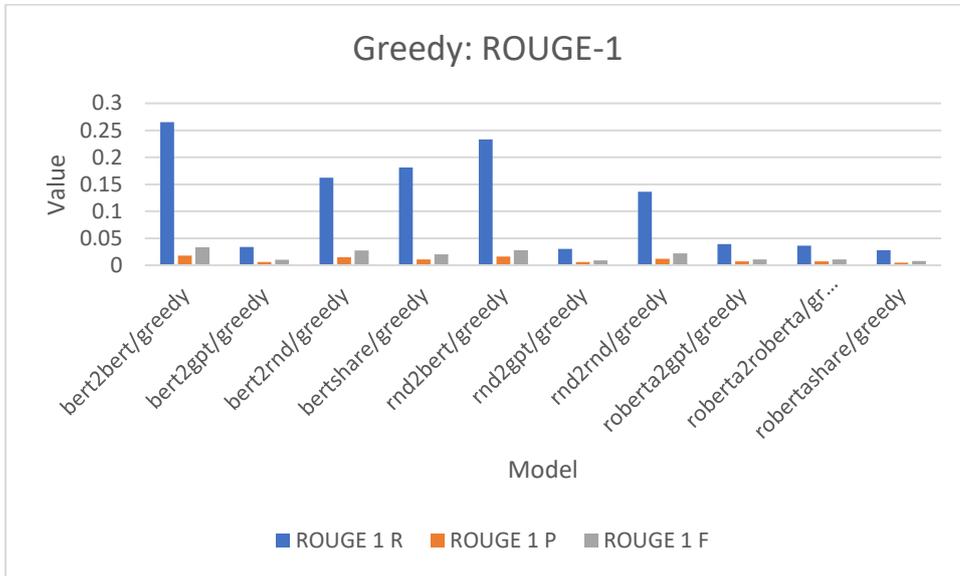

*Figure 29: Appendix C1: Greedy ROUGE1 scores*

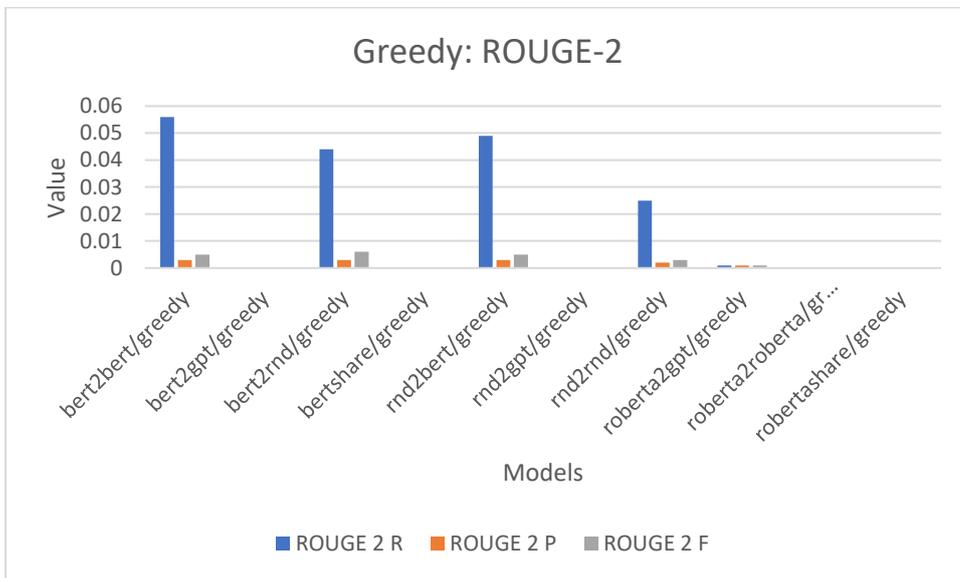

*Figure 30: Appendix C1: Greedy ROUGE-2 Scores*

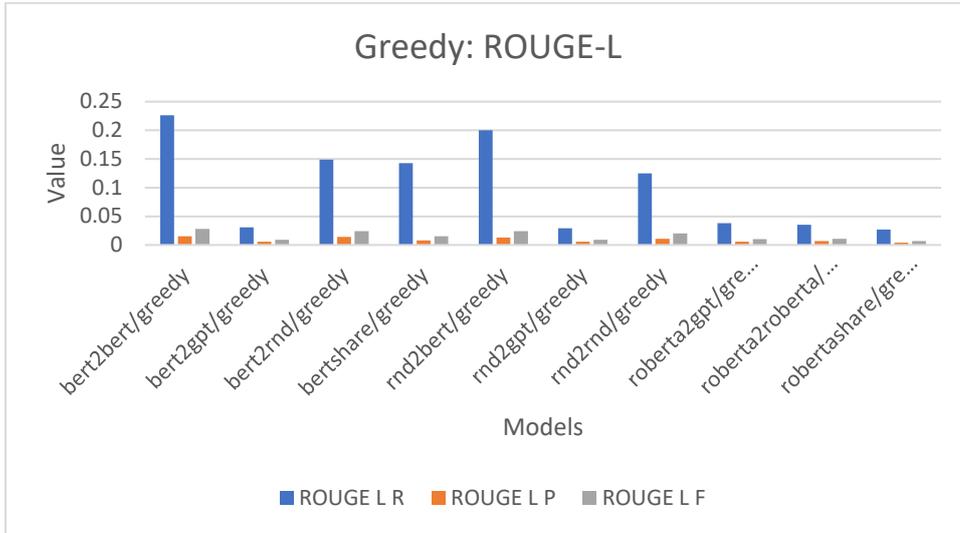

*Figure 31: Appendix C1: Greedy ROUGE-L scores*

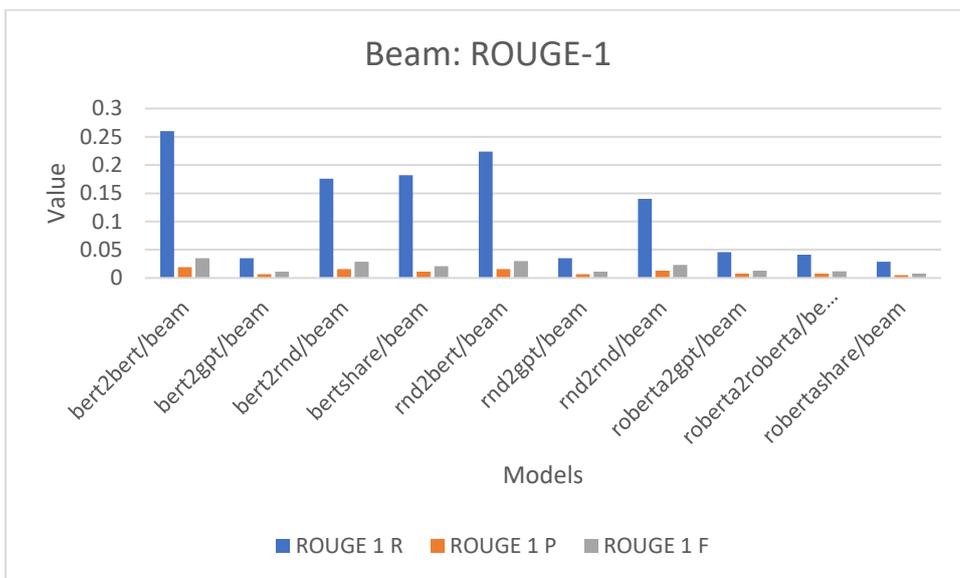

*Figure 32: Appendix C1: Beam ROUGE-1 scores*

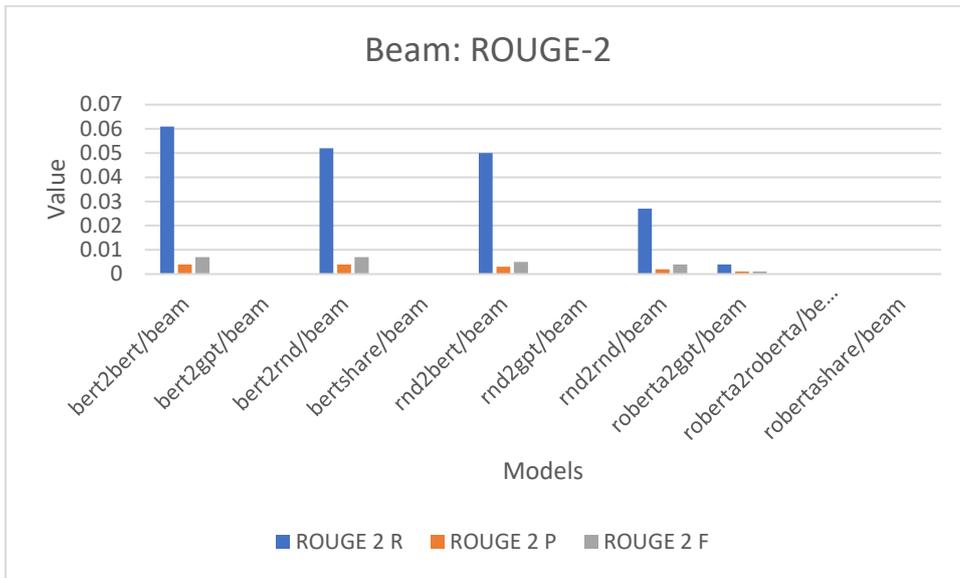

*Figure 33: Appendix C1: Beam ROUGE-2 scores*

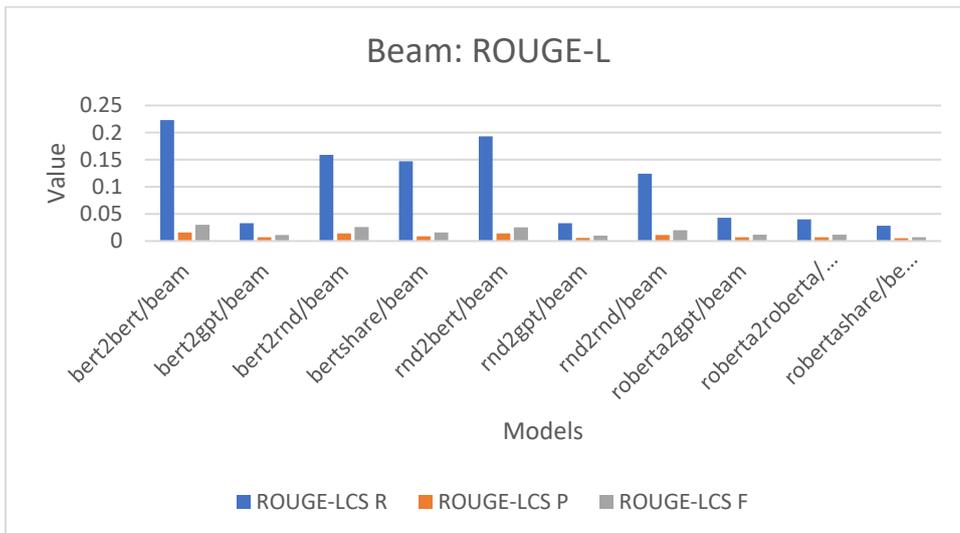

*Figure 34: Appendix C1: Beam ROUGE-L scores*

# Appendix D

List of *Sandhi* patterns changed in the data barring exceptions:

| common_error | correction | exception |
|---|---|---|
| मिति | म् इति | |
| नैव_ | न एव_ | |
| चैव_ | च एव_ | |
| इत्यत्र | इति अत्र | |
| इत्येव | इति एव | |
| अस्मिन्नपि | अस्मिन् अपि | |
| एतदर्थे | एतत् अर्थे | |
| भवत्येव | भवति+एव | |
| भवत्येव। | भवति+एव । | |
| एवञ्च | एवम्+च | |
| उक्तमपि | उक्तम्+अपि | |
| चेति | च इति | |
| तथैव | तथा एव | |
| व्यक्तेश्च | व्यक्तेः+च | |
| मपि | म् अपि | |
| मिव_ | म् इव | |
| मेव_ | म् एव | |
| दृश्यते। | दृश्यते | |
| इत्यनेन | इति+अनेन | |
| इदञ्च | इदम् च | |
| कथञ्च | कथम् च | |
| शब्दो | शब्दः | |
| सन्नेव। | सन् एव । | |
| ह्यत्र | हि+अत्र | |
| इत्थञ्च | इत्थम् च | |
| अस्त्येव | अस्ति एव | |
| अस्तीति | अस्ति इति | |
| इत्यादौ | इति आदौ | |
| एतच्च | एतत् च | |
| तस्येदमिति | तस्य इदम् इति | |
| तस्येदम् | तस्य इदम् | |
| सोऽयमिति | सः अयम् इति | |
| अयमप्यभिसम्बन्धो | अयम् अपि अभिसम्बन्धः | |
| अयमपि | अयम् अपि | |
| नावश्यमिति | न अवश्यम् इति | |
| लक्षणेति | लक्षणा इति | |
| ईदृशस्थले | ईदृशस्थले | |
| वस्तुतस्तु | वस्तुतः तु | |
| भवत्येव | भवति एव | |
| काङ्क्षा | काङ्क्षा | |
| मपेक्षते | म् अपेक्षते | |
| इत्यादौ | इति+आदौ | |
| योस्तु | योः+तु | |
| कैश्चित्तु | कैश्चित् तु | |
| यैष | य एषः | |
| यैषा | या एषा | |
| प्रपञ्चितञ्चेतत्सर्व | प्रपञ्चितम्+च+एतत्+सर्वम् | |
| निष्ठेति | निष्ठा इति | |
| चाधुना | च अधुना | |
| चेत्यादीनि | च इति आदीनि | |
| शब्दोऽयं | शब्दः अयम् | |
| इत्यर्थे | इति अर्थे | |
| इत्यर्थोऽपि | इति अर्थः अपि | |
| अयमप्यर्थो | अयम् अपि अर्थः | |
| शब्दस्यार्थो | शब्दस्य अर्थः | |
| एवार्थोऽस्य | एव अर्थः अस्य | |
| अयमप्यंशः | अयम् अपि अंशः | |
| तीति | ति इति | प्रतीति, प्रतीतिः |
| तीत्यत्र | ति इति अत्र | |
| ती-त्यादौ | ति इत्यादौ | |
| तीत्यादि | ति इत्यादि | |
| तीत्यादौ | ति इत्यादौ | |
| मीति | मि इति | |
| ○ऽ | ○ अ | |
| ोऽ | ः अ | |
| मित्यर्थः | म् इति अर्थः | |
| रपि | ः अपि | |
| रिति | ः इति | |
| स्यापि | स्य अपि | |
| स्तत्र | ः तत्र | |
| मित्येवं | म् इति एवम् | |
| मित्येवम् | म् इति एवम् | |
| ेत्युच्यते | इति उच्यते | strip() not needed |
| स्येति | स्य इति | |
| रित्यस्य | ः इति अस्य | |
| दात्मानम् | त् आत्मानम् | |
| तैरपि | तैः अपि | |
| मवाप्नोदिति | म् अवाप्नोत् इति | |
| मित्युक्तम् | म् इति उक्तम् | |
| णापि | णा अपि | |
| स्यायम् | स्य अयम् | |
| दित्यर्थे | द् इति अर्थे | |
| दित्यर्थः | द् इति अर्थः | |
| कस्तर्हि | कः तर्हि | |
| कस्मिन्नपि | कस्मिन् अपि | |
| कस्मिन्नर्थे | कस्मिन् अर्थे | |
| कस्मिंश्चित् | कस्मिंश्चित् | |
| कस्मिंश्चिदपि | कस्मिंश्चिद् अपि | |
| चिदपि | चित् अपि | |
| चिदर्थ | चित् अर्थ | |
| योरपि | योः अपि | |
| कोदृशं | कीदृशम् | |
| कोदृशी | कीदृशी | |
| स्यैव | स्य एव | |
| मावश्यकम् | म् आवश्यकम् | |
| मेवास्ति | म् एव अस्ति | |
| नैवम् | न एवम् | |
| मित्येव | म् इति एव | |
| चाऽ | च अ | |
| चेति | च इति | |
| चेत्यनेन | च इति अनेन | |
| चेदपि | चेत् अपि | |
| चेदेकमपि | चेत् एकम् अपि | |
| चेदेको | चेत् एकः | |
| चेन्न | चेत् न | |
| चेयं | च इयम् | |
| चेयम् | च इयम् | |
| चैतत् | च एतत् | |
| चैकस्य | च एकस्य | |
| चैकशतं | च एकशतं | |
| चैतन्मत | च एतत् मत | |
| चोक्तम् | च उक्तम् | |
| ष्वपि | षु अपि | |
| दिति | त् इति | अदिति, अदितिः |
| मित्युच्यते | म् इति उच्यते | |
| त्रापि | त्र अपि | |
| मिदं | म् इदम् | |
| मङ्गीकृत्य | म् अङ्गीकृत्य | |
| मङ्गीकरोति | म् अङ्गीकरोति | |
| नैक | न एक | |
| मस्ति_ | म् अस्ति_ | न्यस्तमस्तिष्क |
| मस्ति। | म् अस्ति। | न्यस्तमस्तिष्क |
| ‌ंश्च | न्+च | |
| इत्या | इति आ | |
| इत्यु | इति उ | |
| इत्ये | इति ए | |
| दित्यर्थं | त् इति अर्थं | |
| दित्यर्थः | त् इति अर्थः | |
| दित्युच्यते | त् इति उच्यते | |
| तीत्यतः | ति इति अतः | |
| प्राप्रोति | प्राप्नोति | |
| योरनन्तरं | यो:+अनन्तरं | |
| नैतत् | न+एतत् | |
| नैतदेवम् | न+एतत+एवम् | |
| नैतस्मात् | न+एतस्मात् | |
| इत्यस्मिन् | इति+अस्मिन् | |
| इत्यस्मात् | इति अस्मात् | |
| इत्यस्य | इति+अस्य | |
| इत्यस्यार्थो | इति+अस्य+अर्थः | |
| इत्यादि | इति+आदि | |
| इत्यादौ | इति+आदौ | |
| इत्याशयः | इति+आशयः | |
| इत्याहुः | इति+आहुः | |
| इत्युक्तप्रकारेण | इति+उक्तप्रकारेण | |
| इत्युच्यते | इति+उच्यते | |
| इत्येतत् | इति+एतत् | |
| स्थुल | स्थूल | |
| प्रतिपादनायोच्यते | प्रतिपादनाय+उच्यते | |
| मुच्यते | म्+उच्यते | |
| चोक्तं | च+उक्तं | |
| चोक्तम् | च+उक्तम् | |
| चापाततो | च+आपाततः | |
| श्चेति | ः च+इति | |
| चायमविच्छेदेन | च+अयम्+अविच्छेदेन | |
| मवलोकनेनेदं | म्+अवलोकनेन+इदं | |
| इत्यस्यो | इति अस्य उ | |
| चाभिधीयते | च+अभिधीयते | |
| एवार्थमाश्रित्य | एव+अर्थम्+आश्रित्य | |
| एतदेव | एतत्+एव | |
| एवायम् | एव+अयम् | |
| तस्यैव | तस्य+एव | |
| तस्यैषा | तस्य+एषा | |
| तस्यैवाधारेण | तस्य+एव+आधारेण | |
| ष्वेव | षु एव | new |
| मासीत् | म् आसीत् | |
| ञ्चेति | म् च इति | |
| मेवैकं | म् एव एकं | |
| मुभाभ्यां | म् उभाभ्यां | |
| ष्वेक_ | षु एकः | |
| नैवोपलभ्यते | न+एव+उपलभ्यते | |
| स्यास्य_ | स्य+अस्य | |
| चास्य_ | च+अस्य | |
| ञ्च_ | म् च | |
| तेनान्तरेण_ | तेन+अन्तरेण | |
| तेनोक्त | तेन+उक्तम् | |
| भगवत्येव | भगवति एव | |
| मेतत् | म् एतत् | |
| नवं. | नवम्बरः | |
| डिसं. | डिसम्बरः | |
| मनुसृत्य | म् अनुसृत्य | |
| मभिलक्ष्य | म् अभिलक्ष्य | |
| मस्ति_ | म् अस्ति_ | |
| म्ः | म् | |
| ततो_ | ततः | |
| मावहन्ति | म् आवहन्ति | |
| मादिश्यते | म् आदिश्यते | |
| नास्ति। | न+अस्ति। | |
| रस्ति | ः अस्ति | |
| मभितः | म् अभितः | |
| दृष्ट्वैव | दृष्ट्वा एव | |
| तस्यास्तु | तस्याः तु | |
| ः_ | म्_ | |
| मिच्छन् | म् इच्छन् | |
| तस्यानुसारं | तस्य अनुसारं | |
| ई स तः | | |
| ई स तमे | | |
| ई स तमे | | |
| एतस्मादतिरिक्तम् | एतस्मात् अतिरिक्तम् | |
| चाभिहितम् | च अभिहितम् | |
| कृतास्ति | कृता अस्ति | |
| कियदंशः | कियत् अंशः | |
| तास्ति_ | ता अस्ति | |
| तास्ति।_ | ता अस्ति ।_ | |
| माह_ | म् आह_ | |
| मव्यक्तम्_ | म् अव्यक्तम्_ | |
| अतो | अतः | |
| मकरोत् | म् अकरोत् | |
| _नास्ति_ | _न अस्ति_ | |
| तेष्वेकः | तेषु एकः | |
| तेनेदम् | तेन इदम् | |
| _स_ | _सः_ | |
| एको_ | एकः_ | |
| स्ः | सू | |
| इतिअनेन | इति अनेन | |
| मलिखत्_ | म् अलिखत्_ | |
| चाभूत्_ | च अभूत्_ | |
| मिच्छति_ | म् इच्छति_ | |
| महं_ | म् अहं_ | |
| माचरति | म् आचरति | |
| चाधिकृत्य | च+अधिकृत्य | |
| वर्तते[ | वर्तते | |
| वर्तत। | वर्तते | |
| मापद्य_ | म् आपद्य | |
| मधिकृत्य | म् अधिकृत्य | |
| मुपगताः | म् उपगताः | |
| तास्ति | ता अस्ति | |
| तास्ति। | ता अस्ति। | |
| मासीत्। | म् आसीत्। | |
| योर्मध्ये | योः मध्ये | |
| मारभन्ते | म् आरभन्ते | |
| ! | । | |
| मवलमब्य | म् अवलम्ब्य | |

# Appendix E

Language Model Data (Some data has not been quoted due to copyright and licensing issues)

Mann Ki Baat

| | |
|---|---|
| 17680 | भारतीयविदुषीणाम् दीर्घा परम्परा प्रवर्तते । |
| 17680 | वेदानाम् ऋचानाम् आविष्करणे भारतस्य अनेकासाम् विदुषीणाम् ऋषिकाणाम् च सुबहु योगदान म् आसीत् । |
| 17680 | लोपामुद्रा गार्गी मैत्रेयी अपाला चेत्यादयः न जाने कति कति नामानि सन्ति अद्यत्वे वयम् कन्याम् रक्ष कन्याम् पाठय इति प्रयतामहे परं चसहस्राब्देभ्यः प्राक् अस्मदीयेषु शास्त्रेषु स्कन्दपुराणे कथितम् । |
| 17681 | अर्थात् एका पुत्री दशपुत्रैः तुल्या भवति । |
| 17681 | दशभिः पुत्रैः यत् पुण्यम् प्राप्यते एकया पुत्र्या तत् पुण्यम् लभ्यते । |
| 17681 | तथ्यम् इदम् अस्मदीयसमाजे नार्याः महत्वम् दर्शयति । |
| 17681 | अत एव अस्मदीये समाजे नारी शक्तिस्वरूपा इति प्रतिष्ठापितम् । |
| 17681 | एषा नारीशक्तिः सम्पूर्णम् अपि देशम् अशेषम् समाजम् कृत्स्नम् अपि कुटुम्बं चएकतासूत्रेण आबध्नाति । |
| 17681 | भवन्तु नाम ताः वैदिककालिकाः विदुष्यः लोपामुद्रागार्गी मैत्रेयीप्रभृतयः तासाम् विद्वता वा अक्कामहादेवीमीराबाईप्रभृतीनाम् ज्ञानम् वा भक्तिः भवतु वा सा अहिल्याबाईहोलकरस्य शासनव्यवस्था आहोस्वित् भवतु सा राज्ञाः लक्ष्मीबाईवीराङ्गनायाः वीरता नारीशक्तिः सर्वदैव अस्मान् अनारतम् प्रेरयति स्म सततं चप्रेरयति । |
| 17681 | देशस्य मानम् सम्माननं चविवर्धयन्ती प्रवर्तते । |
| 17682 | राष्ट्रपतिमहोदयः असाधारणमहिलानाम् वृन्दमेकम् अमिलत् याः स्वस्वक्षेत्रेषु सर्वप्रथमम् किम् अपि उत्कृष्टम् अकुर्वन् । |
| 17682 | नूतनम् कीर्तिमानम् प्रतिष्ठापितवत्यः । |
| 17682 | देशस्य एताः महिलाः सन्ति इति नौसेनायाः प्रथमा पोतचालिका इति यात्रिरेलयानस्य प्रथमा महिलाचालिका रेलयानस्य प्रथमा महिलाचालिका इति अग्निशमनयानस्य प्रथमा महिलाचालिका प्रथमा महिला बसयानचालिका इति दक्षिणीयध्रुवम् सम्प्राप्ता प्रथमा महिला ऐवरेस्ट इति सर्वोच्चपर्वतशिखरम् आरूढवती प्रथमा महिला एवम् हि प्रत्येकम् अपि क्षेत्रे इति अस्मदीयाः प्रथमाः नारीशक्तयः समाजस्य रूढिवादिताम् अपाकुर्वत्यः असाधारणाः उपलब्धीः अवाप्नुवन् कीर्तिमानानि च प्रतिष्ठापितवत्यः । |
| 17682 | एताः इदम् प्रदर्शितवत्यः यत् कठोरश्रमम् निष्ठाम् दृढसंकल्पं चआधृत्य सर्वाः अपि बाधाः विघ्नान् चातिक्रम्य नूनम् नवीनम् मार्गमुपकल्पयितुम् शक्यते । |
| 17682 | तादृशः नवीनः मार्गः यः न केवलम् स्वीयसमकालीनानाम् जनानाम् कृते अपि तु भाविप्रजानाम् कृतेपि प्रेरकः स्यात् । |
| 17682 | नूतनया शक्त्या नवीनेन च उत्साहेन तान् आपूरयेत् । |
| 17682 | एताः समुपलब्धवती प्रथमाः महिलाः आलक्ष्य एकम् पुस्तकम् अपि सज्जीकृतम् अस्ति येन हि अशेषदेशः आसाम् नारीशक्तीनाम् विषये अवगताः स्युः एतासाम् जीवनेभ्यः कार्येभ्यः च प्रेरिताः भवेयुः । |
| 17682 | पुस्तकम् इदम् इति अत्र रूपेण अपि समुपलभ्यते । |
| 17683 | साम्प्रतम् देशे समाजे च सञ्जायमाने सकारात्मके परिवर्तने देशस्य नारीशक्तेः महत्वपूर्णा भूमिका वर्तते । |
| 17683 | अद्य वयम् यदा महिलाशक्तीकरणस्य चर्चाम् कुर्वन्तः स्मः तदा अहमेकम् रेलस्थानकम् सन्दर्भयितुमीहे । |
| 17683 | रेलस्थानकम् महिलाशक्तीकरणं चभवन्तः नूनम् विचारयिष्यन्ति यदनयोः मध्ये कः सम्बन्धः मुम्बय्याः मातुङ्गास्टेशन इति रेलस्थानकम् भारतस्य प्रप्रथमम् रेलस्थानकम् वर्तते यत्र सर्वाः अपि कर्मकराः महिलाः एव सन्ति । |
| 17683 | सर्वेषु विभागेषु महिलाकर्मचारिण्यः आहत्य चत्वारिंशदधिकाः महिलाः सन्ति । |
| 17683 | ऐषमः गणतन्त्रदिवसीयानि सामूहिकप्रयाणानि दृष्ट्वा इति अत्र अन्येषु च इति सामाजिकमाध्यमेषु जनैः लिखितम् यत् अत्र मुख्यम् आकर्षणम् आसीत् सीमसुरक्षाबलस्य बाइकयानचालिकानाम् वृन्दम् यस्मिन् सर्वाः अपि महिलाचालिकाः आसन् । |
| 17683 | एताः साहसपूर्णप्रयोगम् कुर्वन्त्यः आसन् अपि चेदम् दृश्यम् वैदेशिकान् अतिथीन् अपि आश्चर्यचकितान् करोति स्म । |
| 17683 | शक्तीकरणम् हि आत्मनिर्भरतायाः एव अनन्यतमम् रूपम् । |
| 17683 | साम्प्रतम् अस्मदीया एषा नारीशक्तिः नेतृत्वम् करोति । |
| 17683 | आत्मनिर्भरा भवति । |
| 17683 | सद्यः स्मरामि यत् छत्तीसगढस्य अस्मदीयाः आदिवासिमहिलाः आश्चर्यम् अजनयन्। |
| 17683 | एताः नूतनोदाहरणम् प्रस्तुतवत्यः। |
| 17683 | आदिवासिमहिलानाम् यदा उल्लेखः भवति तदा अस्माकम् सर्वेषाम् मनस्सु एकम् सुनिश्चितम् चित्रम् उत्पद्यते। |
| 17683 | यत्र अरण्यम् भवति पादमार्गः भवति तस्मिन् काष्ठभारवाहिन्यः महिलाः प्रचलन्त्यः दृश्यन्ते। |
| 17683 | परं छत्तीसगढस्य अस्मदीयाः आदिवासिनार्यः अस्मदीया एषा नारीशक्तिः देशस्य सम्मुखम् एकम् नवीनम् चित्रम् व्यरचयत्। |
| 17683 | छत्तीसगढस्य दंतेवाड़ाक्षेत्रम् यद्धि माओवाददुष्प्रभावितम् वर्तते। |
| 17683 | हिंसा अत्याचरणम् बम्‌इति विस्फोटकम् भुशुण्डिका लघुभुशुण्डिका माओवादिनः एतानि आधृत्य भयानकम् परिवेशम् निर्मितवन्तः। |
| 17683 | एतादृशेषु भयावहेषु क्षेत्रेषु आदिवासिमहिलाः इति चक्रिकायानानि चालयित्वा आत्मनिर्भराः भवन्ति। |
| 17683 | अल्पीयसि कालखण्डे अनेकाः महिलाः अमुना कार्येण संयुताः जाताः। |
| 17683 | एवम् हि लाभत्रयम् सिद्ध्यति एकतः स्ववृत्तितया एताः शक्ताः भवन्ति अपरतः च अमुना कार्येण माओवाददुष्प्रभावितस्य क्षेत्रस्य परिदृश्यम् अपि परिवर्तते। |
| 17683 | तथा च युगपदेव अनेन पर्यावरणसंरक्षणस्य कार्यम् अपि सबलम् भवति। |
| 17683 | अत्रत्यम् जनपदप्रशासनम् अपि प्रशंसार्हम् वर्तते येन अनुदानादिकम् उपलभ्यित्वा आभ्यः प्रशिक्षणादिकम् अपि प्रदीयते एवम् हि जनपदप्रशासनेन आसाम् महिलानाम् सफलतावाप्तौ महत्वपूर्णा भूमिका निर्व्यूढास्ति। |
| 17684 | वयम् पुनः पुनः शृण्वन्तः स्मः यत् जनाः कथयन्ति कुछ बात है ऐसी कि हस्ती मिटती नहीं हमारी। |
| 17684 | किञ्चिद् वैशिष्ट्यम् अस्ति तादृशम् यदस्तित्वम् अविनाशि अस्मदीयम् तत् किम् अस्ति तदस्ति आनम्यता परिवर्तनशीलत्वम् रूपान्तरत्वम्। |
| 17684 | यद्धि कालबाह्यम् अस्ति तत् त्याज्यम् यद्धि आवश्यकम् तस्य परिष्कारः स्वीकर्तव्यः। |
| 17684 | अपि च अस्मदीयसमाजस्य विशेषता वर्तते आत्मपरिष्कारस्य अनारतम् प्रयासः एषास्ति भारतीया परम्परा एषा अस्माकम् संस्कृतिः अस्मभ्यम् रिक्थत्वेन अधिगता अस्ति। |
| 17684 | कस्यचन अपि जीवनसमाजस्य अभिज्ञानम् भवति तस्य आत्मपरिष्कारस्य तन्त्रम्। |
| 17684 | सामाजिककुप्रथाः कुरीतीः च विरुध्य सहस्राब्देभ्यः अस्माकम् देशे व्यक्तिगतेषु सामाजिकेषु च स्तरेषु सततम् प्रयासाः भवन्तः आसन्। |
| 17684 | नातिचिरम् बिहारराज्ये रोचकः प्राथमिकः प्रयासः अभवत्। |
| 17684 | राज्ये सामाजिक कुरीतीः समूलम् उन्मूलयितुम् त्रयोदशसहस्रकिलोमीटरमिताधिका विश्वस्य दीर्घतमा मानवशृङ्खला विरचिता। |
| 17684 | अमुना अभियानेन बालविवाहयौतुकप्रथासदृशीः कुरीतीः विरुध्य जनेषु जागृतिः प्रसारिता। |
| 17684 | यौतुकबालविवाहसदृशीः कुरीतीः विरुध्य अशेषराज्यम् योद्धुम् संकल्पितवत्। |
| 17684 | आबालमहिलावृद्धाः युवानः च सोत्साहम् अस्मिन् अभियाने सहभागिनः आसन्। |
| 17684 | पटनानगर्याः ऐतिहासिकात् गान्धिमैदानस्थलात् आरभ्य मानवशृङ्खला एषा राज्यस्य सीमप्रदेशम् यावत् अविच्छिन्नरूपेण संकलिता जाता। |
| 17684 | समाजस्य सर्वेऽपि जनाः समुचितरूपेण विकासस्य लाभान् अवाप्नुयुः इति कृत्वा एतदावश्यकम् यत् अस्माकम् समाजः एताभ्यः कुरीतिभ्यः मुक्तः स्यात्। |
| 17684 | आगच्छन्तु वयम् सर्वे मिलित्वा समाजात् एताः कुरीतीः अपास्तुम् प्रतिज्ञाम् करवाम तथा च एकम् नवीनम् शक्तम् सबलम् समर्थ भारतम् निर्मातुम् प्रयतेम। |
| 17684 | अहम् बिहारस्य जनान् राज्यस्य मुख्यमन्त्रिणम् तत्रत्यम् प्रशासनम् मानवशृङ्खलायाम् सहभागिनः प्रत्येकम् अपि जनान् च अभिनन्दामि वर्धयामि यत् ते समाजकल्याणदिशि विशिष्टाम् व्यापिनीं च प्रक्रियाम् आरभन्त। |
| 17685 | मम प्रियाः देशवासिनः कर्नाटकस्य मैसूरुतः श्रीमान् दर्शनः इति अत्र अलिखत् तस्य पितुः उपचारार्थम् प्रतिमासम् षट्सहस्ररूप्यकात्मकः व्ययः भवति स्म। |
| 17685 | सः पूर्वम् प्रधानमन्त्रिजनौषधियोजनाविषये न एव जानाति स्म। |
| 17685 | परं चसम्प्रति यदा सः जनौषधिकेन्द्रविषये सूचनाम् अलभत ततः च औषधिक्रयणम् व्यदधात् ततः प्रभृति एष व्ययः प्रतिशतम् पञ्चसप्ततिमित्या अपचितः। |
| 17685 | सः वाञ्छति यदहम् मन की बात प्रसारणे एतद्विषये उद्घोषयेयम् येन अधिकाधिकाः जनाः एतद्विषये अवगताः स्युः अमुना च लाभान्विताः भवेयुः। |
| 17685 | विगतेभ्यः कतिपयेभ्यः दिनेभ्यः अनेके जनाः एतद्विषये माम् लिखन्ति स्म सूचयन्ति स्म। |
| 17685 | अहम् अपि सामाजिकसंचारमाध्यमेषु दृष्टवान् यदनेके जनाः अनया योजनया लाभान्विताः अभूवन्। |

| 17685 | तथा च यदा एतादृशी सूचना अवाप्यते तदा हर्षप्रकर्षः अनुभूयते । |
|---|---|
| 17685 | गभीरः सन्तोषः जायते । |
| 17685 | एतदपि मह्यम् अतितराम् अरोचत यत् श्रीमान् दर्शनः स्वीयमनसि विचारमिमम् अनुभूतवान् यत् यत्किम् अपि तेन लब्धम् तद् अन्येपि अवाप्नुयुः । |
| 17685 | अस्याः योजनायाः उद्देश्यम् अस्ति स्वास्थ्यपरिचर्या सुलभा स्यात् अपि च जीवनस्य सरलता प्रोत्साहिता भवेत् । |
| 17685 | जनौषधिकेन्द्रेषु प्राप्याः औषध्यः आपणेषु विक्रीयमाणाभ्यः चिन्हिताभ्यः औषधीभ्यः प्रायेण प्रतिशतम् पञ्चाशत्तः नवतिमित्या न्यूनार्घाः भवन्ति । |
| 17685 | अमुना जनसामान्यम् विशेषेण प्रतिदिनम् औषधीम् सेवमानानाम् वरिष्ठनागरिकाणाम् कृते अतितराम् आर्थिकम् साहाय्यम् प्राप्यते भूरिशः सञ्चयो भवति । |
| 17685 | साम्प्रतम् अशेषदेशे त्रिसहस्राधिकानि जनौषधिकेन्द्राणि स्थापितानि सन्ति । |
| 17685 | अनेन न केवलम् औषधयः न्यूनार्घाणि प्राप्यन्ते अपि तु वैयक्तिकउद्यमिनाम् कृतेपि वृत्तितायाः नवीनाः अवसराः अवाप्यन्ते । |
| 17685 | प्रधानमन्त्रिभारतीयजनौषधिकेन्द्रेषु चिकित्सालयेषु च अमृतसंग्रहेषु एताः न्यूनार्घाः औषधयः लभ्यन्ते । |
| 17685 | अत्र इदम् एवउद्दिष्टम् यत् देशस्य निर्धनतमाः जनाः गुणवत्तायुताम् सुलभाम् न्यूनार्घा चस्वास्थ्योपचर्याम् अवाप्नुयुः येन हि स्वस्थस्य समृद्धस्य च भारतस्य निर्माणम् सम्भवम् भवेत् । |
| 17686 | मम प्रियाः देशवासिनः महाराष्ट्रस्य श्रीमान् मंगेशः इति अत्र चित्रमेकम् प्राहिणोत् । |
| 17686 | तच्चित्रम् तादृशम् आसीत्न्मम ध्यानम् सहसा तदाकृष्टम् जातम् । |
| 17686 | तस्मिन् पौत्रः निजपितामहेन साकम् इति स्वच्छताभियाने सहभागित्वम् आवहति । |
| 17686 | अहम् ज्ञातवान् यत् अकोलाक्षेत्रीयाः नागरिकाः स्वच्छभारताभियानस्य अन्तर्गतम् मोरानद्याः स्वच्छीकरणार्थम् स्वच्छताभियानम् आयोजितवन्तः । |
| 17686 | मोरानदी पूर्वम् आवर्षम् सजला वहति स्म साम्प्रतम् सा ऋतुनिष्ठा सञ्जाता । |
| 17686 | अपरो वेदनाविषयः आसीत् यत् एषा नदी पूर्णरूपेण आरण्यकैः ग्रासैः जलकुम्भिभिः समाकीर्णा सञ्जाता । |
| 17686 | नद्याम् अस्याः तटप्रदेशे च प्रभूतः अवकरप्रक्षेपः क्रियते स्म । |
| 17686 | एका इति कार्ययोजना सज्जीकृता तथा च मकरसंक्रान्तितः एकदिनपूर्वम् जान्युआरिमासे त्रयोदशदिनाङ्के इति अस्य प्रथमचरणस्य अन्तर्गतम् किलोमीटरचतुष्कस्य क्षेत्रे चतुर्दशस्थानेषु मोरानद्याः तटम् अभितः स्वच्छीकृतम् । |
| 17686 | इत्यस्याः पुण्यकार्ये अकोलाक्षेत्रस्य षट्सहस्राधिकाः नागरिकाः शताधिकानि स्वैच्छिकसंघटनानि महाविद्यालयाः विद्यार्थिनः बालाः वृद्धाः मातरः भगिन्यः सर्वेपि अत्र सहभागित्वम् आवहन् । |
| 17686 | ऐषम् जान्युआरिमासे विंशतिदिनेपि एतत् स्वच्छताभियानम् सततम् प्रवर्तितम् तथा च अहम् सूचितः यत् यावदवधि मोरानदी पूर्णतया स्वच्छा न एव भवति अभियानम् एतत्प्रतिशनिवासरम् प्रातः प्रवर्तिता । |
| 17686 | एतत् प्रमाणयति यत् जनः यदि किम् अपि कर्तुम् दृढसंकल्पम् करोति चेत् तदा किञ्चित् अपि अशक्यम् न अस्ति । |
| 17686 | जनान्दोलनमाध्यमेन बृहत्तमानि परिवर्तनानि कर्तुम् शक्यन्ते । |
| 17686 | अहम् अकोलावासिनः अत्रत्ये जनपदनगरनिगमयोः प्रशासने च एतत्कार्यम् जनान्दोलनरूपेण प्रवर्तयितुम् संलग्नान् सर्वान् नागरिकान् च तेषामेषाम् प्रयासानाम् कृते भूरिशः वर्धयामि अपि च कामये यत् भवताम् एते प्रयासाः देशस्य अन्येषाम् जनानाम् कृतेपि प्रेरणादायिनो भवेयुः । |
| 17687 | केरलस्य आदिवासिमहिला लक्ष्मीकुट्टी इत्यस्याः कथाम् श्रुत्वा भवन्तः सुखदमाश्रयम् अनुभविष्यन्ति । |
| 17687 | लक्ष्मीकुट्टी कल्लारे शिक्षिका अस्ति तथा च साम्प्रतम् अपि गहनारण्येषु आदिवासिक्षेत्रे तालपत्रैः विनिर्मिते कुटीरे निवसति । |
| 17687 | सा स्वीयस्मृत्याधारेण एव पञ्चशतम् वनस्पतीनाम् औषधीः व्यरचयत् । |
| 17687 | सर्पदंशस्य उपचारार्थम् उपयुज्यमानायाः औषधेः निर्माणे तस्याः प्रावीण्यम् सुसिद्धम् । |
| 17687 | लक्ष्मीवर्या वनस्पतीयौषधानाम् निजज्ञानकारणात् अनारतम् समाजम् सेवते । |
| 17687 | अज्ञातनाम्न्याः अस्याः अभिज्ञानम् कृत्वा समाजे अनया अनुष्ठिताय योगदानाय सा पद्मश्रीति अलंकरणेन सम्मानिता अस्ति । |
| 17687 | अद्य एकस्य अपरस्य नाम्नः समुल्लेखात्नाहम् आत्मानम् वारयितुम् शक्नोमि । |
| 17687 | पश्चिमबङ्गालस्य पञ्चसप्ततिवर्षीया सुभासिनीमिस्त्री इत्यस्याः नाम । |
| 17687 | सा पुरस्कारार्थम् वृता अस्ति । |

| 17687 | सुभासिनीमिस्त्री तादृशी महिला अस्ति या चिकित्सालयस्य निर्माणार्थम् अपरेषाम् गृहेषु भाण्डानि स्वच्छीकृतवती शाकादिक्रयम् च व्यदधात्। |
|---|---|
| 17687 | यदा सा त्रयोविंशतिवर्षीया आसीत् तदा उपचारस्य अभावकारणात् अस्याः पत्युः मृत्युरभवत् एषा एव घटना ताम् निर्धनानाम् कृते चिकित्सालयनिर्माणार्थम् प्रैरयत्। |
| 17687 | सम्प्रति अस्याः कठोरश्रमेण विनिर्मिते चिकित्सालये परस्सहस्राणाम् निर्धनानाम् निःशुल्कम् उपचारो विधीयते। |
| 17687 | दृढमहम् विश्वसिमि यत् अस्माकम् बहुरत्नावसुन्धरायाम् तादृशि अनेकानि नररत्नानि सन्ति असंख्यानि नारीरत्नानि च सन्ति तम् ताम् चन एकोपि जानाति न वा तेषाम् सम्यग् अभिज्ञानम् भवति। |
| 17687 | एतादृशानाम् जनानाम् अभिज्ञानम् न भवति चेत् समाजस्य अपि हानिः भवति। |
| 17687 | पद्मपुरस्कारो नाम माध्यमम् अस्ति परं चअहम् देशवासिभ्योपि निवेदयामि यत् अस्मान् परितः समाजस्य कृते समर्पितजीवनाः समाजार्थम् निष्ठावन्तः काम् अपि काञ्चित् अपि विशेषताम् सन्धारयन्तः आजीवनम् कार्यानुष्ठातारः लक्षलक्षाधिकाः जनाः सन्ति। |
| 17687 | नूनम् कदाचित् अपि ते समाजधारायाम् समावेशनीयाः। |
| 17687 | न ते मानसम्मानार्थम् कार्याणि आचरन्ति परं चतेषाम् एभिः कार्यैः वयम् प्रेरिताः भवामः। |
| 17687 | कदाचित् विद्यालयेषु कदाचित् महाविद्यालयेषु च एतादृशाः महनीयजनाः समामन्त्रणीयाः तेषाम् चअनुभवाः श्रवणीयाः। |
| 17687 | पुरस्कारेभ्योपि अग्रे समाजेपि केचन प्रयासाः भवेयुः। |
| 17688 | मम प्रियाः देशवासिनः प्रतिवर्षम् वयम् जान्युआरिमासे नवमे दिनाङ्के प्रवासिभारतीयदिवसम् आमन्यामहे। |
| 17688 | अयम् एवदिवसः अस्ति यदा पूज्यः महात्मागाँधी दक्षिणाफ्रीकातः भारतम् प्रत्यागतः। |
| 17688 | अस्मिन्नेव दिने वयम् भारते विश्वस्मिन् विश्वे च निवसताम् भारतीयानाम् मध्ये अभिनबन्धनस्य उत्सवमायोजयामः। |
| 17688 | ऐषम् प्रवासिभारतीयदिवसावसरे वयम् कार्यक्रममेकम् आयोजितवन्तः यत्र विश्वस्मिन् विश्वे निवसन्तो भारतीयमूलस्य सर्वेपि सांसदाः महापौराः चआमन्त्रिताः आसन्। |
| 17688 | भवन्तः इदम् श्रुत्वा प्रसन्नताम् अनुभविष्यन्ति यत् तस्मिन् कार्यक्रमे दक्षिणअफ्रीकातः अपरेभ्योपि देशेभ्यः च यत्र यत्र अस्मदीयाः महापौराः मूलभारतीयाः सांसदाः च सन्ति ते सर्वेपि सहभागित्वमावहन्। |
| 17688 | प्रसीदामितराम् यत् विभिन्नेषु देशेषु निवसन्तः भारतीयमूलाः जनाः तान् देशान् तु सेवन्ते एव युगपदेव ते भारतेन साकम् अपि स्वीयान् दृढतरान् सम्बन्धान् संधारयन्ति। |
| 17688 | ये यूरोपस्य भिन्नभिन्नेषु देशेषु अस्माकम् मूलभारतीयाः निवसन्ति तेषु केचन इति क्षेत्रे कार्याणि कुर्वन्ति केचन आयुर्वेदक्षेत्रे समर्पिताः सन्ति अपरे स्वीयसंगीतमाध्यमेन समाजस्य मनांसि रञ्जयन्ति इतरे च केचन स्वीयललितकवितारचनाभिः सहृदयान् आनन्दसन्दोहेन आप्लावयन्ति। |
| 17688 | केचन इति जलवायुपरिवर्तनविषये अनुसन्धानम् विदधति अन्ये च भारतीयग्रन्थान् आश्रित्य कार्यनिरताः सन्ति। |
| 17688 | कश्चन भारवाहियानम् चालयित्वा गुरुद्वारानिर्माणम् अकरोत् अपरेण च केनचित् मस्जिद्निर्माणम् कृतम्। |
| 17688 | अर्थात् यत्र कुत्र अपि अस्मदीयाः जनाः सन्ति ते तत्रत्याम् धराम् येन केनापि प्रकारेण सुसज्जिताम् कृतवन्तः। |
| 17688 | अहम् धन्यवादान् वितरामि यूरोपीयसंघस्य अस्य उल्लेखनीयकार्यस्य कृते भारतीयमूलस्य जनानाम् अभिज्ञानार्थम् अपि च अमुना माध्यमेन भारतस्य नागरिकेभ्यः अपि सूचनाप्रदानार्थम्। |
| 17688 | तथा च एतन्माध्यमेन अशेषजगतः जनान् सूचयितुम् अपि। |
| 17689 | जान्युआरिमासे त्रिंशत्तमेदिने पूज्यबापूवर्यस्य पुण्यतिथिः वर्तते यः अस्मान् सर्वान् अभिनवम् मार्गम् प्रादर्शयत्। |
| 17689 | दिवसमेनम् हुतात्मदिनत्वेन वयम् आयोजयामः। |
| 17689 | एतस्मिन् दिने वयम् देशस्य रक्षायै हुतात्मभ्यः महद्भ्यः प्रातः एकादशवादने श्रद्धाञ्जलिम् अर्पयामः। |
| 17689 | शान्तेः अहिंसायाः च मार्गः एव बापूमहात्मनः अनन्यमार्गः आसीत्। |
| 17689 | भवतु नाम भारतम् इदम् वा जगत् स्यात् इदम् कुटुम्बम् वा एकाकी जनो वा सकलः समाजः पूज्यबापूवर्यः येषाम् सिद्धान्तानाम् आदर्शानां चक्रते जीवति स्म पूज्यबापूः यत् किम् अपि अस्मान् समदिशत् तत्सर्वम् अद्यापि अतितराम् उपयोगि वर्तते। |
| 17689 | ते केवलम् सिद्धान्ताः एव नैवासन्। |
| 17689 | वर्तमानकालेपि वयम् पदे पदे पश्यामः यत् बापूवर्यस्य सन्देशः कियान् समीचीनः अस्ति। |
| 17689 | यदि वयम् संकल्पयेम यत् बापूवर्यस्य मार्गमनुसरेम यावन्तम् अपि प्रचलितुम् शक्नुमः प्रचलेम तदा एतस्मात् समुचिततरः कः अन्यः श्रद्धाञ्जलिः भविता। |
| 17690 | विगतेषु दिनेषु इज़राइलप्रधानमन्त्रिणा साकमहम् गुजराते अमदाबादे इति कार्यक्रमस्य उद्घाटनाय अवसरम् लब्धवान्। |
| 17690 | तत्राहम् सूचितः यत् कश्चन अन्यतमो युवा तादृशमेकम् इति अङ्कीयोपकरणम् विकासितवान् यस्य साहाय्येन कश्चन वक्तुम् अक्षमो जनः तस्योपकरणमाध्यमेन स्वीयम् हार्दम् लिखेदनेव तद् ध्वनिरूपेण परिणमते तथा च भवान् तेन सम्भाषितुम् तथा एव शक्नोति यथा केनचित् भाषणक्षमेन साकम् कर्तुम् पारयति। |
| 17690 | अवगच्छामि यत् इति कृत्रिमप्रज्ञाम् वयम् अनेकासु विधासु समुपयोक्तुम् प्रभवामः । |

# Anantaa Journal

| | |
|---|---|
| 6100000 | वैयाकरणमतेषु कारकसंख्याविषये अनवद्यम् तत्त्वम् हि कर्मादीनि कारकाणि षट् । |
| 6100000 | परन्तु सम्यग्व्याकरणपरिशीलितमतिना भर्तृहरिणा प्रतिपाद्यते सप्तम कारकविशेषः । |
| 6100000 | सः अयम् विशेषः शेषाधिकारे षष्ठ्यर्थ इति नाम्ना हरिणा प्रतिपाद्यते । |
| 6100000 | तच्छरूपो भवति सम्बन्धरूपः । |
| 6100000 | अत एव सम्बन्धस्य अपि कारकत्वम् अङ्गीकृतम् हरिणा । |
| 6100000 | तथा च शेषषष्ठ्याः कारकषष्ठ्याः च कथम् विवेक इति अस्मिन् प्रबन्धे निरूप्यते । |
| 6100001 | कर्मादीनाम् व्यतिरिच्य षष्ठ्यर्थ इति सप्तमकः कारकभेद इति हरिवचनम् । |
| 6100001 | सः अयम् कारकभेदः शेषरूपः शेषरूपसम्बन्धः वेति विविच्यते । |
| 6100001 | कारकम् नाम क्रियया सह अन्वयसामर्थ्यम् पदम् । |
| 6100001 | इति अस्य अर्थः हि प्रत्यक्षाप्रत्यक्षभावेन यदि क्रियया सह तत्पदस्य केनापि प्रकारेण अन्वयो भवति चेत् तस्मिन् कारकत्वम् उपपन्नम् भवति । |
| 6100001 | यथा राज्ञः पुरुषः इति अस्मिन् स्थले राज्ञः इति पदस्य क्रियया सह प्रत्यक्षयोगाभावात् कारकत्वम् उपपन्नम् क्रियया सह अन्वयात् । |
| 6100002 | ष्ठयर्थः इति अपरः कारकभेदः इति अस्मिन् विषये प्रायेषु एव व्याकरणकण्ठग्रन्थेषु आलोचना विहिता वैयाकरणैः । |
| 6100002 | तद्विषये भगवता पाणिनिना सूत्रमेकम् सूत्रितम् तद्धि शेषे षष्ठी इति । |
| 6100003 | वैयाकरणशिरोमणिभूतकौमुदीकारैः अपि सूत्रस्य अस्य वृत्तिः तु एव प्रतिपादिता कारकप्रतिपादिकार्थव्यतिरिक्तः स्वस्वामिभावादिसम्बन्धः शेषः तस्मिन् वाच्ये षष्ठी स्यात् । |
| 6100003 | वाक्यपदीये अपि ग्रन्थकृता शेषस्वरूपम् एवम् प्रतिपादितम् । |
| 6100004 | कारकेभ्यः अन्यः कर्मादिविशेषलक्षणेभ्यः षड्भ्यः अन्यो यः सम्बन्धः सः शेषः इति उपयुक्तेतरवचनशेषशब्दाश्रयेण नागेशविरचिते शेखरे तु एवम् प्रतिपादितम् शेषस्वरूपम् शेष उक्तादन्यः । |
| 6100004 | सः च तत्त्वरूपेण स्वत्वादिसम्बन्धत्वेन रूपेण च । |
| 6100005 | कारकविभक्तीनाम् क्रियाजनकत्वसमानाधिकरणकर्तृत्वादिशक्तिरूपेण बोधकत्वम् । |
| 6100005 | तत्र अपि तन्मूलके सम्बन्धे विवक्षिते षष्ठी इति । |
| 6100005 | सः अयम् सम्बन्धः श्रुतायाम् अश्रुतायाम् वा क्रियायाम् सत्याम् एव विशेषणीभूतसम्बन्धिवाचकपदात् उत्तरम् जायमानया षष्ठया अभिधीयते । |
| 6100005 | तत्र उदाहरणम् अश्रुतायाम् क्रियायाम् इति अस्य राज्ञः पुरुषः इति । |
| 6100005 | तत्र राजकर्तृकपुरुषसम्प्रदानकदानक्रियानिरूपितम् राज्ञि कर्तृत्वम् पुरुषे च सम्प्रदानत्वम् इति क्रियाकारकपूर्वकः राजपुरुषयोः स्वस्वामिभावरूपः सम्बन्धः इति पूर्वभाविकारकत्वम् तयोः उत्तरावस्थायाम् अपि अनुगतम् इति एवम् भवति शेषाख्यः सम्बन्धः कारकत्वम् । |
| 6100006 | श्रुतायाम् क्रियायाम् अभिधीयते इति अस्य उदाहरणम् यथा मातुः स्मरति सर्पिषो जानीते इति आदि । |
| 6100006 | अत्र मातुः स्मरति इति अत्र कर्मत्वविवक्षायाम् मातरम् स्मरणम् इति अर्थः । |
| 6100006 | कर्मत्वस्य शेषत्वविवक्षायाम् तु देवदत्तकर्तृकम् मातृसम्बन्धिः स्मरणम् इति अर्थः । |
| 6100006 | राज्ञः पुरुषः इति आदौ तु अश्रुतायाम् क्रियायाम् सम्बन्धाभिधानम् । |
| 6100006 | अतः सम्बन्धस्य अपि कारकत्वम् वाक्यपदीयकुन्मते । |
| 6100006 | सम्बन्धस्य नाम षष्ठ्यर्थस्य वस्तुतः हि क्रियाम् विना न जायते कः अपि सम्बन्धः । |
| 6100006 | यथा राज्ञः पुरुषः इति अस्य विद्यते राज्ञि कर्तृत्वं पुरुषे च सम्प्रदानत्वं तत्तु अनुभवसिद्धम् । |
| 6100006 | तत्त्वम् इदम् उदाहरणेन स्पष्टीकर्तव्यम्। |
| 6100006 | यथा राजा भिक्षुकाय धनम् द्रव्यम् वा ददाति। |
| 6100006 | अतः अत्र राजनि स्वसत्त्वनिवृत्तिपूर्वकपरसत्त्वोत्पादनक्रियायाः कर्तृत्वं भिक्षुके च सम्प्रदानत्वम् इति गम्यते। |
| 6100006 | अत्र तु क्रियाकारकपूर्वकः राजभिक्षुकयोः स्वस्वामिभावादिसम्बन्धः शेषाख्यः षष्ठ्या अभिधीयते। |
| 6100007 | लोकप्रसिद्धानि कानिचन उदाहरणानि यथा वृक्षस्य शाखा देवदत्तस्य अधः इति आदौ तु वृक्षस्य अधिकरणत्वम् तथा परस्य तु देवदत्तस्य कर्तृत्वम् प्रतीयते। |
| 6100007 | एवम् च मातुः स्मरति भगवतः नारायणम् अनुकरोति इति आदिषु वाक्येषु मात्रादौ कर्मत्वम् विद्यते परन्तु तथा अपि कर्मणः विवक्षा षष्ठी विधीयते। |
| 6100007 | अत्रेदमवधेयम् यत् असतः वस्तुनः एव भवति अविवक्षा परन्तु वस्तुनि विद्यमाने सत्यपि कथमविवक्षा स्यात् इति समाधानम् प्रदर्शयते भाष्यकारौः कथम् पुनः सतो नामाविवक्षा यथा दृश्यते लोके एव केचन प्रयोगाः यथा अनुदरा कन्या तथा अलोमिकैडिका इति। |
| 6100007 | अत्र वक्तुराशयः हि एवं अलोमिका इति उक्ते सत्सु अपि बहुषु लोमसु अत्यन्ताल्पत्वात् लोम्नामसत्कल्पानि तस्याः। |
| 6100007 | एवम् एव विचार्य एव तदभिधानम् अलोमिका इति भवति। |
| 6100007 | एवम् एव अनुदरा कन्या इति अत्र अपि विद्यमाने अप्युदरे सूक्ष्मत्वात् क्षीणत्वाद्रा तस्या अनुदरेत्यैवमभिधा। |
| 6100007 | पूर्वस्य अपि सत्सु अपि लोमसु अल्पत्वात् परस्य च विद्यमानेषु अपि उदरे क्षीणत्वात् अनुदारिति एवम्बिधया अभिधया यथार्थस्य वर्जनम् परिलक्ष्यते परन्तु कृते वर्जने अपि यथार्थे अलोमिका इति आदिप्रयोगः तु न असाधुः। |
| 6100007 | उभयत्र अपि प्रयुक्तो नञ् तत्तु अल्पार्थे। |
| 6100007 | एवम् एव विद्यमानस्य उदरस्य लोम्नो वा प्रयोगे तु तत्सत्ता तिरोभूयते एवम् भूयः दृश्यते शासे अपि यथा विद्यमानम् अपि कर्म कर्मणः अविवक्षायाम् षष्ठीम् प्रयोजयति। |
| 6100007 | कदाचित् असतः अपि विवक्षा भवति। |
| 6100007 | तद्यथा समुद्रः कुण्डिका विन्ध्यो वर्धितकम्। |
| 6100007 | कुण्डिका नाम पात्रविशेषः। |
| 6100007 | अत एव अत्र कुण्डिकायाम् समुद्रः इति प्रयोगः न उपपन्नः। |
| 6100007 | यतः हि समुद्रस्य अतिविशालत्वात् तथा कुण्डिकायाः च अत्यन्ताल्पायतनत्वात्। |
| 6100007 | तथाहि कुण्डिकायाः अति प्रशस्ततया विशालतया वा तस्याम् समुद्रः इति व्यवहारः उपपद्यते। |
| 6100007 | अत्र तु वर्धितकम् नामान्नराशिः तस्य अतिविशालत्वात् विन्ध्यपर्वतसंज्ञा भवति। |
| 6100007 | सकलवैयाकरणगोष्ठिगदितम् तदिदम् शेषपदम् द्विविधम्। |
| 6100007 | तत्र शेषपदस्य द्वैविध्यम् अपि व्याख्याकृताम् पक्षद्वयम् अवाप्यते। |
| 6100007 | तत्र कारकाणामविवक्षा एव शेषपदम् इति भाष्यकारादिभिः प्रतिपादितम् मतम्। |
| 6100007 | तैस्तु एवम् उच्यते यत् कर्मादीनाम् सताम् अपि कथमविवक्षा सम्भवति सः अयम् शेषपदार्थः स्वस्वामिभावादिसम्बन्धरूपेण अधीयमाणः षष्ठीविभक्ति प्रयोजयति। |
| 6100007 | परन्तु इतरेषाम् नव्यानाम् मते तु स्वस्वामिभावादिसम्बन्ध एव शेष इति। |
| 6100007 | सः च सम्बन्धः कारकप्रातिपदिकार्थव्यतिरिक्तः सन् षष्ठीम् प्रयोजयति। |
| 6100007 | अस्तु तावत् अत्र आलोचनातः प्राप्नते यत् कः अपि सम्बन्धः समागते सति सः अयं सम्बन्धः अवश्यम् एव द्विनिष्ठः भवति तर्हि आलोच्यमाने अस्मिन् सूत्रे कस्मात् षष्ठी विधेया। |
| 6100008 | अत्र वार्तिकोक्तम् परम् नाम विशेषणम्। |
| 6100008 | यथा राज्ञः पुरुषः इति आदौ राजपुरुषयोर्मध्ये विद्यते स्वस्वामिभावादिसम्बन्धः। |
| 6100008 | तत्र पदयोः राजपुरुषयोर्वा षष्ठी विधीयते राजन् इति शब्दात् परन्तु कथन्न पुरुष शब्दात् इति इयम् आकाङ्क्षा तु स्वाभाविकी आयाति। |
| 6100008 | अतः अत्र मञ्जुषायाम् प्रतिपादितम् दृष्टिः यदपि सम्बन्धः षष्ठी उत्पत्तिः तु भेदकात्। |
| 6100008 | अत्र भेदकः नाम विशेषणम्। |
| 6100008 | अतः सम्बन्धे सति वाक्ये तत्र विधीयमानत्वात्। |

| 6100009 | विशेषणात् एव न तु विशेष्यपदात् । |
| 6100009 | विशेष्यपदात् तु प्रथमा एव वाच्या । |
| 6100009 | यच्च युगपदेवोभाभ्याम् विशेषणविशेष्याभ्याम् षष्ठी समुत्पद्यते तदा तत्र अपि उभयसम्बन्धिभ्यां षष्ठीद्वये सम्बन्धद्वयम् आयाति । |
| 6100009 | परन्तु तत् न युक्तम् । |
| 6100009 | अतः वाक्ये षष्ठी एकैव भवति सा च विशेषणात् उत्पद्यमाना अपि तु सा च षष्ठी सम्बन्धस्य वाचिकापि । |
| 6100010 | केचन वैयाकरणा उपदिशन्ति यत्तत्र परार्थो हि शेषपदस्य अर्थः । |
| 6100010 | परार्थो नाम विशेषणार्थः । |
| 6100010 | अतः राज्ञः पुरुष इति अस्मिन् उदाहरणे विशेषणस्य परार्थत्वात् एव तत्र राज्ञः इति अस्मिन् पदे शेषभावः प्रतीयते अतः ततः षष्ठी विधीयते । |

## Wikipedia (Paragraph number not marked)

| 1000006 | एतत् पश्चिमवेलायां काजळी नद्याः तटे विराजते। |
| 1000007 | तमस्य वर्षस्य जनगणनेः अनुसारं रत्नागिरिस्य जनसङ्ख्या। |
| 1000008 | अत्रत्या मुख्यभाषा मराठी। |
| 1000009 | रत्नागिरिस्य रत्नागिरी हापुस इति आम्रप्रजातिः भल्लातकः च प्रसिद्धाः। |
| 1000010 | रत्नागिरिनगरे विद्यमानः थिबा राजप्रासादः प्रसिद्धः। |
| 1000011 | ओडिशाविश्वविद्यालयः एकः दूरस्थः शिक्षा राज्यविश्वविद्यालयं भवति। |
| 1000012 | यः संबलपुर ओडिशा भारते तिष्ठति। |
| 1000013 | विश्वविद्यालयः ओडिशा राज्ये विधानमंडलः एक अधिनियमः स्थापितः। |
| 1000014 | अस्य विश्वविद्यालयस्य सम्पूर्णे ओडिशाराज्ये अधिकार वर्तते। |
| 1000015 | कटीलु कर्णाटकराज्ये विद्यमानं किञ्चन प्रसिद्धं तीर्थक्षेत्रम्। |
| 1000016 | दक्षिणकन्नडमण्डलस्य केन्द्रात् मङ्गळूरुनगरात् अनतिदूरे एव अस्ति कटीलु। |
| 1000017 | अत्रत्यं प्रसिद्धं श्रीदुर्गापरमेश्वरीमन्दिरम् उडुपी मङ्गळूरुराष्ट्रियमार्गस्य पार्श्वे अस्ति। |
| 1000018 | नन्दिनीनदीतीरे प्रशान्तस्थले सुन्दरः प्राचीनः देवालयः अस्ति। |
| 1000019 | एषः द्वीपदेवालयः। |
| 1000020 | गर्भगृहे शक्तिस्वरूपिणी लिङ्गाकारेण अस्ति। |
| 1000021 | ततः शङ्खचक्रधारिणी चतुर्भुजा वरदाभयमुद्रायुक्ता देवी लोहबिम्बरूपेण विराजते। |
| 1000022 | मूर्तिः लिङ्गस्य अपेक्षया प्राचीना न। |
| 1000023 | वैष्णवाः एव अत्र अर्चकाः शाक्ततन्त्रमार्गानुसारं पूजयन्ति। |
| 1000024 | अत्र विशेषदिनानि नाम मेषसंक्रमणस्य अनन्तरम् अष्टादिनानि। |
| 1000025 | तथैव सिंहशुक्रवारस्य पवित्रदिनम्। |
| 1000026 | देवालयस्य स्तम्भेषु सुन्दरशिल्पानि प्रसिद्धः शिल्पी रञ्जाळ गोपालशेणै उत्कीर्णवान् अस्ति। |
| 1000027 | मङ्गळूरुतः किं मी दूरे विद्यते। |

| 1000028 | तम् वर्षः ग्रेगोरी कालगणनायाम् एकः साधारण वर्षः आसीत्। |
| 1000029 | जनाः सितम्बर मासस्य पञ्चमे दिनाङ्के शिक्षकदिवसम् आचरन्ति। |
| 1000030 | स्वतन्त्रभारतस्य द्वितीयराष्ट्रपतेः डॉ सर्वपल्ली राधाकृष्णन् इत्याख्यस्य जन्मदिनम् एव शिक्षकदिवसः कथ्यते। |

A short sample of the summarization data:

| | |
|---|---|
| किन्तु कः असौ कात्यायन् इति तु न शक्यते। इदमित्थन्तया वक्तम्। केचित्पाणिनिशिष्यम् कात्यायनम् एव कच्चायनस्य कर्तारम् मन्यन्ते । अपरे तु तमशोकसमकालिकम् । केचितु कच्चायने चान्द्रव्याकरणकाशिकयोः प्रभावमालक्ष्य तत्काशिकान्तरवत इत्यपि मन्यन्ते यथार्थस्तु गुहायाम् निहितः । अश्वघोषस्यास्मरणादेव कच्चायनकारस्य तदवरवर्तित्वम् तु न एव सिध्यति। प्राकृतप्रकाशस्तु वाररुच एव मन्यते । वररुचिना हि पालिप्रकृतिकशब्दाः किमर्थ न निरूपिता इति विषयः अयमन्वेषणस्य । सः हि महाराष्ट्री पैशाची शौरसेनी मागधी प्राकृतान्येन व्याकरोति । असौ हि विक्रमसमकालिकत्रिपिटकस्य सङ्कलनम् वा प्रणयनमशोकात्प्रागेव सम्पन्नम् आसीत्इति तु तस्यम् श्रमणेभ्यः अप्पमादवग्गो श्रवणादेव ज्ञायते । | तत्र कच्चायनम् हि व्याकरणम् कात्यायनप्रणीतम् मन्यते |
| तस्य च वृत्तिभागः सङ्घनन्दिना प्रणीतः प्रयोगः च ब्रह्मदतेन न्यासस्तु विमलबुद्धिनेति । उक्तम् एव | कच्चायनो हि केवलम् सूत्राणाम् कर्ता |
| संस्कृतस्य प्राचीनतमग्रन्थाः वेदाः सन्ति। वेद शास्त्र पुराण इतिहास काव्य नाटक दर्शनादिभिः अनन्तवाङ्मयरूपेण विलसन्ती अस्ति एषा देववाक् । न केवलम् धर्म अर्थ काम मोक्षात्मकाः चतुर्विधपुरुषार्थहेतुभूताः विषयाः अस्याः साहित्यस्य शोभाम् वर्धयन्ति अपितु धार्मिक नैतिक आध्यात्मिक लौकिक वैज्ञानिक पारलौकिकविषयैः अपि सुसम्पन्नाः इयम् देववाणी। | संस्कृतवाङ्मयम् विश्ववाङ्मये अद्वितीयम् स्थानम् अलङ्करोति |
| कालान्तरे एतस्य लेखनम् ब्राह्मीलिप्या अभवत्। तदनन्तरम् एतस्य लेखनम् देवनागर्या आरब्धम् । | संस्कृतलेखनम् पूर्वम् सरस्वतीलिप्या आसीत् |
| स्वराक्षराणाम् उच्चारणसमये अन्येषाम् वर्णानाम् साहाय्य्यम् नापेक्षितम्। स्वयम् राजन्ते इति स्वराः। | अक्षराणि स्वराः व्यञ्जनानि च इति द्विधा विभक्तानि |

# Appendix F

Study using MURIL:

MURIL

- The publicly available checkpoints were used through MURIL-Share encoder-decoder combination.

- The following summaries were generated for each input summary thereby indicating that the input was not conditioning the output. More resources would be required to do this:

---

Example: 146
Document: सः विक्रमस्य द्वादशशताब्द्यन्तिमभागे मालावारप्रदेशो गृहीतजनुषा बिल्वमङ्गलेन कृष्णलीलांशुकापराभिधानेनाप्यनेके ग्रन्थाः प्रणीताः सन्ति येषु गोविन्दाभिषेकम् नाम द्वादशसर्गात्मकम् महाकाव्यम् प्रसिद्धम् अस्ति । काव्यम् इदम् श्रीचिह्नकाव्यमित्यपि प्रसिद्धम् । यथार्थनामेदम् काव्यम् श्रीकृष्णभक्तिवर्णनपरम् । अत्र प्राकृतव्याकरणस्य नियमा अपि सोदाहरणम् निरूपिताः ।

------------------------------------------------------------------

Gold: कृष्णलीलांशुकस्य अपरम् नाम बिल्वमङ्गलः इति

------------------------------------------------------------------

murilshare2/greedy: ##म् इति भवति आसीत् अभवत्ः अस्ति एव च वर्ततेनाम् सन्ताः न अपि अकरोत् स्मति◯तःस्यत् न्ियि शक्यतेन्तेनम् यत् आसन् स्वयतेाम् करोतिनि कृतम् तस्य्येयम्◯ानिवान्यति प्रान्यितुम् कृतवान् सहभ्याम् तेषाम् भवन्तिमवियत् कृतः कर्तुम्◯ भव उपुम् कृतेयाः उक्तन्◯ाय जातःतःय कार्यम्◯◯न ।

murilshare2/beam: आसीत्म् इति भवति अभवत्ः अस्ति एव च वर्ततेनाम् सन्ताः न अपि अकरोत् स्मति◯तःस्यत् न्ियि शक्यतेन्तेनम् यत् आसन् स्वयतेाम् करोतिनि कृतम् तस्य्येयम्◯ानिवान्यति प्रान्यितुम् कृतवान् सहभ्याम् तेषाम् भवन्तिमवियत् कृतः कर्तुम् भव् उपुम् कृतेयाः उक्तन् जातः◯यतःय कार्यम्◯◯न ।

Example: 112
Document: मांसस्य परीक्षणम् कृत्वा एव भक्ष्यम् । चरः शरीरावयवाः स्वभावः धातवः क्रियाः लिङ्गं प्रमाणं संस्कारः मात्रा इति आदीनि नवविधानि परीक्षणानि सन्ति । यदि परिक्षणम् न क्रियते तर्हि रोगाः भवितुम् शक्नुवन्ति ।

------------------------------------------------------------------

Gold   : एतादृशम् मांसम् कदाऽपि न एव भक्ष्यम्

------------------------------------------------------------------

murilshare2/greedy: ##म् इति भवति आसीत् अभवत्ः अस्ति एव च वर्ततेनाम् सन्ति नाः अपि अकरोत् स्मति◯तःत्स्य न्ियि शक्यतेन्तेनम् यत् आसन्यते स्वाम् करोतिनि कृतम् तस्य्यम्◯ ◯ानिवान्यति प्रान्यितुम् कृतवान् सहभ्याम् तेषाम् भवन्तिमवियत् कृतः कर्तुम् भव उपुम् कृतेयाः उक्तन् जातः◯ायतःय कार्यम्◯◯न ।

murilshare2/beam: आसीत्म् इति भवति अभवत्ः अस्ति एव च वर्ततेनाम् सन्ति नाः अपि अकरोत् स्मति◯तःस्यत् न्ियि शक्यतेन्तेनम् यत् आसन्यते स्वाम् करोतिनि कृतम् तस्य्येयम्◯ानिवान्यति प्रान्यितुम् कृतवान् सहभ्याम् तेषाम् भवन्तिमवि कृतःयत् कर्तुम्◯ भव उपुम् कृतेयाः उक्त जातः◯ायन्तःय कार्यम्◯◯न ।